\documentclass[10pt,twocolumn,switch]{article}
\usepackage{arxiv}

\usepackage{amssymb}

\usepackage{amsmath,amsfonts}
\usepackage[bbgreekl]{mathbbol}
\usepackage{algorithmic}
\usepackage{algorithm}
\usepackage{array}
\usepackage[caption=false,font=normalsize,labelfont=sf,textfont=sf]{subfig}
\usepackage{textcomp}
\usepackage{stfloats}
\usepackage{url}
\usepackage{booktabs}
\usepackage{graphicx}
\usepackage[pagebackref=true]{hyperref}

\usepackage{xcolor}
\usepackage{pifont} 
\usepackage{makecell}
\usepackage{graphicx}
\usepackage[linguistics]{forest}
\newcommand{\tabincell}[2]{\begin{tabular}{@{}#1@{}}#2\end{tabular}}
\usepackage{tikz}
\usetikzlibrary{matrix,shapes,arrows,fit,tikzmark}

\usepackage{mathtools}

\usepackage{multirow}
\usepackage{float}

\usepackage{tikz-dependency}

\def\state{x}
\def\out{y}
\def\inp{u}

\def\leb{\mathcal{L}}
\def\spn{\mathop{\rm span}\nolimits}

\usepackage{mathtools}
\newcommand{\defeq}{\vcentcolon=}

\newcommand{\loss}{\mathcal{L}}

\def\R{\mathbb{R}}
\def\C{\mathbb{C}}

\newcommand\parafango[1]{\noindent\textbf{#1.
\enspace}}

\def\span{\mathop{\rm span}\nolimits}

\newcommand{\mt}[1]{\textcolor{blue}{#1}}
\definecolor{dgreen}{rgb}{0, 0.5, 0}

\newcommand{\sigla}[1]{\texttt{#1}}

\newcommand{\din}{{d_{\text{in}}}}
\newcommand{\dout}{{d_{y}}}
\newcommand{\dstate}{{d_{\text{s}}}}
\newcommand{\dk}{{d_{\text{k}}}}

\newcommand{\dv}{{d_{\text{v}}}}
\newcommand{\dkern}{{d_{\text{ker}}}}
\newcommand{\dseq}{L}

\usepackage{pgfplots}
\usepackage{pgfplotstable}
\usepackage{amsmath}
\pgfplotsset{compat=1.18}
\usepackage{xcolor}

\newcommand{\e}[1]{\exp\left(#1\right)}
\newcommand{\similarity}[1]{\text{sim}\left(#1\right)}

\DeclareMathOperator*{\Id}{I}

\usepackage[final]{changes}
\definechangesauthor[name={matteo}, color=magenta]{per}
\makeatletter
\AddToHook{cmd/added/before}{\def\Changes@AuthorColor{magenta}}
\AddToHook{cmd/deleted/before}{\def\Changes@AuthorColor{red}}
\AddToHook{cmd/replaced/before}{\def\Changes@AuthorColor{blue}}
\makeatother

\title{
 State-Space Modeling in Long Sequence Processing: \\  A Survey on Recurrence in the Transformer Era
}

\author{ \href{https://orcid.org/0000-0002-9133-8669}{\includegraphics[scale=0.06]{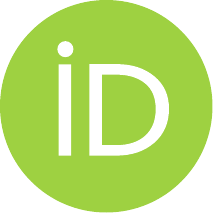}\hspace{1mm}Matteo Tiezzi$^{2}$}, 
\And \hskip -1.7cm 
\href{https://orcid.org/0009-0003-5138-9159}{\includegraphics[scale=0.06]{orcid.pdf}
\hspace{1mm}Michele Casoni$^{1}$},\\
 	\hskip -4cm $^{1}$ DIISM\\
	\hskip -4cm University of Siena\\
	\hskip -4cm 53100, Siena, Italy \\
\And \hskip -1.7cm
\href{https://orcid.org/0000-0002-9052-8743}{\includegraphics[scale=0.06]{orcid.pdf}\hspace{1mm}Alessandro Betti$^{3}$},\\
	\hskip -2.7cm $^{2}$ IIT\\
	\hskip -2.7cm Ist. Italiano di Tecnologia\\
	\hskip -2.7cm 16152, Genova, Italy \\
\And \hskip -2.6cm 
\href{https://orcid.org/0000-0001-6337-5430}{\includegraphics[scale=0.06]{orcid.pdf}\hspace{1mm}Marco Gori$^{1}$},\\
	\hskip -1cm $^{3}$ IMT\\
	\hskip -1cm Scuola Alti Studi\\
	\hskip -1cm 55100, Lucca, Italy \\
\And \hskip -2.7cm
\href{https://orcid.org/0000-0002-0415-0888}
{\includegraphics[scale=0.06]{orcid.pdf}\hspace{1mm}Stefano Melacci$^{1}$}\And   \vspace{-1cm} $\ $  \\
 \texttt{\scriptsize matteo.tiezzi@iit.it,m.casoni@student.unisi.it,alessandro.betti@imtlucca.it,marco.gori@unisi.it,stefano.melacci@unisi.it}
}

\date{}

\renewcommand{\headeright}{}
\renewcommand{\undertitle}{}
\renewcommand{\shorttitle}{A Survey on Recurrence in the Transformer Era}

\begin{document}

\twocolumn[\begin{@twocolumnfalse}
\maketitle

\vspace{-1.3cm}

\begin{abstract}
Effectively learning from sequential data is a longstanding goal of Artificial Intelligence, especially in the case of long sequences. From the dawn of Machine Learning, several researchers have pursued  algorithms and architectures capable of processing sequences of patterns, retaining information about  past inputs while still leveraging future data, without losing precious long-term dependencies and correlations. While such an ultimate goal is inspired by the human hallmark of continuous real-time processing of sensory information, several solutions have simplified the learning paradigm by artificially limiting the processed context or dealing with sequences of limited length, given in advance. These solutions were further emphasized by the  ubiquity of Transformers, which initially overshadowed the role of Recurrent Neural Nets. However, recurrent networks are currently experiencing a strong recent revival due to the growing popularity of (deep) State-Space models and novel instances of large-context Transformers, which are both based on recurrent computations that aim to go beyond several limits of currently ubiquitous technologies. The fast development of Large Language Models has renewed the interest in efficient solutions to process data over time. This survey provides an in-depth summary of the latest approaches that are based on recurrent models for sequential data processing. A complete taxonomy of recent trends in architectural and algorithmic solutions is reported and discussed, guiding researchers in this appealing research field. The emerging picture suggests that there is room for exploring  novel routes, constituted by learning algorithms that depart  from the standard Backpropagation Through Time, towards a more realistic scenario where patterns are effectively processed online, leveraging local-forward computations,  and opening new directions for research on this topic.


\end{abstract}

\keywords{Recurrent Neural Networks \and Transformers \and State-Space Models \and Long Sequences \and Learning over Time}

\vspace{1.2cm}

\end{@twocolumnfalse}]


\section{Introduction}
\vskip -0.3cm 
Human cognition is inherently intertwined with the seamless processing of sequential patterns \cite{elman1990finding,simon1963human,conway2001sequential}. Sequential data are  ubiquitous in the context of perception: from the flow of natural language during conversations to the sequence of cues processed in visual perception, and more. Mimicking the human ability to comprehend and learn from sequences has long been an aspiration within the realm of Artificial Intelligence (\sigla{AI}), with the ultimate goal of mirroring the human capacity to retain long-term insights from past experiences while remaining attuned to upcoming information \cite{sodhani2020toward,lipton2015critical}.
The relevance of processing sequential data is evidenced by the increasing amount of tasks tackled with Deep Learning for scientific and industrial applications, such as conversational AI \cite{brown2020language}, natural language understanding \cite{wang2018glue}, video representation learning and processing \cite{selva2023video}, lifelong and continual learning \cite{sodhani2020toward,ehret2020continual},
time-series analysis \cite{BianchiMKRJ17}, temporal graphs \cite{wu2020comprehensive},  and various other domains\cite{baccouche2011sequential,jurtz2017introduction}.

Early attempts to deal with sequential data date  back to the dawn of Machine Learning~\cite{mcculloch1943logical,rumelhart1985learning,hopfield1982neural,elman1990finding,DBLP:journals/neco/Pearlmutter89},
and they focused on the proposal of novel \textit{architectural} and \textit{algorithmic} solutions that departed from popular approaches  designed {for} static data, and were better suited to the context of processing sequences. Indeed, Recurrent Neural Networks (\sigla{RNNs}) go beyond feed-forward models for static data thanks to the way hidden states are structured, serving as a form of memory and being influenced  by previous-time self-connections \cite{elman1990finding,salehinejad2017recent}. Given the novelty of such architectural topologies, several ad-hoc learning rules were proposed \cite{williams1989learning,williams1990efficient}, among which the Backpropagation Through Time (\sigla{BPTT}) \cite{werbos1990backpropagation} algorithm emerged and became prominent. However, \sigla{BPTT} requires the recursive unfolding of  the network over the processed sequences, storing intermediate computations across  the whole input sequence, and virtually obtaining a ``deep'' feed-forward computational graph on which standard back-propagation of errors is applied \cite{ororbia2020continual}.

The limits and drawbacks of this solution become evident when the goal is to emulate abilities which are typical of human cognition, such as real-time/online sequence processing. Indeed, the unfolded networks represent very deep pathways to be traversed by error information. As a result, \sigla{RNNs} trained by \sigla{BPTT} suffer from \textit{vanishing gradients} \cite{bengio1994learning,sodhani2020toward} making \textit{credit assignment} much more difficult \cite{ororbia2018deep}, i.e., the task of computing the impact on
the overall error of the individual units in the network.
Early solutions often necessitated compromises to make learning affordable, simplifying the learning paradigm by reducing the context window (e.g., Truncated \sigla{BPTT} \cite{williams1990efficient}) or assuming the availability of complete sequences in advance.
Yet, the aforementioned issues and these simplifications constrained the ability to capture intricate long-term dependencies and correlations that characterize many real-world sequential datasets \cite{gu2021efficiently}.

\begin{figure*}[!ht]
    \centering
    \includegraphics[width=0.95\textwidth]{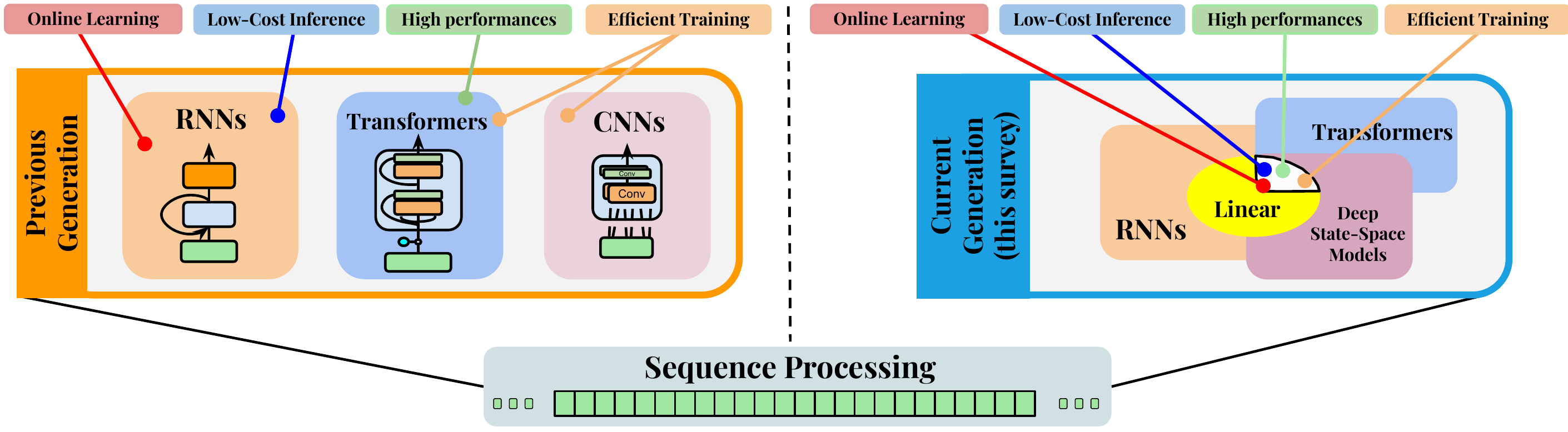}
    \caption{{\bf Left:} Since the advent of Transformers \cite{vaswani2017attention}, the neural network community has promoted these architectures as alternatives to recurrent (RNNs) and convolutional (CNNs) models for sequence processing, i.e., {\it``attention is all you need''}, thus they are represented by disjoint boxes in the picture, with the main features characterizing each of them listed at the top. {\bf Right:} In the current generation of neural models, described in this survey, attention has ``married'' recurrence, generating a significant intersection between the different ``worlds''. RNNs are state-space models, and in recent years we have observed the emergence of several variants of (deep) State-Space Models, where linearity (yellow set) is one of the main keys behind their success (together with element-wise recurrence and gating functions), and convolutional layers are now parts of various architectures. Interestingly, linearity intersects all the three families of models (RNNs, Transformers, Deep State-Space Models---the latter are almost fully based on linear models). The increasing interest in lifelong-Learning and applications running on limited-computation devices and with privacy constraints, such as instances of edge computing, is shifting the interest toward online learning (previously a prerogative of stateful/RNNs models). The white area is the meeting point of all the important features listed on top, and where there might be more room for novel research activities.} 
    \label{fig:main}
\end{figure*}

However, the ultimate goal of mimicking the human capacity to learn over sequences in real-time has inspired several researchers.
Amongst others, Williams and Peng proposed\textit{``an on-line algorithm, designed to be used to train a network while it runs; no manual state resets or segmentations of the training stream are required''} \cite{williams1990efficient}, as well as other well-known variants such as Real Time Recurrent Learning (\sigla{RTRL}) \cite{williams1989learning}, which, however, suffer from scalability issues \cite{javed2023online}.
Architectural solutions were investigated, for instance through the introduction of gating mechanisms (Long Short Term Memories,  \sigla{LSTMs} \cite{hochreiter1997long}) that partially improve the control over vanishing gradients and the error flow. Remarkably, the original work on gating in \sigla{LSTMs} \cite{gers2000learning} was devised as a remedy to the fact that        
\textit{``a continual input stream eventually may cause the internal values of cells to grow without bound''}.
Still, none of these solutions {was} capable of overcoming the challenges posed by limited computational resources and the quest for efficiency in sequence processing in \sigla{RNNs}, ultimately resulting in their inability to process sequences longer than a few thousand steps \cite{voelker2019,li2018independently}.
Going beyond instances of \sigla{RNNs}, still in the context of neural networks, the scientific literature includes approaches to long sequence processing that are based on Convolutional Neural Networks (\sigla{CNNs})---see \cite{he2023unified} and references therein.
\sigla{CNNs} enable the capability of parallelizing inference/learning over the temporal dimension when the entire sequence is available in advance \cite{bai2018empirical}. The local nature of the convolution results from the design choice of using filters that span a limited time range around each time instant. However, dealing with small filters hinder s the ability to capture very long-term dependencies, a role that is instead fulfilled by stacking multiple convolutional layers.

The ubiquity of Transformers \cite{vaswani2017attention} in recent years has dominated the sequence processing scenario due to several advantages over other existing architectures. In fact, the training procedure is parallelizable on modern hardware when the full sequence is available \cite{zhuang2023survey}. The self-attention mechanism introduces several advantages, such as the ability to handle both local and long-range relations \cite{tay2020long}, while completely avoiding the vanishing gradient issue, thanks to the direct connection of any token pairs in the sequence.
However, the {\it quadratic} complexity characterizing self-attention yields a significant computational cost and memory demand, particularly pronounced when handling long input sequences, which becomes a strong limitation in the case of on-the-edge devices with limited computational resources. These issues have instigated a profusion of research endeavors aimed at improving the scalability attributes of Transformers, often entailing trade-offs with certain traits that underlie their efficiency \cite{wang2020linformer,dao2022flashattention}.
Several of these approaches are inspired by insights derived from \sigla{RNNs}, that are fast-inference machines that scale {\it linearly} with the sequence length~\cite{peng2023rwkv}.


{In parallel to these developments, and during the same years in which Transformers were gaining widespread adoption, an alternative and independent research direction emerged, based upon the usage of State-Space Models (\sigla{SSMs}) for sequence modeling. Inspired by methods developed for continuous-time recurrent neural networks \cite{voelker2019, gu2020hippo}, this line of work proposed \sigla{SSMs} as a principled way to handle long-range dependencies while preserving efficiency. In particular, a subclass of linear SSMs gained traction, promoting the idea of removing nonlinearities from the state-update rule to enable efficient computations \cite{gu2021efficiently}.}
When discretized with ad-hoc integrators, continuos time \sigla{SSMs} were used as additional modules to gated \sigla{RNNs} \cite{voelker2019, gu2020hippo}. This intuitions inspired a plethora of works aiming at injecting linear \sigla{SSMs} into Deep Architectures \cite{gu2021combining, gupta2022simplifying,smith2022simplified}, frequently considering diagonal recurrent matrices  and further blurring the line between these model classes. More recently, the authors of \cite{orvieto2023resurrecting} proposed a recurrent model that bridges \sigla{RNNs} and the intuitions behind the aforementioned categories of \sigla{SSMs}, referred to as Linear Recurrent Units (\sigla{LRUs}). Basically, they showed  that appropriately parametrizing and structuring \sigla{RNNs} (linear and diagonal recurrence, specific initializations, etc.) the advantages of  Deep-\sigla{SSMs} can be fully exploited also in standard \sigla{RNNs}. When paired with appropriate gating functions and very structured deep networks, these models can reach state-of-the art results in language modeling \cite{de2024griffin}.
While it is useful to talk about Deep-\sigla{SSMs} and \sigla{RNNs} in a separate manner to better emphasize the recent literature on linear models (frequently based on diagonal recurrent matrices and additional neural layers on top of them), we note that {\sigla{RNNs} are also instances of state-space models}.
We summarize in Table~\ref{tab:ingredients} the aforementioned ingredients constituting the novel components of current solutions, which we will explore in greater detail in the remainder of the paper.
While previous approaches did not consider many of these architectural components—such as linear and element-wise recurrence and gating mechanisms— they have proven fundamental for effectively managing increasingly demanding long-term sequences.
When the sequence length is taken to the extreme, i.e., dealing with potentially infinite sequences, the current learning paradigm based upon \sigla{BPTT} must be reconsidered by considering online learning scenarios.

\begin{table*}[t]
\centering
\caption{Features of previous (upper part) and current generation of neural models (bottom part), from the perspective of recurrence.
While previous generation did not consider many of those features that usually characterized recurrent models, the current generation (Transformers, state-space models, modern \sigla{RNNs}) share several of such features (linear and element-wise recurrence, gating mechanisms). This trend have proven fundamental when considering extremely long-term dependencies, both in term of performances and scalability. 
}
\label{tab:ingredients}
\resizebox{0.8\textwidth}{!}{\begin{tabular}{clcccc}
\toprule
\bf   &    & \bf \tabincell{c}{Linear Rec.} & \bf Element-wise Rec.    & \bf \tabincell{c}{Gating}  & \bf Online Learning\\
\midrule
\multirow{3}{*}{\rotatebox{90}{\scriptsize\textsc{\makecell{Previous \\ Gen}}}} & Transformers        & - &  - & - & - \\
& \sigla{RNNs} & - &  \ding{51} & \ding{51} & \ding{51}  \\
& ConvNets (for sequences) & - &  - & - & -  \\

\midrule
\multirow{4}{*}{\rotatebox{90}{\scriptsize\textsc{\makecell{Current \\ Gen}}}}
& Sec.\ref{sec:transformers}:  Transformers embracing Recurrence & \ding{51} &  \ding{51} & \ding{51} & -  \\
&Sec.\ref{sec:ssm}:   Deep State-Space models (\sigla{SSMs}) & \ding{51} &  \ding{51} & \ding{51} & -  \\
&Sec.\ref{sec:other_rnns}:   Enhancing \sigla{RNNs} & \ding{51} &  \ding{51} & \ding{51} & -  \\
&Sec.\ref{sec:forward}:   Learning in \sigla{RNNs} & \ding{51} &  \ding{51} & \ding{51} & \ding{51}  \\

\bottomrule
\end{tabular}}

\end{table*}

\parafango{Motivation and Scope} In the present era, a confluence of factors, including the rise of large-scale language models \cite{brown2020language} and the need to optimize their performance during both training and inference, has resulted in a resurgence of interest in recurrent architectures \cite{orvieto2023resurrecting,peng2023rwkv} and their optimization schemes, marking a new chapter in the narrative of sequence processing, going beyond the \textit{``attention is all you need''} conjecture \cite{vaswani2017attention} (Figure~\ref{fig:main}).
This survey presents a comprehensive exploration of this recent evolution, covering the latest architectural solutions inspired by recurrent computations, ranging from Transformers with extensive context to the rekindling of interest in linear recurrent networks, driven by multiple variants of deep state-space models.
Meanwhile, innovative learning algorithms that challenge the conventional Backpropagation Through Time (\sigla{BPTT}) have emerged. These algorithms represent a significant shift from the traditional methodology, addressing the practical need for online sequence processing \cite{marschall2020unified}. They leverage localized and forward-facing strategies to navigate the nuances of real-time pattern recognition \cite{javed2023online}.
The primary goal of this survey is to offer an exhaustive taxonomy of contemporary trends in both architectural design and algorithmic innovation for sequence processing with recurrent architectures. By shedding light on the cutting-edge techniques, we aim to provide researchers with a comprehensive guide through the dynamic landscape of sequence processing, guiding them in their pursuit of effective solutions.
As the \sigla{AI} community stands on the threshold of unprecedented opportunities catalyzed by the advent of large language models, we invite researchers to embark on this journey into the intricate realm of sequence processing.
This survey also highlights emerging research directions that lie at the intersection of online learning from sequential data and lifelong learning \cite{mai2022online}.
Figure~\ref{fig:main} summarizes the big picture of this paper, emphasizing the{``marriage''} between \sigla{RNNs}, Transformers, and the evolution of state-space models based on linear units.
In a nutshell, stateful models are not only back in the context of learning from sequential data, but they also bring together approaches that were previously considered distinct by the scientific community.
New research opportunities may emerge  at the intersection with online lifelong learning.


\begin{figure*}[!ht]
    \centering
    \includegraphics[width=\textwidth]{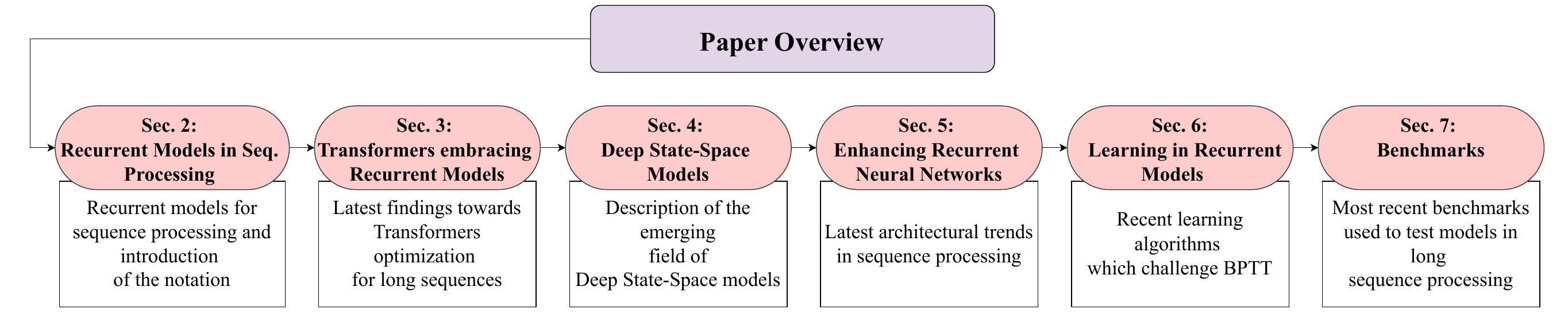}
    \caption{Overview of the organization of this survey.}
    \label{fig:Sec1-guide}
\end{figure*}

\parafango{Paper Overview} Figure~\ref{fig:Sec1-guide} describes the way this paper is organized. In each section that surveys multiple categories of models, we will further provide a picture to summarize its contents and main methods.
Moreover, we  include references to previous efforts in surveying sequence processing based on recurrent neural networks.
\added{In Table~\ref{tab:symbols}, we report the main mathematical symbols used throughout the paper, together with their descriptions and dimensionalities.}
We start this survey by describing, in Section~\ref{sec:related}, recurrent models for sequence processing and introducing the notation we will use in the remainder of this manuscript.
We will formalize our derivations using a notation that is applied consistently across apparently different models, to facilitate comparisons among them.
Section~\ref{sec:transformers} surveys the latest findings regarding the optimization of Transformers for long sequence processing, focusing on methods that aim to improve self-attention mechanisms inspired by recurrent approaches.
Section~\ref{sec:ssm} provides a thorough description of the emerging field of deep state-space models.
In Section~\ref{sec:forward}, we focus on learning dynamics, describing the recent learning algorithms that challenge the conventional Backpropagation Through Time (\sigla{BPTT}), in pursuit of local and forward-facing methods designed to emulate the human ability to process sequences in an online manner.
Then, in Section~\ref{sec:benchmarks}, we survey the most recent benchmarks used to evaluate how well models capture long-term dependencies.
Finally, in Section~\ref{sec:discus}, we analyze some open problems and issues that have been recently pointed out in the context of the described works, as well as highlight possible future avenues for research, and we draw our conclusions in Section~\ref{sec:conclusions}.

\begin{table}
\caption{\added{Summary of the main mathematical symbols used throughout the paper, including RNNs, Transformers, and Deep State-Space Models, along with their description and dimensionality.}}
\centering
\small
\begin{tabular}{lll}
\toprule
\textsc{Symbol} & \textsc{Description} & \textsc{Dimension} \\
\midrule
\( L \) & Sequence length & - \\
\( x \) & Hidden state & \( d_s \) \\
\( u \) & Input vector & \( d_{\text{in}} \) \\
\( y \) & Output vector & \( d_y \) \\
\( A \) & State transition matrix & \( d_s \times d_s \) \\
\( B \) & Input matrix & \( d_s \times d_{\text{in}} \) \\
\( C \) & Output matrix & \( d_y \times d_s \) \\
\( D \) & Feedthrough matrix & \( d_y \times d_{\text{in}} \) \\
$\theta$& Learnable parameters &-\\
$\sigma$& Activation function &-\\
\midrule
\( q, k \) & Query, key, value vectors & \( d_k \) \\
\( v \) & Value vectors & \( d_y \) \\
\( U \) & Stacked input (sequence) & \( L \times d_{\text{in}} \) \\
\( Q, K \) & Stacked query/key (sequence) & \( L \times d_k \) \\
\( V \) & Stacked value (sequence) & \( L \times d_y \) \\
\( W_q, W_k \) & Linear query/key projection matrices & \( d_k \times d_{\text{in}} \) \\
\( W_v \) & Linear value projection matrices & \( d_y \times d_{\text{in}} \) \\
\bottomrule
\end{tabular}
\label{tab:symbols}
\end{table}

\section{Recurrent Models in Sequence Processing}
\label{sec:related}

In several real-world scenarios, data is organized in a sequential manner, where ordering matters and requires appropriate computational models.
This is the case of natural language text, speech, vision, time series, and others.
Leveraging the temporal relations and ordering of patterns can help in discovering many additional cues in data \cite{elman1990finding}.
Let us describe a sequence of $\dseq$ patterns with the notation
\begin{equation}
(u_1,\dots, u_{t-1}, u_t, u_{t+1}, \dots, u_\dseq),\end{equation}
where $\dseq$ is a positive integer that represents the sequence length and $u_t \in \R^{\din}$, $t=1,\dots, \dseq$, is the $t$-th pattern.\footnote{This holds both when $u_t$ is considered to be the external input of a computational model or the result of intermediate computations in a multi-layer network.}
Note that while $t$ is commonly interpreted as the index of a pattern in the sequence, in many settings (especially in the context of perception or  time-series) we also have the use of the time at which the sample was provided or, equivalently, of the length of the time interval that passed between consecutive patterns.
We also remark that, in principle, $\dseq$ could  be infinite.
In this paper, to keep the notation simple, we use $t$ both as the index of a pattern or step and as the time variable, depending on the context.\footnote{In the following, the symbol $t$ will be written as a subscript when referring to a discrete-time index, for example $u_t$, and enclosed in parentheses when representing a continuous-time variable, for example $u(t)$.}       

\parafango{Stateless Models} The most straightforward approach for handling sequences, which have been considered since before recurrent architectures became ubiquitous, is to explicitly inject time by means of a spatial representation (as discussed in \cite{elman1990finding}, for example). The order of events in a sequence is simply considered to be an additional dimension of the pattern, which can be processed in a single shot. This also allows to design models that process in parallel the information at the different time instants, i.e., without having to wait for computations on past data to finish. There exists several works in the late eighties described by this route \cite{elman1988learning,hanson1987parsnip}, which has been also followed by more recent works based on convolutions \cite{gu2021efficiently} and Transformers \cite{vaswani2017attention}. The former virtually slides fixed-length filters over the input sequence \cite{gehring2017convolutional,smith2022simplified}, the latter leverages positional encodings, in both  cases parallelizing computations over time \cite{dufter2022position}. 
Overall, these models are {\it stateless}, in the sense that they do not try to build and progressively update an internal representation of the sequence observed so far. As a consequence, they require the whole sequence to be available and/or to reconsider it all if it gets updated, thus not being suited, for example, for continuous online-learning~\cite{collectionlessAI}.    

\begin{figure}[hb]
    \centering
    \includegraphics[width=1.\columnwidth]
{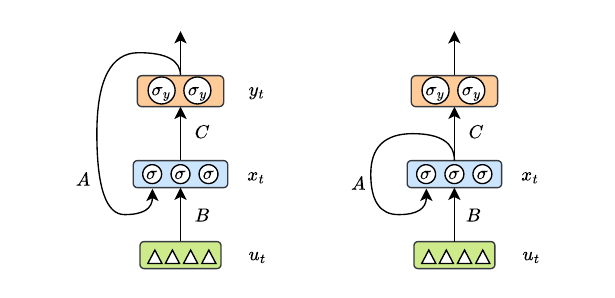}
    \caption{Jordan (left) and Elman (right) implementations of recurrent models. $A$, $B$, $C$ represent linear projections, and input (triangles), hidden (circles-blue boxed), output (orange-boxed) units are emphasized ($\sigma$, and $\sigma_{y}$ are activation functions).}
    \label{fig:jord_elm}
\end{figure}

\parafango{Recurrent Models} 
An alternative approach consists in focusing on {\it stateful} models that specifically embrace time in their computational scheme, developing a form of memory that gets updated over time (i.e., the {\it state}).
This is achieved by introducing feedback connections, yielding instances of what are commonly referred  as Recurrent Models.
Processing a sequence in Recurrent Models consists of processing one pattern {at-a-time}, following the original order of the data.
Internal units are used to model the so-called {\it state} or {\it context}, and, due to the feedback connections, they are responsible for integrating information from the current patterns and information from the previous time step, in order to update the state representation.      
As a result, the state effectively encodes the temporal characteristics of the sequential input observed so far \cite{elman1990finding}.
Several seminal works investigated this very natural direction \cite{rumelhart1985learning,werbos1988generalization}.
Jordan \cite{jordan1997serial} introduced a network with one-to-one connections from output units to state units, as shown in Figure \ref{fig:jord_elm}-left.
Feedback connections enable the hidden units to access the prior outputs of the network, thereby shaping subsequent behaviors.
This property grants the network a form of memory, facilitating the association of static patterns (referred to as ``Plans'') with serially ordered output patterns (sequences of ``Actions'').
Elman \cite{elman1990finding} proposed the best known instance of a Recurrent Model, where state units interact exclusively with internal nodes of the network, and not with the external world (i.e., state units are fed with the hidden information of the previous step, instead of what comes from the output layer), as shown in Figure \ref{fig:jord_elm}-right.
There exists an important connection between stateful Recurrent Models and the generic notion of {\it state-space model}, which has been used in many fields of engineering \cite{kitagawa1998self}.
This notion is common in control theory to model a dynamical system, and it is generic enough to cover a large number of possible implementations, including linear and non-linear systems, whether time-varying or time-invariant.
However, in the context of the majority of the literature in Machine Learning, it is rarely mentioned together with Recurrent Models.
{Koopman theory offers a principled alternative to \sigla{RNNs} for modeling nonlinear dynamical systems by embedding their evolution into a linear function space via the Koopman operator. This perspective enables more stable and interpretable models, as testified by works that incorporate  Koopman theory into deep learning through physics-informed autoencoders that outperform traditional \sigla{RNNs}~\cite{pmlr-v119-azencot20a}. See Section \ref{sec:forward} for further discussion on the connections between \sigla{RNNs} and dynamical systems.}

State-space models define the temporal evolution of state variables by first-order differential equations or difference equations.
The state changes over time as a function of its value at a given instant and the currently provided external input.
This definition intrinsically covers the typical feedback connection of Recurrent Models, which can be considered instances of state-space models, as we will formalize in what follows.


\parafango{Recurrent Neural Networks} Elman's architecture \cite{elman1990finding} was the pillar to the foundations of the widespread notion of Recurrent Neural Network (\sigla{RNN}), summarizing a neural model with a special hidden-layer including lookback connections, that we refer to as \sigla{RNN} layer. Given an input sequence $ (u_1, u_2, \dots, u_L)$, an \sigla{RNN} layer with $\dstate$-dimensional hidden state $x_t$ computes a sequence of $\dout$-dimensional outputs $(y_1, y_2,\dots, y_L)$ through a recurrent model,
\begin{equation}
    x_{t} = \sigma(Ax_{t-1}+ Bu_t), \quad y_t = \sigma_{y}(C x_t + D u_t),
    \label{eq:RNN}
\end{equation}
starting from some $x_0\in \R^\dstate$, with learnable \textit{state matrix} $A\in\R^{\dstate \times \dstate}$,   \textit{input matrix} $B\in\R^{\dstate\times \din}$, an \textit{output matrix} $C\in\R^{\dout \times \dstate}$ and an optional \textit{feed-through matrix} $D\in\R^{\dout \times \din}$.\footnote{$D\ne 0$ basically introduces a skip connection~(standard in modern architectures) that was not included in the original Elman network \cite{elman1990finding}.} We denote with $\sigma$ and $\sigma_{y}$ non-linearities on the state and output computation, respectively, often chosen to be the hyperbolic tangent or sigmoid activation. If $\sigma$ is the identity function, then we say the \sigla{RNN} layer is \textit{linear}.
The relation between
discrete computation of Eq.~\eqref{eq:RNN} and a continuous-time formulation becomes evident once we explicitly introduce the dependence on time and model the variation of the state as follows,
\begin{equation}\dot x(t) = -x(t)+\sigma(A x(t) +Bu(t)),
\, y(t)= \sigma_{y}(Cx(t)+Du(t)).
\label{eq:ssmsux}
\end{equation}
Moving the term $-x(t)$ to the left-hand side of the first equation yields an evident connection to Eq.~\eqref{eq:RNN}. This formulation is well-suited to trace another important link with the already introduced notion of {\it state-space model}. The general form of a state-space model is actually close to the one of Eq.~\eqref{eq:ssmsux}, when also including the direct dependence on time $t$ in both the equations, i.e., $\dot x(t) = f(x(t), u(t), t)$ and $y(t) = h(x(t), u(t), t)$, being $f$ and $h$ two generic functions. The discrete counterpart of the first equation, when considering time invariant systems, intrinsically yields the classic feedback of Recurrent Models.

\begin{figure}[h]
    \centering
    \includegraphics[width=0.9\columnwidth]{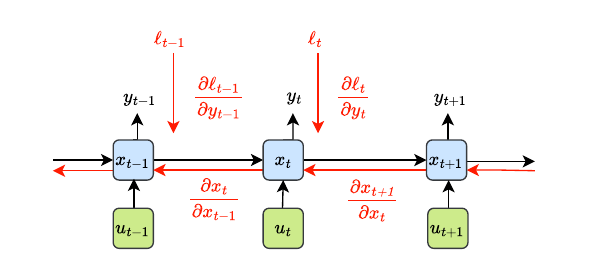}
    \caption{\sigla{BPTT}. This sketch \cite{pascanu2013difficulty} emphasizes the way the information propagates (black) and the way the learning signal goes backward in time and down from the instantaneous loss (red). We denote
by $\ell_t$ the loss value at time $t$. }
    \label{fig:bptt}
\end{figure}

\parafango{Learning in Recurrent Models}
Recurrent Models are commonly exploited to learn a mapping from the input sequence $ (u_1, u_2, \dots, u_L)$ to a target output, being it another temporal sequence  $(\hat{y}_1, \hat{y}_2,\dots, \hat{y}_L)$  or a single output vector $\hat{y}_L$. The former can be considered a more general formulation that includes the latter as a degenerate case. At each step $t$, an instantaneous loss function $\ell(y_t, \hat{y}_t)$ quantifies to what degree the predicted output $y_t$ matches the target output $\hat{y}_t$.
Let us collect the model learnable parameters in $\theta := \{A, B, C, D\}$.
\sigla{BPTT} \cite{rumelhart1985learning} is the de-facto standard algorithm to learn with Recurrent Models, and works by minimizing the following global empirical risk function over the whole sequence,
\begin{equation}
 \mathcal{L}(\theta) = \frac{1}{L} \sum_{t=1}^{L} \ell \left( y_{t}, \hat{y}_{t} \right),      
 \label{eq:loss_rnn}
\end{equation}
by gradient descent. Exploiting the chain rule \cite{werbos1990backpropagation} it can be shown that the gradient for each step $k$ is a sum of products of partial gradients,
\begin{equation}
\begin{aligned}
\frac{\partial  \mathcal{L}}{\partial \theta} &= \sum_{t=1}^{L} \frac{\partial \ell(y_{t}, \hat{y}_{t})}{\partial \theta} \\
&= \frac{1}{L}   \sum_{t=1}^{L} \frac{\partial \ell(y_{t}, \hat{y}_{t})}{\partial {y}_{t}} \frac{\partial {y}_{t}}{\partial x_{t}} \sum_{j=1}^{t} \Big( \prod_{s=j}^t \frac{\partial x_{s}}{\partial x_{s-1}}\Big) \frac{\partial x_{j-1}}{\partial \theta},
\end{aligned}
\label{eq:bptt}
\end{equation}
where $\prod_{s=j}^t \frac{\partial x_{s}}{\partial x_{s-1}}$ is the term that transports the error back in time from step $t$ to step $j$. A straightforward way to visualize \sigla{BPTT} is to virtually replicate the model at each time step, generating a deep feed-forward network over the input sequence. The model parameters are virtually ``copied'' at each time step, that is what is called temporal unfolding of the model.
In Figure  \ref{fig:bptt} we report  a sketch to emphasize the way the information propagates over time (black) and the main learning signals from the instantaneous loss function and those going backward over time (red), where for ease of notation we denoted the instantaneous loss $ \ell \left( y_{t}, \hat{y}_{t}\right) $ with $\ell_t$.

\parafango{Issues in BackPropagation Through Time} We can rewrite the term $ \prod_{s=j}^t \frac{\partial x_{s}}{\partial x_{s-1}}$ of Eq.~\eqref{eq:bptt} in the form of a product of Jacobi matrices \cite{pascanu2013difficulty},
\begin{equation}
\begin{aligned}
\prod_{s=j}^t \frac{\partial x_{s}}{\partial x_{s-1}} = \prod_{s=j}^t A' \text{diag}\big(\hat{\sigma}(x_{s-1})\big)
\label{eq:product}
\end{aligned}
\end{equation}

where $A'$ is the matrix transpose, $\text{diag}(\cdot)$ converts a vector into a diagonal matrix, and $\hat{\sigma}$ is the derivative of the activation function $\sigma$ in Eq.~\eqref{eq:RNN}, which is applied in an element-wise fashion to its input vector.
Unless the partial terms $\frac{\partial x_{s}}{\partial x_{s-1}}$ are close to 1, the product in Eq. \eqref{eq:product} could explode or vanish \cite{hochreiter1991untersuchungen}. In details, in the simplified case of a linear model (i.e., replacing $\sigma$  with the identity function) the power iteration method helps in deriving tight boundaries. In particular, a sufficient condition for {long term information to vanish} is that of having the largest eigenvalue of the recurrent weight matrix $A$ smaller than one, i.e. $\lambda < 1$. Conversely,  $\lambda > 1$ is a necessary condition for gradients to explode (see \cite{pascanu2013difficulty} for further details). When gradients undergo vanishing during their backward propagation through time, the critical credit assignment aspect of backpropagation is compromised. In particular, the information about minor state changes in the distant past loses the capacity to influence future states. Conversely, when gradients explode, gradient-based optimization algorithms encounter substantial difficulties in traversing the cost surface. This is primarily due to the assumption that gradient-based optimization relies on that small parameter adjustments yield small changes in the objective function.
As the number of time steps considered in the sequence of states increases, the amplifying or diminishing effects associated with the state-to-state transformations at individual time steps can grow exponentially, making it difficult for these models to capture long-range dependencies in the data.

\parafango{Properties of Recurrent Models} The sum-product of $\dseq$ terms in Eq.~\ref{eq:bptt} results in  $\mathcal{O}(\dseq^2)$ scaling for each processed sequence.
In practice, efficient gradient propagation schemes have been devised to achieve a cost linear in the length of the sequence \cite{kag2021training,gruslys2016memory}.
In terms of memory consumption, \sigla{BPTT} requires storing all the $\dseq$ intermediate states, resulting in a $\mathcal{O}(\dseq)$ overhead (see also \cite{gruslys2016memory} for alternative methods to achieve more memory-efficient \sigla{BPTT}).
Since \sigla{RNN} are stateful models, it is not possible to parallelize their inference/training procedure over the time instants to which the components of the sequence belong.\footnote{See Section \ref{sec:ssm} for special cases that can be parallelized.}
Computing the $t$-th state is conditioned on the current input $u_t$ and the preceding state $x_{t-1}$, which summarizes all the past knowledge and behaves as an information bottleneck.     
In other words, \sigla{RNNs} are {\it causal} models, since they can only leverage past information available in the input sequence.
Indeed, to produce an output $y_{t+1}$, the model can only access past information up to $u_{t}$.
A major advantage of this computational scheme is that, during inference, \sigla{RNNs} only require constant computational and storage resources per time step, i.e., $\mathcal{O}(1)$.       
From the theoretical point of view, \sigla{RNNs} have been proven to be able to simulate Universal Turing Machines \cite{siegelmann2012neural}, i.e., they can combine symbolic and sub-symbolic capabilities by running algorithms on
 a neural substrate. In this direction, some works introduced the unrealistic assumption of neurons with unbounded precision that equals the number of symbols used in the Turing tape \cite{siegelmann1992computational,siegelmann2012neural}. Chung et al. \cite{chung2021turing} relaxed such assumptions by leveraging a dynamically growing memory module made of neurons of fixed precision.
Additionally, \sigla{RNNs} are Universal Approximators \cite{schafer2006recurrent}, i.e., they are able to  approximate any open dynamical system with an arbitrary accuracy (see \cite{li2022approximation} and references therein for further Universal Approximation results for both discrete and continuous \sigla{RNNs}). {We further mention several studies on the way recurrent models (with trained and untrained state transition function) handle memorization, such as those that fall in the context of Echo State Networks \cite{gallicchio2017deep}.
The reader can refer to Section \ref{sec:forward} for further details on the literature on this topic.}



\textbf{Other Surveys on Recurrent Models.}  
Due to the large popularity of \sigla{RNNs}, there are several books \cite{graves2012supervised}, surveys, and reviews that describe their properties and architectural advances.
Detailed overviews were included in recent surveys \cite{lipton2015critical,salehinejad2017recent}, covering models up to 2015--2017, respectively. An in-depth analysis of gating-based architectures has been given in \cite{yu2019review}. Bianchi et al. \cite{BianchiMKRJ17} provide a general overview on \sigla{RNNs} and learning algorithms before focusing on short-term load forecasting.
The \sigla{BPTT} algorithm and its biological plausibility are investigated in \cite{lillicrap2019backpropagation}.
An important branch of research looks towards the discovery of online learning algorithms in the context of \sigla{RNNs}, and \cite{marschall2020unified} proposes an interesting framework that summarizes  recent findings. There are also surveys  that focus on the application of \sigla{RNNs} in confined areas, such as continual \cite{cossu2021continual,ehret2020continual} and online learning \cite{parisi2020online}.

\parafango{Recent Findings} The usage of Recurrent Models for sequence processing became less popular since 2017, due to the already established popularity of Convolutional models \cite{bai2018empirical} and, more importantly, due to the growing ubiquity of Transformer-based solutions \cite{vaswani2017attention}, driven by the ease of parallelizing training over the time instants on which the input sequences are defined. However,processing long sequences is hard and expensive with such architectures, fostering the need for novel paths to achieve efficient architectures \cite{tayefficient}.
As a result, the scientific literature has recently experienced a resurgence of interest in \sigla{RNNs} \cite{katharopoulos2020transformers,orvieto2023resurrecting}. Transformers' quadratic complexity with respect to the sequence length has pushed towards the discovery of novel methods to optimize inference \cite{tayefficient,LIN2022111}, and many solutions based on stateful recurrent computations have been introduced \cite{katharopoulos2020transformers,peng2023rwkv,polihyena,sun2023retentive}. At the same time, the need to preserve extremely long-term dependencies on sequences has led to the adoption of the so-called (deep) State-Space Models (\sigla{SSMs})  \cite{gu2020hippo,smith2022simplified}. This research field originated from the existing knowledge on \sigla{RNNs} \cite{voelker2019}, but then it took its own direction. It has been very recently formally reconnected to \sigla{RNNs} by a careful architectural design \cite{orvieto2023resurrecting,de2024griffin}.  This is not surprising, since, as anticipated, \sigla{RNNs} are indeed instances of \sigla{SSMs}.
{In this section, we established the foundational dichotomy between stateless sequence processors and stateful recurrent models, formalised the Elman-type \sigla{RNN}, and analysed \sigla{BPTT} together with its vanishing/exploding gradient pathology. The discussion highlighted \sigla{RNNs}’ theoretical universality, their $\mathcal{O}(1)$ inference cost, and their recent resurgence prompted by the scaling limits of Transformers. These insights set the stage for hybrid approaches that attempt to combine recurrent efficiency with modern attention mechanisms.}

{The classical \sigla{RNN} framework outlined above clarifies both its expressive power and its well-known training bottlenecks (vanishing/exploding gradients and limited parallelism). These limitations have recently motivated a wave of research that injects recurrence into the dominant Transformer family, seeking the best of both worlds—parallel training and explicit long-range memory. The next section surveys this convergence and shows how “attention meets recurrence”.}

\section{Transformers Embracing Recurrent Models}
\label{sec:transformers}

\definecolor{level1}{HTML}{f48e82}%
\definecolor{level3}{HTML}{63aff3}%
\definecolor{level2}{HTML}{74df7b}%
\definecolor{level4}{HTML}{ffe080}%
\definecolor{level5}{HTML}{cafcda}

\begin{figure*}[h!]
    \centering
    \includegraphics[width=\linewidth]{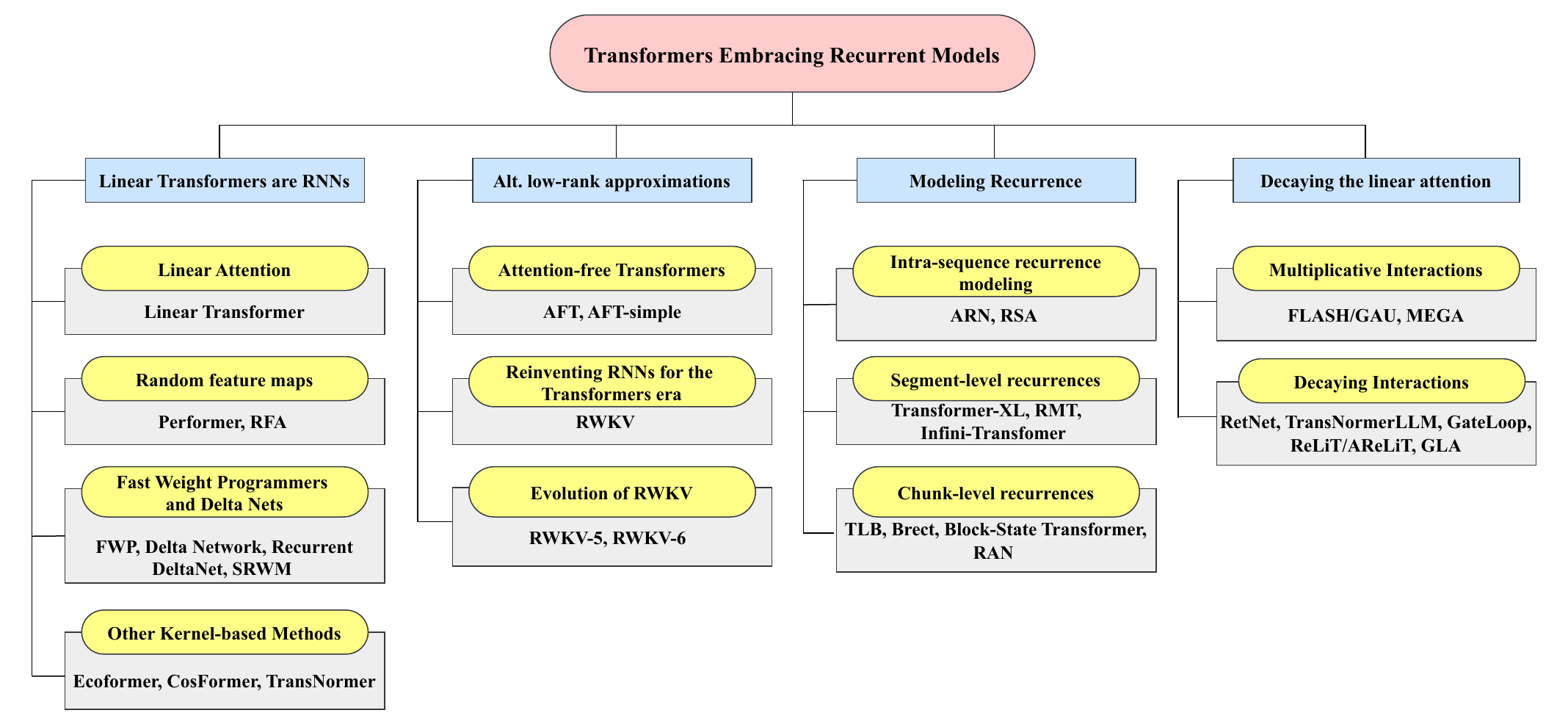}
    \caption{Conceptual overview of the organization of Section~\ref{sec:transformers},which categorizes Transformer-based models according to architectural and functional aspects. We report the names of the subsections (blue boxes) and the covered categories of models (yellow). The names of some representative methods are also indicated for each category (gray). }
    \label{fig:Sec3-guide}
\end{figure*}

Transformers \cite{vaswani2017attention} emerged as a disruptive alternative to Recurrent Models, promoting the idea of going beyond stateful models that process the sequence one element at a time, motivated by the {\it ``attention is all you need''}  motto. Transformers are able to handle the elements of an  input sequence \cite{tayefficient} in parallel and to capture both  local and long-range dependencies. These properties are due to their self-attention-based computational scheme, which compares all the possible pairs of input patterns constituting the sequence. In the following, we present a brief analysis aimed at summarizing the properties of self-attention and subsequently underline its drawbacks. Finally, we will delve into the latest findings on self-attention variants and alternatives that, perhaps unexpectedly, can be seen through the lens of Recurrent Models.
Figure~\ref{fig:Sec3-guide} showcases the organization of this section.

\parafango{Transformers and Self-Attention} Given the input sequence $(u_1, u_2, \dots, u_\dseq)$ with  $u_t \in \R^\din$, $t=1,\ldots,^\dseq$,   
Transformers implement a sequence-to-sequence function $\mathrm{tr}\colon\R^{\dseq \times \din} \to \R^{\dseq \times \dout}$, that yields $(y_1, y_2, \dots, y_\dseq)$. The whole function 
 $\mathrm{tr}$ is based on a self-attention procedure, followed by a feed-forward network, and it is commonly applied multiple times in a multi-layer fashion.\footnote{We purposely avoid describing these components and others, such as skip connections, layer normalization, and multi-head attention. We also do not mention the distinction between Transformer encoders and decoders, since they do not introduce useful information for the point we make. Please refer to \cite{tayefficient} and references therein for detailed descriptions of the Transformer architecture.}
Self-attention acts across the temporal dimension of the input sequence, evaluating pairwise interactions between the input components, regardless of their positions in the sequence,\footnote{Positional embeddings are commonly exploited to make Transformers position-dependent.} 
and it is evaluated (in parallel) on each element of the input sequence. When evaluated on a generic $u_t$, it returns a sum over $(u_1, u_2, \dots, u_\dseq)$, weighed by scores that depend on the similarities between $u_t$ and the $\dseq$ elements of the input sequence. Formally, 
\begin{equation}
\begin{aligned}
   y_t &= \sum_{i=1}^\dseq \frac{ \similarity{q_t, k_i} }
                {\sum_{j=1}^\dseq \similarity{q_t, k_j}} v_i,\quad t=1, \ldots, \dseq,
    \label{eq:general-attn}
\end{aligned}
\end{equation}
where $q_{\cdot}$, $k_{\cdot}$, $v_{\cdot}$ are the so-called {\it queries}, {\it keys} and {\it values}, respectively, computed by projecting (three linear projections) the elements of the input sequence using three trainable matrices $W_q , W_k \in \R^{\dk \times \din}$ and  $ W_v \in \R^{\dout \times \din}$, where $\dk$ is the keys and query size while $\dout$ is the self-attention output size. We have $q_z=W_q u_z$, $k_z=W_k u_z$ and  $v_z=W_v u_z$, for $z = 1,\ldots,L$. 
The similarity function $\similarity{q_{\cdot},k_{\cdot}}$ returns a positive score that increases with the similarity between its arguments.
The canonical choice for the self-attention function is the \textit{softmax dot-product} (multiplicative) attention, where the similarity score is computed as the exponential of the dot
product between a query and a key, $\similarity{q_{\cdot},k_{\cdot}} = \e{{q_{\cdot}' k_{\cdot}/\sqrt{d_k}}}$, where $'$ is the transpose operator and all mono-dimensional arrays are assumed to be column vectors (here and in the rest of the paper). 
Before going into further details, we introduce the matrix notation that we will use throughout this section to simplify the descriptions. 
Matrix $U$ is composed of $L$ rows, each storing the transpose of $u_t$, for $t=1,\ldots,L$. 
Similarly, for all the already introduced symbols that are associated with the $L$ elements of the input sequence, their uppercase versions (with no subscripts) indicate matrices with $L$ rows (e.g., in the case of $y_t$, $q_t$, $k_t$, $v_t$, $t=1,\ldots,L$, we have matrices $Y$, $Q$, $K$, $V$, respectively).
We can rewrite Eq.~\ref{eq:general-attn} in matrix notation as follows,
\begin{equation}
\begin{aligned}
    Q &= UW'_q, \quad K = UW'_k, \quad V = UW'_v,\\
    Y &= \mathrm{softmax}\left(\frac{QK'}{\sqrt{d_k}}\right)V
\end{aligned}
\label{alg:parallel_attention}
\end{equation}

where $\mathrm{softmax}(A)$ is the softmax function that operates on each row of $A$.
The above formulation is usually referred to as the \textit{parallel form} of attention, since it computes the Transformer outputs for all the time instants of the sequence (i.e., the full matrix $Y$) in parallel given the full input matrix $U$, meaning that all the sequence must be available in advance. This is one of the main advantages of Transformers: during training, which is usually performed using the full input sequence, computations can be efficiently parallelized over the sequence dimension—hence the term \textit{parallel training}.


\parafango{Causal Attention}
Unlike the aforementioned scenario where the full sequence is available in advance, there are settings where future information in the sequence cannot be accessed, thus requiring the implementation of a form of \textit{causal attention}.
 For instance, Transformers can work in an {\it autoregressive} setting, where the input sequence is progressively populated in an iterative manner (adding an element predicted by the Transformer itself at each time step, e.g., during inference while decoding). The standard self-attention of Eq. \eqref{eq:general-attn} is not causal, given that future positions $j>i$ can influence the current one. However, autoregressive Transformers can be obtained by causal masking, i.e., masking the attention computation such that the $i$-th position can only be influenced by preceding ones. In this causal setting, the parallel form of Eq.~\eqref{alg:parallel_attention} can be rewritten as  $Y=\mathrm{softmax}\left(\frac{QK'}{\sqrt{d_k}} \odot{M}\right)V$, where $M \in \R^{\dseq \times \dseq}$ is a mask that prevents the model from attending to future tokens, i.e., $M_{ij}=1$ if $i \ge j$ and $M_{ij}=-\infty$ if $i < j$. 
 During inference, causal Transformers can be interpreted as exploiting a \textit{recurrent form}, which is obtained by considering only the time steps up to the current one in Eq. \eqref{eq:general-attn}, i.e.,
$ 
   y_t = \sum_{i=1}^t \frac{ \similarity{q_t, k_i} }
                {\sum_{j=1}^t \similarity{q_t, k_j}} v_i,\  t=1, \ldots, \dseq.
$
Note, that, through this causal mechanism, attention is performed over sets of keys and values that grow over time: to compute  $y_{t-1}$, the sequences $(k_i)_{i=1}^{t-1}$ and $(v_i)_{i=1}^{t-1}$ are needed; to compute $y_t$, the additional key and value $k_t$ and $v_t$ must be considered in the computation as well, thus expanding the aforementioned sequences of keys and values over time also referred to as the \sigla{KV-cache} \cite{pope2023efficiently}.

\parafango{Computational Cost and Scaling Issues} As stated above, one of the key contributions  of the Transformer architecture is that inference can be parallelized over the temporal dimension \cite{tayefficient}, if the full sequence is available in advance, thus overcoming the intrinsic limitations of stateful Recurrent Models.
 In fact, all the matrix-by-matrix products in Eq.~\ref{alg:parallel_attention} can be computed as $L$ independent vector-by-matrix products. Thus, there is evident room for parallel implementations, given the full input $U$.
Despite the significant benefits brought by parallelization, Transformers are characterized by serious computational and memory costs. 
In fact, unlike Recurrent Models that keep track of the prior context via an explicit state with fixed size, Transformer self-attention has a ``global'' receptive field that directly attends the whole sequence at inference.
As a result, the main bottleneck is constituted by the self-attention, which scales quadratically with the sequence length, $\mathcal{O}(\dseq^2)$. Indeed, from the definition in Eq.~\ref{alg:parallel_attention}: computing $QK'/\sqrt{d_k}$ takes $\mathcal{O}(\dseq^2\dk)$; the $\mathrm{softmax}$ function involves elementwise exponentiation (and sum) of its matrix argument, taking $\mathcal{O}(\dseq^2)$ time, and division of each element by the corresponding row sum, which takes $\mathcal{O}(\dseq^2)$; multiplication of $\mathrm{softmax}(\cdot)$ and $V$ takes $\mathcal{O}\left(\dseq^2 \text{max}(\dk, \dout)\right)$ time \cite{keles2023computational}; overall, inference has $\mathcal{O}(\dseq^2)$ complexity in the sequence length..
In the aforementioned {\it autoregressive} setting, where the input sequence is progressively populated and keys and values grow over time in the \sigla{KV-cache}, the computational cost of autoregressive inference is $\mathcal{O}(\dseq_t)$, being $\dseq_t$ the length of the accumulated sequence at time $t$. 
All the aforementioned complexities make it harder to apply Transformers when scaling up the input sequence length. This issue has forced practitioners to artificially limit the sequence size, considering a custom context window length, thus hindering the model's capability of capturing long-term dependencies. 
Investigations on such limitations have spurred a number of efforts aimed at improving Transformers' scalability \cite{tayefficient}, with the final goal of retaining their performance, parallel computations, and merging them with efficient inference complexity. Among efforts aimed at optimizing the attention mechanism \cite{beltagy2020longformer, kitaev2020reformer,choromanski2020masked,dao2022flashattention} or completely replacing it \cite{zhai2021attention}, many recent works have been inspired by the advantages and 
peculiarities of Recurrent Models. 
We report an overview of the complexities of different variants of recent Transformer models in Table \ref{tab:complexity_comparison}, which will be discussed in the following.

\begin{table}
\caption{Complexities for sequences of length $L$, comparing representatives of the categories of Transformers discussed in Section~\ref{sec:transformers}, vanilla Transformers and Recurrent Models (Section~\ref{sec:related}). In order to bridge the discussion in the main text of paper, here we assumed $d_k=\din$, for the sake of simplicity. The symbol $T$ denotes the segment length in the case of \sigla{Transformer-XL}.}
\centering
\setlength{\tabcolsep}{2pt}
\small
\begin{tabular}{r|lccc}
\toprule
 & \textsc{Model} & \textsc{Recurrent} & \textsc{Time} & \textsc{Space}\\
\midrule
Sec.~\ref{sec:related} & Recurrent Net & Yes & $\mathcal{O}(\dseq\din^2)$ & $\mathcal{O}(\dseq\din)$  \\
Sec.~\ref{sec:transformers} & Transformer & No &  $\mathcal{O}(\dseq^2\din)$ & $\mathcal{O}(\dseq^2 + \dseq\din)$ \\
\midrule
\multirow{1}{*}{Sec.~\ref{sec:linear}} & Linear Trans.  & Yes & $\mathcal{O}(\dseq \din^2)$ & $\mathcal{O}(\dseq\din + \din^2)$ \\
\multirow{1}{*}{Sec.~\ref{sec:alt}} & RWKV  & Yes & $\mathcal{O}({\dseq\din})$ & $\mathcal{O}({\din})$ \\
\multirow{1}{*}{Sec.~\ref{sec:subrec}} & Transformer-XL  & Yes & $\mathcal{O}({T^2 \din})$ & $\mathcal{O}({T^2 + Tm})$ \\
\multirow{1}{*}{Sec.~\ref{sec:decay}} & RetNet  & Yes & $\mathcal{O}({\dseq\din^2})$ & $\mathcal{O}({\din^2})$ \\
\bottomrule
\end{tabular}
\label{tab:complexity_comparison}
\end{table}

\subsection{Linear Transformers are RNNs:       
 Attention  Through the Lens of Kernels}
\label{sec:linear}

Reaching the ambitious goal of reducing the cost of autoregressive inference from $\mathcal{O}(\dseq_t)$ to $\mathcal{O}(1)$ while attempting to retain the performances of vanilla Transformers and still enabling parallel computations (with crucial impact in reducing training times on large datasets) is extremely challenging. In Figure~\ref{fig:main}, the three aforementioned properties are depicted at the bottom, with blue, pink, and yellow boxes, respectively. There exist multiple categories of models that were recently proposed, sharing the intuition of introducing linearity in the computations (notice that this term is used in a two-fold manner: the attention process does not exploit non-linear functions, and  the goal is to {gain} linear complexity), sometimes going back to the theory of kernels, to recover stateful models of attention (Figure~\ref{fig:Sec3-guide}, first column).

\begin{figure*}[ht]
    \centering
    \includegraphics[width=0.35\textwidth]{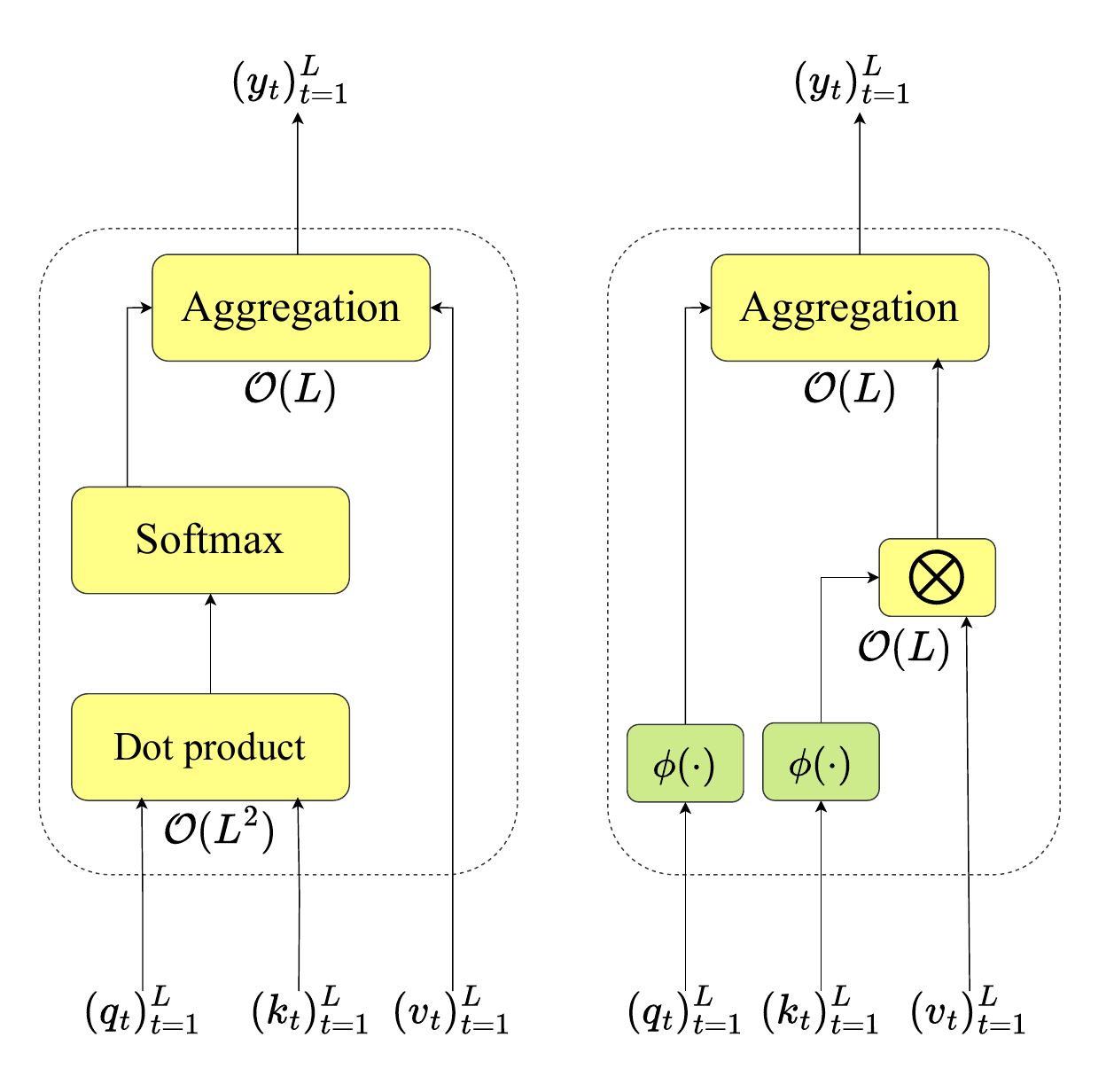}
    \hskip 1cm
    \includegraphics[width=0.42\textwidth]{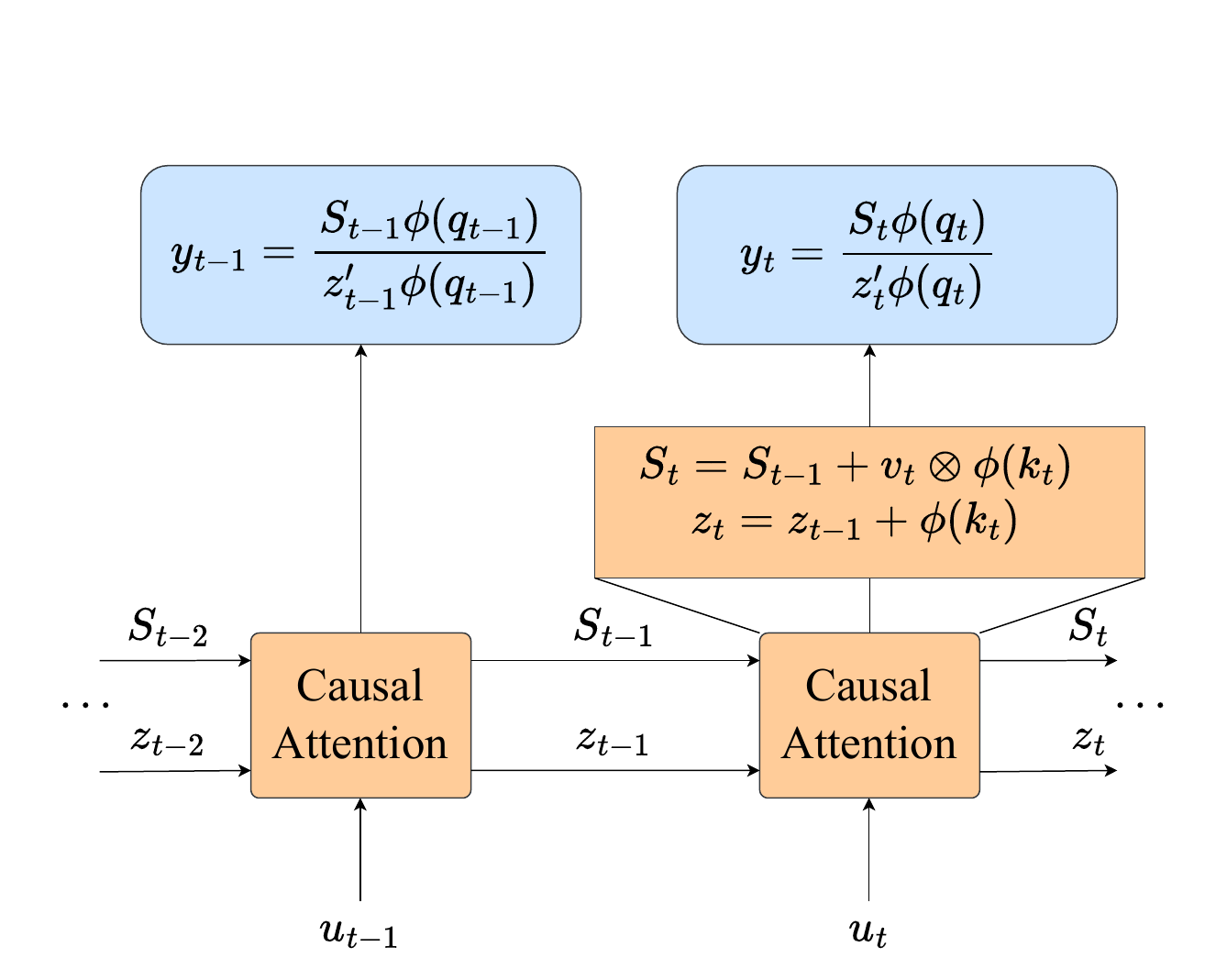}    
    \caption{Left: sketch which highlights the main operations and their complexity in classic softmax attention (left) and in kernel-based linear attention methods (right). Right: how linear attention evolves over time, showing its stateful nature.}    
    \label{fig:kernel_based}
\end{figure*}

\parafango{Linear Attention} The so-called Linear Transformer \cite{katharopoulos2020transformers} achieves a linear complexity on the self-attention operation introducing a similarity function $\similarity\cdot$ in Eq.~\ref{eq:general-attn} described by a kernel function $\mathcal{K}$ \cite{Hofmann_kernel,tsai2019transformer}, $\text{sim}(q, k) := \mathcal{K}(q, k) =  \phi(q)' \phi(k)$, where $\phi\colon \R^\dk \mapsto \R^\dkern$ is a non-linear feature map and the kernel codomain should be positive in order to define proper attention weights, i.e., $\mathcal{K}\colon \R^\dk \times \R^\dk \mapsto \R^+$. Therefore, leveraging the associative property of matrix multiplication, Eq.~\ref{eq:general-attn} can be rewritten as 
\begin{equation}
\label{eq:linearatt}
\begin{aligned}
 S_\dseq &= \sum_{j=1}^\dseq   v_j \otimes \phi(k_j),  \\
z_\dseq &= \sum_{j=1}^\dseq \phi(k_j),\\
y_t&= \sum_{i=1}^\dseq \frac{v_i \phi(k_i)' \phi(q_t)}{\sum_{j=1}^\dseq \phi(k_j)'\phi(q_t)} = \frac{S_\dseq \phi(q_t)}{z_\dseq' \phi(q_t)},
\end{aligned}
\end{equation}

where $\otimes$ denotes the outer product between vectors, and $S_L$ is a matrix. 
This is what is commonly referred to as \textit{cross-attention} or \textit{encoder self-attention} in the Transformers literature, where the whole sequence is available in advance and computing $y_t$ requires the aggregated term $S_L$. However, the two terms $S_L \in \R^{\dk \times   \dkern}$ and $z_L \in \R^\dkern$ are computed once and reused for every query position $t \in \{1, \dots , \dseq\}$. Hence, Linear Transformers reduce the  space and time complexity requirements to $\mathcal{O}(\dseq)$, as depicted in Figure~\ref{fig:kernel_based} (left).
This is evident when we express the numerator of the previous equation exploiting the  \textit{parallel form}, i.e. $\phi(Q)S_L \defeq  \phi(Q)\big(V'\phi(K)\big)$, 
where the feature map $\phi$ is applied rowwise to the matrices $Q$ and $K$. 
If, without any loss of generality, we consider keys, queries, and values of size $\dk$ and a cost to compute $\phi$ of $\mathcal{O}(\dkern)$, then the overall run-time complexity of Linear Transformer is $\mathcal{O}{(\dseq \dkern \dk)}$.
The role of the kernel function $\phi$ has been investigated in several subsequent works \cite{peng2021random,schlag2021linear,choromanski2020rethinking}. Linear Transformer \cite{katharopoulos2020transformers} select $\phi(x) = \text{elu}(x) + 1$, where $\text{elu}(\cdot)$ denotes the exponential linear unit~\cite{clevert2015fast}, a kernel that preserves the dimension of the input key vector $(\dkern = \dk)$, leading to a globally linear complexity of the model with respect to the sequence length, i.e., $\mathcal{O}(\dseq \dk^2)$, implying less computation (in terms of number of operations) in long sequences. 

\parafango{Causal Linear Attention} Interesting properties emerge from the kernel-based formulation when moving to the \textit{autoregressive} setting. As previously anticipated, this intrinsically requires to implement a form of \textit{causal attention}, since future information cannot be accessed. At a generic time $t$, we need to have access to $S_t$ and $z_t$ (i.e., they are aggregated up to what has been seen so far), and Linear Transformers can be seen as stateful models that, at each time step, update an internal state $X_t \defeq (S_t, z_t)$, composed by a recurrent state matrix $S_t$ and a recurrent normalization vector $z_t$, which are updated iteratively with what is referred to as \textit{additive} interaction,

\begin{eqnarray}
\label{eq:rec_transf_s}
S_t &=& S_{t-1} + v_t \otimes \phi(k_t),\\
\label{eq:rec_transf_z}
z_t &=& z_{t-1} + \phi(k_t), \\
y_t &=& \frac{S_t\phi(q_t)}{z_t'\phi(q_t)},
\label{eq:rec_transf_out}
\end{eqnarray}
with some initial condition $S_0$, $z_0$ (usually set to zero). 
Hence, in an autoregressive setting, $y_t$ can be computed in a recurrent fashion, multiplying the current query with the $S_t$ portion of the state, and normalizing it with its $z_t$ portion. This view challenges the traditional distinction between \sigla{RNNs} as ``automata-like'' constant-size stateful models and Transformers as not being so. Indeed, autoregressive Transformers can be equivalently expressed as an RNN-like sequence processor with $2$-dimensional constant-size states that are updated by sum, as depicted in Figure~\ref{fig:kernel_based} (right). Noticeably, this \textit{recurrent} form allows to compress history into the matrix-valued hidden state, thus eliminating the need to maintain and attend over the \sigla{KV cache} during inference.

We remark that, when considering the \textit{parallel} form of causal linear attention, its complexity is still quadratic in $\dseq$. Indeed, due to the presence of the causal masking matrix $M$, it is not possible to exploit
the associative property of matrix multiplication to reduce
the parallel form complexity from quadratic to linear. 
Noticeably, Linear Transformers combine the best of standard attention and recurrence: at  training time (i.e., in non-causal tasks where the full sequence is available in advance), computations can be parallelized and take full advantage of GPUs or other accelerators by leveraging the \textit{parallel form}. When it comes to inference (e.g., sequential decoding), the cost per time
and memory for one prediction is constant, and the internal state $X_t = (S_t, z_t)$ can be updated at every time step like in \sigla{RNNs}, thanks to their \textit{recurrent form}.
Tasks where vanilla  quadratic-time softmax attention
cannot be fully parallelized across time steps can be effectively accelerated, such as in autoregressive decoding, both in the conditional (e.g., machine translation) and unconditional (e.g., sampling from a language model) cases \cite{peng2021random}. 
Additionally, it has been proposed to reduce memory transfer costs by avoiding materializing states in GPU slow \sigla{HBM} memory (see Section \ref{sec:flopsvsmem} for further details).

\parafango{Random Feature Maps} Despite the effective computational advantages, Linear Transformers underperform compared to vanilla self-attention, and the main cause has been attributed to the choice of the kernel function \cite{peng2021random,schlag2021linear,LIN2022111}. 
More recent works \cite{choromanski2020masked,choromanski2020rethinking,peng2021random,schlag2021linear} leverage random feature maps to achieve unbiased estimations of shift-invariant kernels, such as the Gaussian one, in order to approximate effects obtained by using the softmax function in vanilla Transformers. \sigla{Performer}~\cite{choromanski2020masked,choromanski2020rethinking} uses random feature maps defined by custom functions $f_1,\cdots,f_l\colon \R\rightarrow \R$ and $h\colon\R^\din\rightarrow \R$. Formally, given an input vector $k_t \in \R^\dk$ (it could be a key or a query),
\begin{equation}
\begin{aligned}
    \nonumber \phi(x) = \frac{h(x)}{\sqrt{m}}[f_i(\omega_1' x),\ldots,f_i(\omega_m' x)]_{i=1}^{l},
\end{aligned}
\end{equation}
where the notation $[a_i]_{i=1}^{l}$ indicates a vector obtained by concatenating all the $l$ different $a_i$'s,  $\omega_1,\cdots,\omega_m$ are random vectors independently drawn from a distribution $\mathcal D\in\mathcal P(\R^\din)$, i.e., $\omega_1,\cdots,\omega_m\stackrel{\text{iid}}{\sim} \mathcal D$.
Choromanski et al.~\cite{choromanski2020rethinking} tested trigonometric functions with $h(x)=\exp({\|x\|^2/2}), l=2, f_1=\sin, f_2=\cos$, inspired by the random Fourier feature map~\cite{rahimi2007random},  that has been proved to be successful in speeding up kernel methods \cite{oliva2015fast} and to approximate softmax \cite{rawat2019sampled}. Given that relying on trigonometric functions does not guarantee non-negative attention scores and thus could lead to unstable behaviors, follow-up works~\cite{choromanski2020rethinking} proposed positive random feature maps, such as those based on $h(x)=\exp(-\|x\|^2/2),l=1,f_1=\exp$, guaranteeing unbiased and non-negative approximation of dot-product attention.

A concurrent work, Random Feature Attention (\sigla{RFA}) \cite{peng2021random}, leverages similar intuitions, building a feature map with $h(x)$ set to $1$, with queries and keys $\ell_2$-normalized in advance (see \cite{peng2021random} for more details), which shows benefits with respect to the kernels based on the $\text{elu}$. Additionally, \sigla{RFA} includes a learnable gating mechanism inspired by \sigla{LSTMs} \cite{hochreiter1997long} to explicitly model the notion of \textit{recency bias} and locality, which are not considered in vanilla self-attention. Eq.~\ref{eq:rec_transf_s} and \ref{eq:rec_transf_z} are augmented by means of a gating function returning a score $g_t$ in $(0,1)$, becoming

\begin{equation}
\nonumber 
\begin{aligned}\label{eq:gate_rfa}
    g_t &= \sigma(w_g' u_t), \\
    S_t &= g_t\, S_{t-1} +  (1-g_t)\,v_t \otimes {\phi}\left(k_t\right), \\
    z_t &= g_t\, z_{t-1} + (1-g_t)\,{\phi}\left(k_t\right),
\end{aligned}
\end{equation}
where $w_g$ is a vector of learnable parameters (there could also a bias term), and $\sigma$ is a sigmoid activation. The multiplicative interaction between the learned scalar gates $0< g_t<1$ and the hidden state $X_t = (S_t, z_t)$ exponentially decays past information, favoring more recent contexts.
In addition to using random feature maps to approximate standard dot product attention, \cite{peng2021random} and \cite{choromanski2020rethinking} explore alternative routes, approximating order-1 arc-cosine kernel. In this case with $h(x)=1,l=1,f_1=\mathrm{ReLU}$. This feature map has been show to be effective in various tasks including machine translation and protein sequence modeling.

\parafango{Fast Weight Programmers and Delta Nets} Schlag et al. \cite{schlag2021linear} showed the formal equivalence between kernel-based Linear Transformers and the seminal works on {F}ast {W}eight {P}rogrammers (\sigla{FWPs}) \cite{schmidhuber1992learning,schmidhuber2021fwp}. It turns out that the kernel-based causal perspective of linear attention (Eqs.~\ref{eq:rec_transf_s}-\ref{eq:rec_transf_out} and Figure~\ref{fig:kernel_based}-right) is a \sigla{FWP} with the additional recurrent normalization factor $z_t$. The intuition behind the notion of \textit{fast weights} is to introduce novel dependencies on the weights of the model. A two-network system is proposed, composed of a slow net with slowly-changing weights, which continually reprograms another net with weights that change in a fast manner, making them dependent on the spatio-temporal context of a given input stream. Notably, this finding suggests that the recurrent state matrix $S_t$ ($\approx W^{(t)}$ in \sigla{FWP}) can be seen as a key-value associative ``memory matrix'' which gets reprogrammed. ($i$) The ``write'' operation is obtained by aggregating the results of the outer products of ($\phi$-mapped) keys and values in Eq.~\ref{eq:rec_transf_s}, also referred to as \textit{associations}. ($ii$) The ``retrieve'' operation consists of multiplying the memory matrix by the ($\phi$-mapped) query (see Eq.~\ref{eq:rec_transf_out}). 
Schlag et al. \cite{schlag2021linear} argue that endlessly adding new associations to a memory of finite size (Eq. \eqref{eq:rec_transf_s}) will inevitably reach a capacity limit. To prevent associations from interfering with each other upon retrieval, keys must be orthogonal, i.e., the feature map size must be large enough to avoid working in an overcapacity regime.\footnote{With keys embedded in a $\dkern$-space, there cannot be more than $\dkern$ orthogonal vectors. That is, storing more than $\dkern$ associations will result in retrieval issues. Linear attention based on the $\text{elu}$ function preserves the dimension of the input key vector ($\dkern = \dk$) without modifying the memory capacity, thus, when $\dseq > \dkern$, the model might be in such an overcapacity regime.}
This analysis showcases the limits and sub-optimality of random feature maps, characterized by a $2m$-sized capacity which would require $m$ to go to infinity to yield robust approximations. Interestingly, Schlag et al. introduce a deterministic parameter-free feature map which fosters orthogonality in feature space. Specifically, the feature map they propose is such that $\dkern=2\nu \dk$, and it is defined as

\begin{equation}
\nonumber
\begin{aligned}
    \phi(x)&=\left[\mathrm{ReLU}([x,-x])_i \mathrm{ReLU}([x,-x])_{i+\nu}\right]_{i=1}^{2\dk},
    \end{aligned}
\end{equation}
with a capacity controlling parameter $\nu \in \{ 1, \dots, 2\dk-1\}$.  
Similar intuitions led to the enlargement of the associative memory capacity in a write-and-remove fashion. Differently from \sigla{RFA}, associations in the memory are updated while keeping intact other unrelated ones. Specifically, given a new input key-value pair $(k_t, v_t)$, the model first attempts a retrieve operation using $k_t$, in order to get back an estimated value vector $\bar{v}_t$ accordingly to the currently available memory $S_{t-1}$, i.e., $\bar{v}_t = S_{t-1} \phi(k_t)$. It then writes to the memory matrix the difference between the real $v_t$ and the estimated $\bar{v}_t$, modulated by a gating function $g_t$. This update mechanism is an error-correcting delta rule \cite{widrow1960adaptive} with a dynamic learning rate $g_t$,
\begin{equation}
\begin{aligned}\label{eq:updaterule2}
g_t &= \sigma(w_g'u_t), \\
S_t &= S_{t-1} + g_t(v_t - S_{t-1} \phi(k_t)) \otimes \phi(k_{t}),  \\
y_t &= S_{t}\phi(q_t),
\end{aligned}
\end{equation}
hence the model is referred to as \sigla{Delta Network}.
Notice that the recurrent normalization factor $z_t$ leveraged in Eq. \eqref{eq:rec_transf_out} is not there anymore, since the authors propose to apply a sum-normalization before updating the memory matrix (i.e., they normalize $\phi(q_t),\phi(k_t)$ by the sum of their components). 

\parafango{Beyond Delta Nets} Follow-up works extended \sigla{Delta Networks} by adding  recurrent connections that feedback the previous output $y_{t-1}$ (actually, $\tanh(y_{t-1})$), resulting in \sigla{Recurrent Delta Networks} \cite{irie2021going}.
Such recurrent connections are exploited when computing the current key $k_t$, query $q_t$, value $v_t$, and $g_t$ of \sigla{Delta Networks}, i.e.,
\begin{equation}
\nonumber
    \begin{aligned}
    k_t &= W_k u_t + R_k \tanh(y_{t-1}), \quad \text{(same for $q_t$ and $v_t$)}\\
    g_t &= w_g' u_t + r_q' \tanh(y_{t-1}),\\
    y_t &= S_t \phi(q_t) + R_y \tanh(y_{t-1}),
    \end{aligned}
\end{equation}
where the $R_{\cdot}$'s and $r_q$ are new projection matrices and vector, respectively.
Inspired by Self-Referential Weight Matrix (\sigla{SRWM}) \cite{schmidhuber1993self}, Irie et al. \cite{irie2022modern} proposed a different approach, which can be considered an extension of previous works on \sigla{FWPs}. Compared to \sigla{Delta Networks} of Eq.~\ref{eq:updaterule2}, \sigla{SRWM} is based on a computational scheme in which the output $y_t$ is directly computed by projecting the ($\phi$-mapped) input, $y_t = W_y \phi(u_t)$, being $W_y$ a learnable projection matrix. Such a new projection matrix and the already introduced $W_q$, $W_k$, $w_g$, are the outcome of a programming scheme which consists in a recurrent process, i.e., the one that in Eq.~\ref{eq:updaterule2} is used to update $S$. Once we replace $S$ by another matrix $\tilde{S} = [W_y, W_q, W_k, w_g]$, and by appropriately choosing the number of components in $v_t$, \sigla{SRWM} computes the value vector as a function of $\tilde{S}$,
\begin{equation}
\nonumber
\begin{aligned}
[y_t, q_t, k_t, g_t] &= \tilde{S}_t \phi(u_t), \\
v_t &= \tilde{S}_t \phi(q_t),\\
\tilde{S}_t &= \tilde{S}_{t-1} + g_t(v_t - \tilde{S}_{t-1} \phi(k_t)) \otimes \phi(k_{t}), 
\end{aligned}
\end{equation}
where the first equation summarizes the linear projections of the ($\phi$-mapped) input by $W_y, W_q, W_k, w_g$ in a compact manner.
A very recent work \cite{irie2023practical} investigates the computational power of Transformers with linear attention \cite{katharopoulos2020transformers} and \sigla{Delta Networks} \cite{schlag2021linear}, showing that the just introduced recurrent \cite{irie2021going} and self-referential extensions of the latter \cite{irie2022modern} provides improvements over Linear Transfomers, e.g., allowing for generalisation on the parity problem. 
Mao \cite{mao2022fine} proposes a data-dependent decay-based update, exploiting gated rules \cite{peng2021random}. A low-rank decay matrix $G_t$ with values in $(0,1)$ is factorized by two gating vector-functions, and it is used to decay the state $S_t$,
\begin{equation}
    \begin{aligned}
    G_t &= \sigma(W_f u_t)\sigma(W_z u_t)', \\
    S_t &= G_t \odot S_{t-1} + v_t \otimes k_t,
    \end{aligned}
    \label{eq:palle}
\end{equation}
where $\odot$ is the element-wise product and $W_z \in \R^{\din \times \din}$, $W_f \in \R^{\dk \times \din}$ are newly added learnable parameters.
It clearly differs from the delta rule in \sigla{Delta Networks} \cite{schlag2021linear}, especially due to the introduction of a  finer-grained element-wise operation. Moreover, there is no $\phi$ function in Eq.~\eqref{eq:palle} since this model virtually exploits a linear feature map $\phi(x) = W_\phi x$, that can be subsumed into the query-key projections $W_k, W_q$,  inspired by Kasai et al. \cite{kasai2021finetuning}.
{The hidden states $S \in \R^{\dseq \times \dk \times \dv}$ (which are required for gradient computation during the backward pass) must be stored due to the addition of $G_t$
in Eq. \eqref{eq:palle}, which makes the computation
non-reversible \cite{mackay2018reversible} and the I/O memory management challenging.} 

\parafango{Other Kernel-based Linear Methods} 
There exist other works that we can connect to the principles behind linear attention and recurrence.
\sigla{Ecoformer}~\cite{liu2022ecoformer} exploits an energy-saving kernel-based hashing  (RBF kernel) to map the queries and keys onto low-dimensional binary codes in Hamming space. Kernelized hash functions are learned in a self-supervised manner, driven by the Hamming affinity of the attention scores. 
\sigla{CosFormer} \cite{QinSDLWLYKZ22} uses $\phi(x) = \text{ReLU}(x)$ ensuring {non-negativity} in the attention weights, and a cosine-based re-weighting mechanism to enforce the locality bias in the original softmax attention. 
The authors of \sigla{TransNormer} \cite{qin2022devil} identify issues in kernel-based Linear Transformer, both due to {\it unbounded gradients} and {to} {\it attention dilution}, i.e., the sparse distribution of attention scores in long sequences. They propose to solve them by normalizing the attention computation and leveraging diagonal attention confined to neighbouring tokens in early layers. In particular, \sigla{TransNormer} leverages a linear kernel, replacing the role of the $z_t$ normalization factor of Eq. \eqref{eq:rec_transf_out} by Layer Normalization \cite{ba2016layer} on the attention scores (NormAttention). This results in the most compact form of the update equations of Linear Transformers,

\begin{equation}
\begin{aligned}
 S_t &= S_{t-1} + v_t \otimes  k_t, \\
 y_t &= S_t q_t,
 \end{aligned}
\label{eq:simple_linear_attention}
\end{equation}
from which it is even easier to trace connections to linear \sigla{RNNs} equipped with ``matrix-valued'' hidden state $S_t$, updated by the outer-product $v_t \otimes  k_t$, as already pointed out by several works \cite{schmidhuber1992learning,schlag2021linear,irie2021going,mao2022fine}.
We mention that the idea of proposing an \sigla{RNN}-oriented view of matrix-valued state Transformers has been also recently remarked by Oren et al. \cite{oren2024transformers}, introducing \sigla{Multi-state RNNs}. In such a view, the state can be considered the number of input tokens in the sequence, which is basically related to the time-increasing \sigla{KV-cache}. 

To sum up, all the papers described up to this point highlight the rich and convenient connections between Transformer models and \sigla{RNNs}, when considering specific architectural designs. In this respect, linear Transformers can be considered as linear \sigla{RNNs} acting in a matrix-valued state space, where states are matrices updated via the outer-product $v_t \otimes k_t$. 

\subsection{Alternative Low-rank Approximations}
\label{sec:alt}

Kernel-based methods exploited to linearize Transformer self-attention are effective in reducing the computational cost with respect to the sequence length. However they result in a computational complexity quadratic to the model{’s feature dimension, } $\mathcal{O}(\dseq \dk^2)$, caused by their reliance on matrix dot products, which is inefficient forlarge model sizes \cite{zhai2021attention}.

\parafango{Attention-free Transformers} A recent alternative emerged from a different low-rank factorization of the $\similarity{\cdot,\cdot}$ function in Eq. \eqref{eq:general-attn}, leveraging  intuitions which are related to Linear Attention of Eq. \eqref{eq:linearatt}, even if based on element-wise multiplications that preserve the feature-dimension, i.e., $\similarity{q, k}=\sigma(q)\odot \psi(k)$. 
{A}ttention {F}ree {T}ransformers  (\sigla{AFT}) \cite{zhai2021attention} implement $\sigma(\cdot)$ as an elementwise nonlinear activation function (e.g., sigmoid) and $\psi(k) = e^{k}$, and perform the following operation,
\begin{equation}
y_t =  \sigma(q_t) \odot \frac{ \sum_{i=1}^{\dseq} e^{k_i + w_{t,i}} \odot  v_i}{\sum_{i=1}^{\dseq} e^{k_i+w_{t,i}}}, 
\label{eq:aft}
\end{equation}

where the division is intended to operate element-wise, while $w_{t,i}$ denotes the $(t,i)$-th element of matrix $\mathcal{W} \in \R^{\dseq \times \dseq}$, which is a learnable matrix of pairwise position biases. In fact, for each input token at position $t$, \sigla{AFT} computes a weighted average of values, the result
of which is combined with the query by an element-wise multiplication.
This approach eliminates the need for dot product self-attention, while still preserving global interactions between query and values , and avoids the need to compute and store the attention matrix. 
As noted by the authors, \sigla{AFT} triggers an implicit attention {mechanism with as many heads as the feature dimensions, wherein the attention matrices have a factorized structure.
Such a procedure has a memory complexity that is linear with respect to both the context size and the number of features, making it well-suited for both large inputs and large model sizes, and is generally referred to as \textit{channel directed attention}.
The special configuration in which no position biases are learned, referred to as \sigla{AFT-simple}, can be trivially formalized as

\begin{equation}
\begin{aligned}
y_t &=  \sigma(q_t) \odot \frac{ \sum_{i=1}^{\dseq} e^{k_i} \odot  v_i}{\sum_{i=1}^{\dseq} e^{k_i}}. 
\label{eq:aft}
\end{aligned}
\end{equation}
Masked/causal formulation can be obtained by limiting the upper range of the summations to the current time index $t$ (instead of $L$).
It is easy to see that \sigla{AFT-simple} completely gets rid of dot products (the ``softmax'' is independent on the output index $t$), which results in a complexity of $\mathcal{O}(\dseq \dk)$ rather than $\mathcal{O}(\dseq \dk^2)$.

\parafango{Reinventing RNNs for the Transformer Era} Whilst both \sigla{AFT} and Linear Transformers largely reduce the computational requirements of classic softmax self-attention, they do not match its performances in real-world tasks, such as language modeling at scale. Inspired by \sigla{AFT}, the Receptance Weighted {K}ey {V}alue (\sigla{RWKV}) model \cite{peng2023rwkv} was proposed, driven by the intuition of giving less importance to ``older'' embeddings, thus
replacing \sigla{AFT} pair-wise position embeddings $w_{t,i}$'s with exponential decays. 
In detail, in \sigla{RWKV} each $w_{t,i}$ is implemented with a decay factor, 
i.e., $w_{t,i} \defeq - (t-i)w$ where $w \in \R^\dk_{\geq 0}$.
Then, the \sigla{RWKV} architecture is the outcome of stacking pairs of blocks, where each pair consists of a of \textit{time-mixing} and a \textit{channel-mixing} residual layers, respectively.
\begin{figure}
    \centering
    \includegraphics[width=\columnwidth]{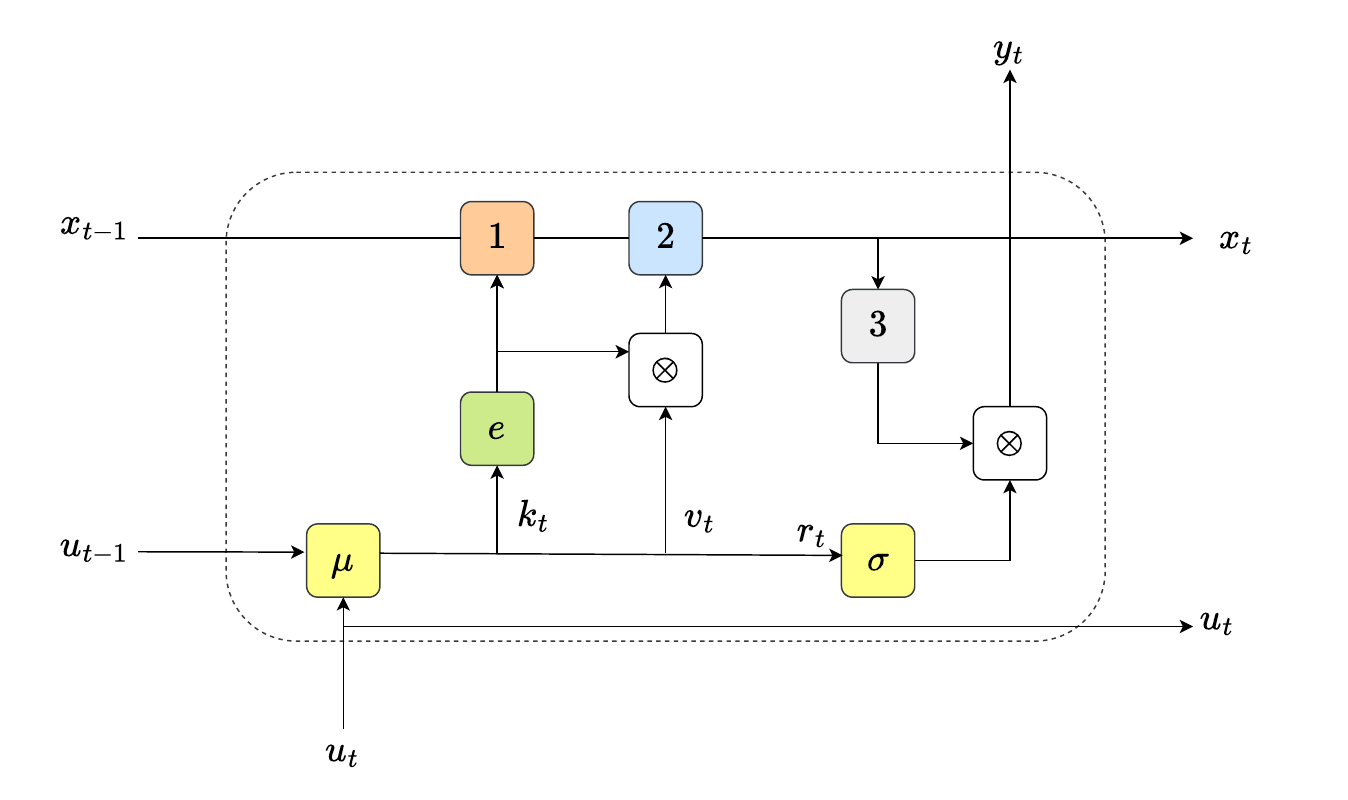}
    \caption{RWKV time-mixing block \cite{peng2023rwkv}. The yellow block ($\mu$) implements the token shift; referring to Eq.~\ref{eq:xyz1}, the orange block (1) denotes the computation of $s_t$ and the blue block (2) denotes the computation of $z_t$ (recall that $x_t$ is the tuple $(s_t, z_t)$). The gray block (3) evaluates the fraction in the computation of $y_t$. See \cite{peng2024eagle} for the recurrent form  of later models (\sigla{RWKV-5/6}).}
    \label{fig:rwkv}
\end{figure}
The {\it time-mixing} layer is based on the computation of the so-called \textit{receptance} $r_t$, of the key vector $k_t$ and of value $v_t$. Then, the output of the block, $y_t$, depends on all such elements, together with the decayed position embeddings,
\begin{eqnarray}
        \hskip -0.6cm r_t &\hskip -0.3cm = \hskip -0.3cm& W_r (\mu_r\odot u_t + (1-\mu_r)\odot u_{t-1}), \label{eq:rwkv1} \\
\hskip -0.6cm k_t &\hskip -0.3cm = \hskip -0.3cm& W_k (\mu_k \odot  u_t + (1-\mu_k)\odot u_{t-1}), \label{eq:rwkv2} \\
\hskip -0.6cm v_t &\hskip -0.3cm = \hskip -0.3cm& W_v (\mu_v \odot  u_t + (1-\mu_v)\odot u_{t-1}),\label{eq:rwkv3} \\
\hskip -0.6cm y_t &\hskip -0.3cm = \hskip -0.3cm& W_o \sigma(r_t)\odot  \frac{ \sum_{i=1}^{t-1} e^{-(t-i- 1)w+k_{i}}\odot v_i + e^{p+k_t}\odot v_t}{\sum_{i=1}^{t-1} e^{-(t-i- 1)w+k_{i}} + e^{p+k_t} }, 
   \label{eq:rwkv4}
\end{eqnarray}
where $\mu_.$, $W_.$, and $p$ are additional trainable parameters. 
In Eq.~\eqref{eq:rwkv1}-\eqref{eq:rwkv3}, recurrence is  implemented by means of a linear interpolation  between  input at $t$ and $t-1$, referred to as \textit{token-shift}.
In Eq.~\eqref{eq:rwkv4}, receptance $r_t$ participates in the computation of a gate that acts as a forget gate, which eliminates unnecessary historical information. 
The recurrent nature of the model goes beyond Eq.~\eqref{eq:rwkv1}-\eqref{eq:rwkv3}, and it also affects Eq.~\eqref{eq:rwkv4}, once we consider autoregressive decoding at inference. In fact Eq.~\eqref{eq:rwkv4} can be written in the following recurrent form,
\begin{equation}
    \begin{aligned}
    s_t &= e^{-w}\odot s_{t-1} + e^{k_t}\odot v_t, \\
    z_t &= e^{-w}\odot z_{t-1} + e^{k_t}, \\
   y_t &= W_y \sigma(r_t) \odot  \frac{s_{t-1} + e^{p+k_t}\odot v_t}{z_{t-1} + e^{p+k_t}}, 
    \end{aligned}
    \label{eq:xyz1}
\end{equation}
where the {\it state} is $x_t \defeq (s_t, z_t)$. The dataflow of the RNN-like time-mixing is shown in Figure \ref{fig:rwkv} (bottom). 
In \sigla{AFT}, $\mathcal{W}$ is a matrix of (pairwise) position biases, while here it represents a channel-wise\footnote{The term channel refers to the feature dimension.} vector multiplied by relative positions. 
Going beyond the just described {\it time-mixing} layer, \sigla{RWKV}, features a {\it channel-mixing} block, that computes its own receptance $r'_t$ and keys $k'_t$ following the same formulation of Eq.~\eqref{eq:rwkv1} and Eq.~\eqref{eq:rwkv2}, respectively (with separate learnable parameters). Then, the channel-mixing block computes its output $y_t$ by  means of $y_t = \sigma(r'_t)\odot (W_v' \cdot \text{ReLU}^2(k'_t))$, leveraging the squared ReLU activation \cite{so2021searching}.\footnote{Implemented as $\text{ReLU}^2(x) \defeq \big(max(0,x)\big)^2$} 
Noticeably, the token-shift operation allows the model to learn the amount of new information that should be stored into each channel of receptance, key, value and gate, resulting in the capability to compose induction heads  within a single layer \cite{elhage2021mathematical}. Indeed, though this formulation, a single head can directly accumulate both past and current token information into separate subspaces. 
The parallel (non-causal)  form  of \sigla{RWKV} has a time complexity of $\mathcal{O}(\dseq\dk^2)$.
Updating attention scores in \sigla{RWKV} requires a serial scan (hence, the model cannot be parallelized along the sequence dimension) and has complexity of $\mathcal{O}(\dseq\dk)$. 
The simplification of dot-product attention to element-wise operations leads to significant efficiency gains, but at the same time it limits the model capability of capturing long-term dependencies. 

\parafango{Evolution of RWKV} In a subsequent work \cite{peng2024eagle}, the authors proposed two evolutions of the model, referred to as \sigla{Eagle} (or \sigla{RWKV-5}) and \sigla{Finch} (or \sigla{RWKV-6}). The former improves upon the previous architecture through the use of expressive multi-headed matrix-valued states (as opposed to vector-valued
 states), a reformulated receptance, and an additional gate $g_t$. 
 For ease of notation, hereinafter the linear interpolation implementing the token-shift operation, Eq.~\eqref{eq:rwkv1}-\eqref{eq:rwkv3}, will be more compactly defined by means of the   $\text{lerp}$ operator, $\text{lerp}(a,b,\mu) \defeq a + (b-a) \odot \mu$, where $\mu$ is a learnable vector. Moreover, we are in a multi-head setting, involving $h$ heads, thus all the vectors belongs to $\R^{\dk/h}$.
The channel-mixing block is exactly the same of the previous model, while the time-mixing block in \sigla{RWKV-5} revises the one in Eq.~\eqref{eq:rwkv1}-\eqref{eq:rwkv4} as follows,
\begin{equation}
\label{eq:rwkv5}
    \begin{aligned}
    r_t &= W_r\text{lerp}(x_t, x_{t-1}, \mu_r),\quad k_t = W_k\text{lerp}(x_t, x_{t-1}, \mu_k), \\
    v_t &= W_v\text{lerp}(x_t, x_{t-1}, \mu_v),\quad g_t = W_g\text{lerp}(x_t, x_{t-1}, \mu_g), \\
        w &= \text{exp}\big(-\text{exp}(\omega)\big), \\
        \text{WKV}_t &= \text{diag}(u) (v_t \otimes k_t) + \sum_{i=1}^{t-1} \text{diag}(w)^{t-1-i} (v_i \otimes k_i), \\
        y_t &= W_o  \text{concat} \big(\text{SiLU}(g_t) \odot \text{LayerNorm}(\text{WKV}_t r_t)\big),
    \end{aligned}
\end{equation}
where $\text{WKV}_t \in \R^{(\dk/h) \times (\dk/h) }$ is the attention score matrix, the $\text{concat}(\cdot)$ operation concatenates the contributions from different heads, $\text{diag}(\cdot)$ yields a square matrix whose diagonal is the fuction argument, while $\text{LayerNorm}$ (layer normalization) operates on each of $h$ heads separately. Please notice that $w$ is obtained from a double exponentiation where $\omega \in \R^{\dk/h}$  are headwise trainable parameters, thus ensuring that $w \in (0,1)$, and guaranteeing that $\text{diag}(w)$ is a contraction matrix.  
It turns out that the attention score $\text{WKV}_t $ in the \sigla{Eagle} model can be expressed in the  recurrent form,
\begin{equation}
    \begin{aligned}
    S_{t}& = \text{diag}(w)  S_{t-1} + v_t \otimes k_t,  \\ 
    \text{WKV}_t &= S_{t} + \text{diag}(u)  (v_t \otimes k_t), \label{eq:yaaaa}
    \end{aligned}
\end{equation}
confirming the recurrent nature of this model as well.
From such a recurrent form, it is easy to notice that the state $S_t$ is a sum over outer products where each channel is individually decayed by the corresponding channel of $w$, at each time
 step. The attention score of Eq.~\eqref{eq:yaaaa} (bottom) is computed by applying a per-channel boost $u$, multiplied with the current token's outer product $v_t \otimes k_t$, giving the current token a special treatment relative to the sum of past tokens contained within the decaying state history. 
The \sigla{Finch/RWKV-6} model further improves the architecture expressivity by injecting data-dependence for both the time-mixing and token-shift modules. Additionally, it proposes to leverage Low Rank Adaptation (\sigla{LoRa}) \cite{hu2021lora}  to effectively exploit the learned data decay vectors in a context-specific manner.
Indeed, the token shift in \sigla{Finch} leverages an augmented data-dependent linear interpolation $\text{ddlerp}()$ implemented as follows,
\begin{equation*}
    \begin{aligned}
    \text{LoRa}(x, A, B, \lambda) &= \lambda + B\text{tanh}(Ax), \\ 
    \text{ddlerp}(a,b, A, B, \lambda) &= a + (b-a)\odot \\
    & \hskip 0.5cm \text{LoRa}(a+ (b-a)\odot \mu_x, A, B, \lambda),
    \end{aligned}
\end{equation*}
where $B \in \R^{\din \times 32}$, $A \in \R^{32 \times \din}$, and $\mu_x, \lambda \in \R^{\din}$ are trainable parameters. In this novel form of data-dependent token-shift the amount of new and old data allocated per channel now depends on the input at both current and prior time steps.  
The time-mixing block extends Eq.~\eqref{eq:rwkv5} as follows (replacing the equation of $w$ and of $\text{WKV}_t$),
\begin{equation*}
    \begin{aligned}
    d_t &= \text{LoRa}(\text{ddlerp}(x_{t}, x_{t-1}, A, B, \lambda), A, B, \lambda), \\
        w_t &= \text{exp}\big(-\text{exp}(d_t) \big), \\
        \text{WKV}_t &= \text{diag}(u) (v_t \otimes k_t) + \sum_{i=1}^{t-1} \text{diag}(\odot_{j=1}^{i-1} w_j) (v_i \otimes k_i).
    \end{aligned}
\end{equation*}
Differently from  \sigla{Eagle/RWKV-5} where $w$ is a fixed vector, in  \sigla{Finch/RWKV-6} each channel of $w_t$ varies over time in a data-dependent manner. The $\text{LoRa}$ operators allows to inexpensively augment learned vectors with additional offsets determined by the incoming input.

\subsection{Modeling Recurrence}
\label{sec:subrec}

Amidst the multiple advantages brought by the vanilla self-attention mechanism, early attempts to tackle language modeling with Transformers were hampered by the inability ($i$) to model intra-sequential relations due to the static order dependencies available in standard positional encodings (i.e., the absence of temporal information), and ($ii$) to propagate inter-sequence information among processed contexts. Several attempts to tackle these two issues are presented in the following. Additionally, such approaches inspired recent sub-quadratic methods that split the overall computation into sequence-chunks that are processed in parallel.


\parafango{Intra-sequence Recurrence Modeling} An interesting line of work leverages recurrent mechanisms to increase the representational power of features extracted by Transformers, e.g., by injecting locality biases or temporal ordering in the obtained outputs. 
Chen et al. \cite{chen2018best} showed that representations learned by RNN-based encoders can be augmented by those learned with a self-attention encoder, resulting in an improvement in performance for RNN-based neural machine translation (NMT) tasks \cite{stahlberg2020neural}. 
Inspired by this work, Hao et al. \cite{hao2019modeling} propose to directly model recurrence in an encoder-decoder Transformer architecture for NMT. This is done by leveraging an additional recurrence encoder, ${E_\text{rec}(\cdot)}$, that operates in parallel with respect to the standard Transformer encoder, hereinafter referred to as $E(\cdot)$. The authors propose to implement ${E_\text{rec}(\cdot)}$ as ($i$) a bidirectional \sigla{RNN} or as ($ii$) an Attentive Recurrent Network (\sigla{ARN}), a model that performs recurrent steps on the features vectors extracted with an attention model from the input representations $U$. For a generic layer $l$, and its input $U^l$,  \sigla{ARN} computes: 
\begin{equation*}
\begin{aligned}
     x_t^l &= F(x_{t-1}^l, c_t^l),    \\
    c_t^l &= \text{ATT}(x_{t-1}^l, U^{l-1}),
\end{aligned}
\end{equation*}
where $F(x_{t-1}^l, c_t^l)$ denotes an \sigla{RNN}-like transition function (e.g., such as Eq. \eqref{eq:RNN} or the one used in \sigla{LSTM}) with state $x_t$, which processes an external input value $c_t$. The operation $\text{ATT}(x_{t-1}^l, U^{l-1})$ is an attention procedure over the layer-input representations $U^{l-1}$, exploiting the previous state $x_{t-1}^l$ as query. 
Outputs from $E(\cdot)$ and $E_\text{rec}(\cdot)$ are fused either by a gating mechanism or by sequentially attending them in the decoder.  
Injecting this kind of recurrent mechanisms into purely attention-based architectures has been proven to boost their performance.  Chen et al. \cite{chen2019recurrent} attribute these results } on the fact that positional embeddings exploited by standard Transformer architectures are based solely on discrete numerical indices (e.g., using trigonometric functions \cite{vaswani2017attention}) and are independent
of word content. Hence, they are not prone to capture semantic  dependencies between words in a sentence. To overcome this issue, the authors in \cite{chen2019recurrent} split the embeddings of each input word/token $u_t$ into two parts, resulting in two input sequences $U^p=(u_t^p)_{t=1}^L$ and $U^r=(u_t^r)_{t=1}^L$. Then, they explicitly learn recurrent positional embeddings (\sigla{RPEs}) on $U_r$ with a non-linear \sigla{RNN}, i.e., \sigla{RPEs} are the elements  of the temporal sequence of states computed by such \sigla{RNN}. At the same time, the other sub-sequence $U^p$ is leveraged to compute positional word embeddings following the standard Transformer pipeline. The concatenation of such embeddings and \sigla{RPE} is then given as input to the Transformer encoder (or decoder), equipped with ad-hoc heads to process the recurrent positional information. Leveraging such recurrent embeddings allows the model  to capture order-related dependencies.

Huang et al. \cite{huang2022encoding} proposed a block-diagonalization of a linear \sigla{RNN} layer that allows to rewrite it into a lightweight relative positional encoding matrix for a multi-head self-attention, named Recurrent Encoding Matrix (\sigla{REM}). The overall intuition is to encapsulate the recurrent dynamics of the \sigla{RNN} layer into the positional encodings of a multi-head attention layer, leading towards a self-attention with recurrence (\sigla{RSA}).  
In particular, by considering a \textit{linear} \sigla{RNN} (obtained from Eq. \eqref{eq:RNN} when $\sigma$ is the identity function), it is possible to write it in the following compact form,  
\begin{equation}
x_t =  Ax_{t-1} + Bu_t =  \sum_{j=0}^{t-1} A^j Bu_{t-j}.
\label{eq:rnn_unroll}
\end{equation}
Thus, the authors propose to block-diagonalize the $A$ matrix such that the \sigla{RNN} can be broken down into a sequence of simpler \sigla{RNNs}.
Under mild assumptions, $A$ has $r$ real non-zero eigenvalues ($\lambda_1, \dots, \lambda_r$) and $s$ pairs of complex nonzero distinct eigenvalues (the pair ($\gamma_k e^{i\theta_k},\gamma_k e^{-i\theta_k})$ with $1 \leq k \leq s$ where $i$ is the imaginary unit). The corresponding real Jordan form is $A=G\Lambda G^{-1}$, where $G \in \R^{\din \times \din}$ is invertible and $\Lambda\in \R^{\din \times \din}$ is a block diagonal matrix. The exponentiation of this matrix is easy to compute, i.e., $A^j=G\Lambda ^jG^{-1}$, and it is possible to break down the recurrence induced by $A$ into that of $p \times p$ block matrices in $\Lambda$, with $p \in (1,2)$.
As a result, the linear \sigla{RNN} layer can be decomposed into three different contributions, namely $(x^R_t, x^{\text{C1}}_t, x^\text{C2}_t)$, the first one corresponding to real eigenvalues (the $1 \times 1$ blocks in $\Lambda$) and the last two to the complex ones (the $2 \times 2$ blocks in $\Lambda$). 
This allows to rewrite Eq. \eqref{eq:rnn_unroll} as $x_t = \sum_{k=1}^r x^R_t(\lambda_k) + \sum_{k=1}^s x^{\text{C1}}_t(\gamma_k, \theta_k) + \sum_{k=1}^s x^{\text{C2}}_t(\gamma_k, \theta_k) + Bu_t$. The first term is defined as follows $x^R_t(\lambda_k) \defeq \sum_{j=1}^{t-1} \lambda^j B^Ru_{t-j}$. See the referenced paper for the complete forms of $x^{\text{C1}}_t$ and $x^{\text{C2}}_t$.
When considering the input matrix $U$, it is interesting to see that the aforementioned decomposition allows to write the \sigla{RNN} as a multihead self-attention with $r+ 2s +1$ heads, with null query and values and where the positional encodings matrices encapsulate the recurrent dynamics. In details,
\begin{equation*}
    (x^*_1, \ldots, x^*_T) =  \big(\text{softmax}(QK') + P^*(\lambda)\big)V,
\end{equation*}
that is differently instantiated for $x^*_t\in (x^R_t, x^\text{C1}_t, x^\text{C2}_t)$. The same holds for $P^*(\lambda_k)$, which is a relative positional encoding (lower triangular in the causal masked case) matrix, referred to as  Recurrence Encoding Matrix (\sigla{REM}). For instance, when considering $x^R_t$, $V=UB^R$ and $P^R(\lambda_k)$ has a specific form (see the referenced paper for the case of $P^{C1}$ and $P^{C2}$).  
These three \sigla{REMs}, $P^R, P^{\text{C1}}, P^{\text{C2}}$  summarize different temporal decays patterns: the regular \sigla{REM}, $P^R$, provides the regular exponential decay induced by the real eigenvalues. Cyclical \sigla{REMs}, ($P^{C1}, P^{C2}$), provide the cyclical damped
cosine or sine decay induced by the pair of complex eigenvalues. \sigla{REM} can be injected into any existing self-attention based Transformer architecture, leading to the Self-Attention with Recurrence (\sigla{RSA}) module,
\begin{equation*}
    \sigla{RSA}(U) = \Big(\big(1 - \sigma(\mu)\big) \text{softmax}(QK') + \sigma(\mu)P^* \Big) V,
\end{equation*}
where $\sigma(\mu) \in [0,1]$ is a gate parametrized by $\mu$,  $\sigma$ the sigmoid function and $P^*$ is a regular or cyclical \sigla{REM}. 
\sigla{RSA} models the token-level recurrence, which is at the most finegrained scale. Subsequently, it can be easily incorporated into the  coarser-grained designs that will be the subject of the next paragraph, and may potentially bring benefits to their performance.
Token Turing machines \cite{ryoo2023token} takes the alternative route of exploiting memory-based mechanisms, inspired by Neural Turing Machines \cite{graves2014neural}. An
external memory, populated with token summarizations, is read/written using a Transformer as the processing unit/controller at each step. Hence, constant compute is achieved, regardless of the length of the history, hence resembling the computational mechanism of recurrent models where the memory is the neural state. 

\begin{figure*}[h]
\vskip 0.0in
\begin{center}
\includegraphics[width=0.55\linewidth]{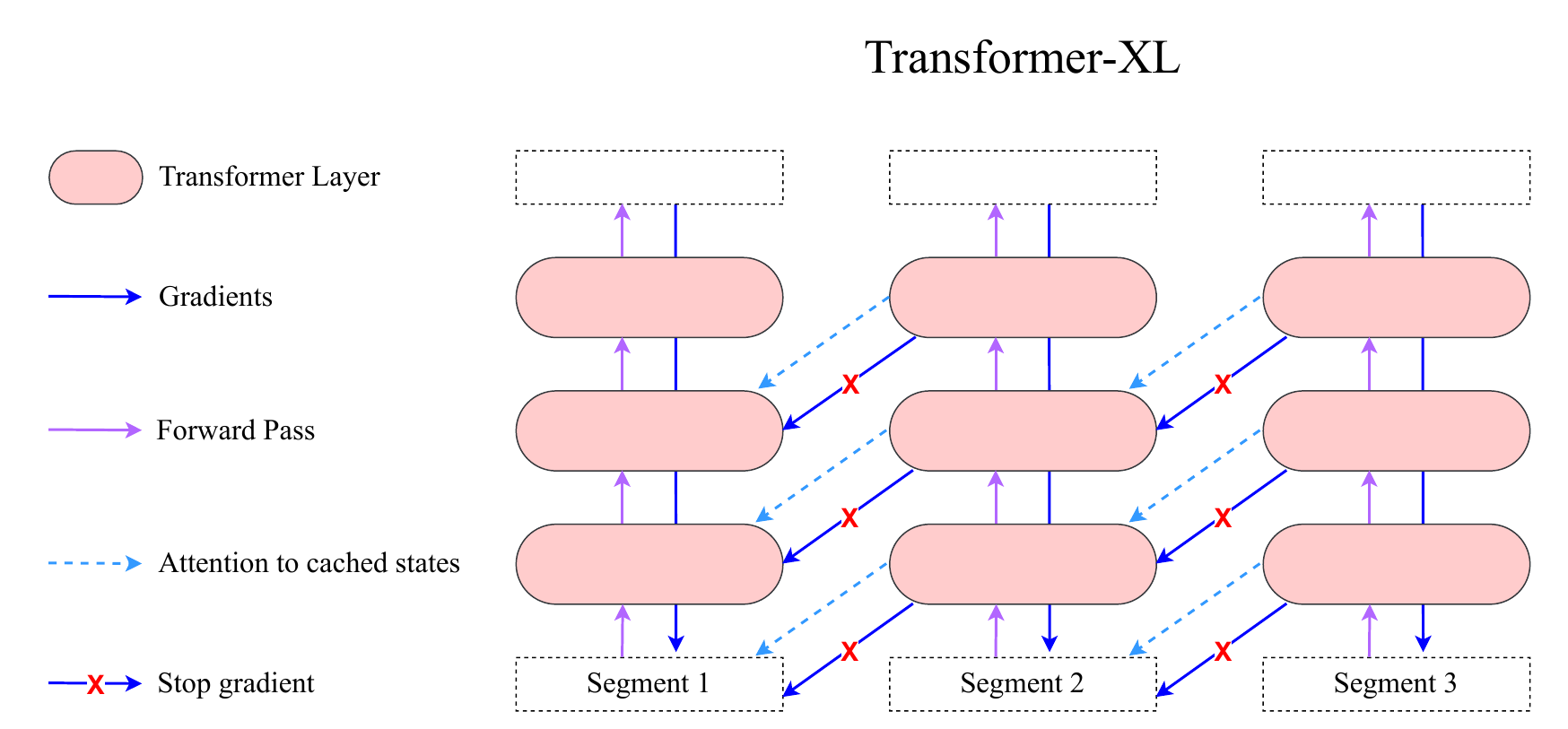}
\hskip 0.2cm
\includegraphics[width=0.39\linewidth]{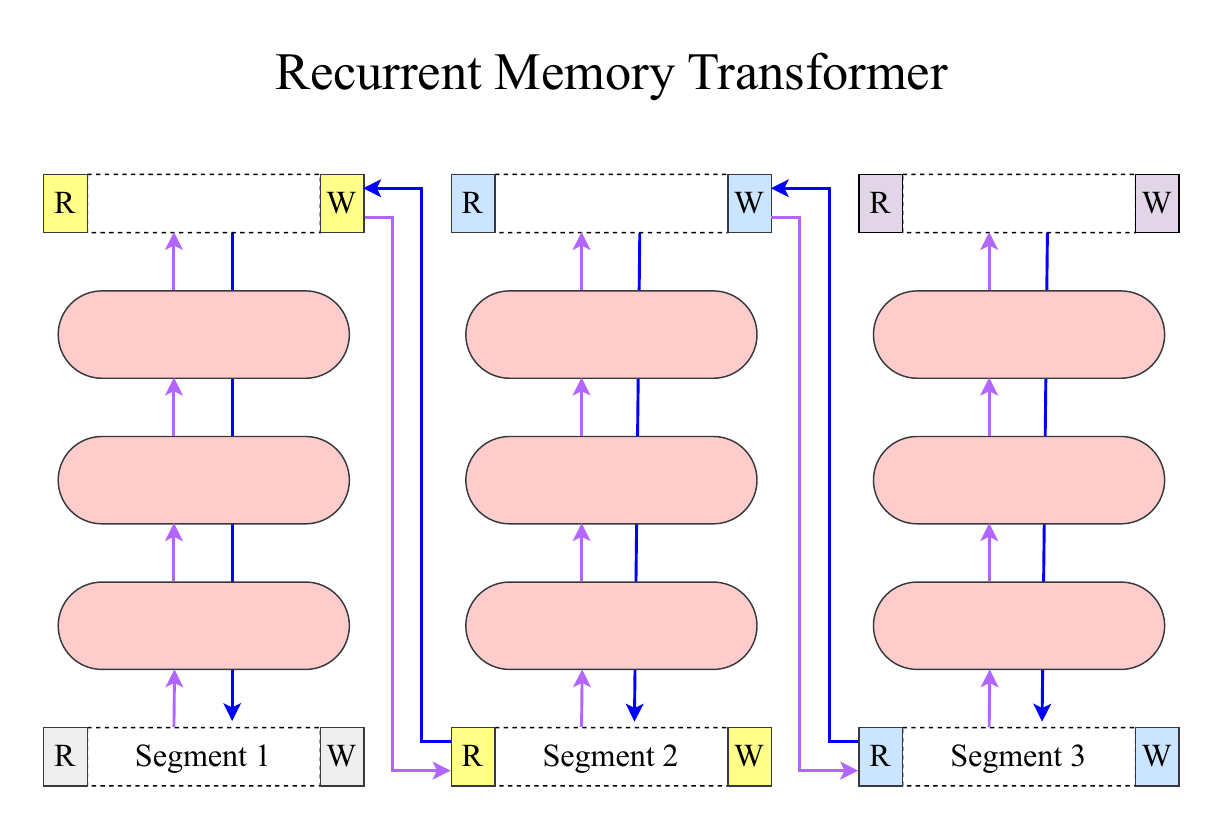}
\caption{\textbf{Comparison of \sigla{Transformer-XL} and Recurrent Memory Transformer (\sigla{RMT})  architectures.} {Recurrent Memory Transformer augments Transformer with global memory tokens, allowing a segment-level recurrence. Special read/write memory tokens (denoted in the figure with $R$ and $W$ for ease of notation -- in the main text they are $M^{\text{read}}$ and $M^{\text{write}}$, respectively) are added to the input sequence. Multiple memory tokens can be used in each read/write block. Updated representations of write memory are passed to the next segment. During training, \sigla{RMT} uses \sigla{BPTT} to propagate gradient to previous segments through memory tokens representation. Effective context length for recurrence with memory is not limited by the depth of a network which is the case for the cache of \sigla{Transformer-XL}.}}
\label{fig:trXL}
\end{center}
\end{figure*}

\parafango{Segment-level Recurrences} Standard Transformer training is performed on separate fixed-length segments of text, derived from the context window span, without any information flow across such segments, resulting in the inability to capture any longer-term dependency beyond the predefined context. Hence, the ability to temporally connect different contexts becomes extremely important.  
The basic feature that can overcome such limitations consists of maintaining a cache memory, to be read as an additional input, in order to store the state of previously processed information. 
This can be easily implemented by exploiting a segment-level recurrence to aggregate the temporal information at a coarser scale, while the token-by-token dependencies are learned by self-attention as usual \cite{LIN2022111}. 
A seminal work in this direction is \sigla{Transformer-XL}\cite{dai2019transformer}, which sequentially processes consecutive segments having size $T$, exploiting a segment-level recurrence. Layer-wise representations from previous segments are cached and exploited as an extended context for the current segment. Considering the $l$-th layer of a Transformer architecture and the $s$-th segment, we denote with $U_{s}^{l}$ the $l$-th layer input representation of the $s$-th input segment (i.e., it corresponds to $U_s$ when $l=0$ and to the output of the previous Transformer layer for the segment $s$, i.e., $Y_s^{l-1}$, when $l > 0$). In \sigla{Transformer-XL}, ${U}_{s}^{l}$ is concatenated with the representations from the previous segments, $Y_{s-1}^{l-1}$, to compose a state that is exploited to produce keys and values as follows,

\begin{equation*}
    \begin{aligned}
    X_{s}^{l}&=[\text{sg}(Y_{s-1}^{l-1})\  |\  U_{s}^{l}]\label{eq:xformerxl}, \\
   Q_{s}^{l}& =X_{s}^{l}W_q, \quad  K_{s}^{l}=X_{s}^{l}W_k, \quad  V_{s}^{l} =   X_{s}^{l}W_v, \\
   Y_s^l &= U_s^{l+1} = \text{tr}^l(Q_{s},K_{s},V_{s}),
\end{aligned}
\end{equation*}
where $\text{sg}(\cdot)$ denotes an operation (i.e., stop-gradient) which avoids the gradient flow during Backpropagation, $[ \cdot | \cdot ]$ denotes concatenation and $\text{tr}^l(\cdot)$ the $l$-th Transformer layer. Notice that the recurrent dependency shifts one-layer downward per segment, as depicted in Figure~\ref{fig:trXL},  which differs from the same-layer recurrence in conventional \sigla{RNNs}. Thus, the largest possible dependency length grows linearly w.r.t. the number of layers as well
as the segment length.
Additionally, it is possible to not limit the state-cache solely to the previous segment, by storing the last $m$ states and using them  as the extra context when processing
the current segment.
Thus, a \sigla{Transformer-XL} with $N$ layers and with a memory of size $m$, has a maximum temporal range of $N \times m$ with a memory complexity in the attention operation of $\mathcal{O}(T^2  + Tm)$ when processing a segment with length $T$. 
\sigla{Transformer-XL} inspired a plethora of model variants \cite{rae2019compressive,wu2022memformer,ding2021ernie,lei2020mart}.  Compressive Transformer~\cite{rae2019compressive} extends the cache with two levels of memory, exploited to store compressed older activations. \sigla{Memformer}~\cite{wu2022memformer} extends the recurrence mechanism from decoder-only architecture to an encoder-decoder architecture. \sigla{R-Transformer} \cite{wang2019r} inputs are first fed to a local \sigla{RNN}, that captures local dependencies, and then to multi-head self-attention module. Please refer to the survey \cite{LIN2022111} (mostly Section 6.4.1) for an overview of these methods. 
Recurrent Memory Transformer (\sigla{RMT}) \cite{bulatov2022recurrent} propose an alternative recurrent approach, where the input segment is augmented with $m$ real-valued \textit{memory} tokens, $M_s$, added both at the beginning and at the end of the input segment  $U_{s}$, as follows, \begin{equation*}
\begin{aligned}
    \hat{U}_{s} &= [M_{s} | U_{s} | M_{s}], \\
\hat{Y}_s &= \text{tr}(\hat{U}_{s}), 
\end{aligned}
\end{equation*}
where the positions in the Transformer output sequence corresponding to the memory $M_{s}$ are interpreted as a read/write memory, i.e., when considering an $N$-layered architecture, the output of the multi-layer transformer can be interpreted as partitioned as follows, $\hat{Y}_s := [M_{s}^{\text{read}} | Y_{s}^N | M_{s}^{\text{write}}]$ (see also Figure~\ref{fig:trXL}). 
The \textit{memory} tokens play a two-fold role: the ones placed at the sequence start allow the standard sequence tokens to attend to memory states produced at the previous segment. The ending group acts as a write memory, that updates its representation based on all the current segment tokens. As a result, $M_{s}^{\text{write}}$ contains updated memory tokens for the $s$-th segment.
Recurrent connection between segments is achieved by passing the memory tokens from the current segment as the input to the next segment,
\begin{equation*}
\begin{gathered}
M_{s+1} := M_{s}^{\text{write}}, \quad
\hat{U}_{s+1} = [M_{s+1} | U_{s+1} | M_{s+1}].
\end{gathered}
\end{equation*}
In \sigla{RMT} the 
Effective context length for recurrence with memory is not limited by the depth of the network,
which is the case for the cache of \sigla{Transformer-XL.}
Moreover, while \sigla{Transformer-XL} stores $m \times T$ vectors per segment, \sigla{RMT} stores only $m$ memory vectors per segment.
\sigla{RMT} is trained with \sigla{BPTT}, and the number of segments to backpropagate is a hyperparameter of the training procedure (the authors tested from 0 to 4 segments). Differently from \sigla{Transformer-XL}, during the backward pass memory-related gradients are not stopped between segments. Figure~\ref{fig:trXL} 
depicts the architectural differences between \sigla{Transformer-XL} and \sigla{RMT}. 
In a recent technical report \cite{bulatov2023scaling}, the authors leverage the \sigla{RMT} architecture to extend the context length of a BERT model \cite{kenton2019bert} up to two million tokens, while maintaining high memory retrieval accuracy. 
Another approach that still falls in the category of models described so far is \sigla{Infini-Transformer} \cite{munkhdalai2024leave}, which incorporates
a compressive memory into a vanilla attention mechanism. It also builds
in both causal local attention and long-term linear attention mechanisms into a single Transformer block. Inspired by Dai et al. \cite{dai2019transformer}, the authors remark that performing attention solely on the current segment (that can be seen as a form of \textit{local attention}) discards the attention states of the previous one. 
To counteract this issue, they propose to maintain the entire context history in the form of a compressive memory. An \sigla{Infini-Transformer} layer contains both global compressive and local fine-grained states. In fact, it maintains as many parallel compressive memories as the number of attention heads, in
addition to the dot-product attention. Each segment is processed via a classic softmax self-attention, producing an output $Y_{\text{loc}}$. Then, a linear attention (see Eq. \eqref{eq:linearatt}) is exploited to retrieve content from the previous segment memory/state, $M_{s-1}$, as follows, 
$$ Y_{\text{mem}} = \frac{\phi(Q)M_{s-1}}{\phi(Q)z_{s-1}},$$
where $\phi(x) = \text{elu}(x) + 1$, $s$ is an index over the segments and $z_.$ is the normalization factor of linear attention (see Eq. \eqref{eq:linearatt}, here we used the parallel form, in matrix notation).  
State/memory update is implemented by a delta-rule-like mechanism \cite{schlag2021linear} (see Section \ref{sec:linear}), first retrieving existing entries (values) and subtracting them
from the new values, before applying the associative bindings, as follows, 

\begin{equation*}
    \begin{aligned}
        M_{s} &= M_{s-1} + \phi(K)'\left(V - \frac{\phi(K)M_{s-1}}{\phi(K)z_{s-1}}\right), \\
        z_{s} &= z_{s-1} + \sum_{t=1}^T \phi(K_t),
    \end{aligned}
\end{equation*}
where $T$ denotes the segment length. 
The new memory states $M_s$ and $z_s$ are then passed to the next segment $s + 1$, building in a recurrence in each attention layer. Finally, local attention $Y_{ \text{loc}}$ and memory retrieved content $Y_{\text{mem}}$ are aggregated via a learned gating mechanism, weighted by a learnable scalar $\beta$, i.e., $Y = \sigma(\beta)Y_{\text{mem}} + (1 - \sigma(\beta))Y_{\text{loc}}$. 

\parafango{Chunk-level Recurrences} 
The recently introduced segment-based approach allows networks to process very long texts  sequentially,  keeping a recurrent memory/context.  
Another category of emerging approaches involving recurrent models that handle portions of text consists of processing sub-portions of the input sequence, referred to as \textit{chunks}, by dividing the sequence into non-overlapping chunks and performing (serial) inter-chunk recurrent computations followed by (parallel) intra-chunk computations. Such a \textit{chunk-wise parallel form} yields sub-quadratic complexity. 
Temporal Latent Bottleneck (\sigla{TLB}) \cite{didolkar2022temporal} divides the input sequence $U = (u_1, \dots, u_\dseq)$ into chunks of fixed size $C$, resulting in $\lfloor\dseq/C \rfloor$ chunks that are sequentially processed one after the other. We denote the $i$-th chunk of the input sequence as $U_{[i]} \defeq U_{iC:(i+1)C} \in \R^{C\times \din}$. Each chunk is processed by a fast-stream (also referred to as \textit{perceptual} module), implemented by a Transformer $\text{tr}(\cdot)$. The fast-stream computation is conditioned via cross-attention on information coming from a slow-stream, referred to as Temporal Latent Bottleneck $\mathcal{G}$, that aggregates cues across chunks and is updated once per chunk. Such a slow-stream is computed recurrently, and it manages a state ${X}$ composed of a set of $N$ $\dstate$-dimensional vectors. The state update in $\mathcal{G}(\cdot)$ is performed via a cross-attention operation in which the queries are obtained by projecting  $X$, while the keys and values come from the output of the perceptual module. Overall, the model performs the following operation,

\begin{equation*}
    \begin{aligned}
    \hat{Y}_{[i]} &= \text{tr}(U_{[i]}, X_{[i]}), \\
   X_{[i + 1]} &= \mathcal{G}(X_{[i]}, \hat{Y}_{[i]}).
\end{aligned}
\end{equation*}
The recurrent update of the \sigla{TLB} state $X$ is performed at lower rates with respect to the computations happening in the perceptual module, fostering the distillation of more stable, condensed and slowly changing features, while the perceptual module captures faster changing local cues.  
From the computational point of view, leveraging a chunk-based approach allows to achieve a  complexity of  $\mathcal{O}\left(\frac{L}{C}(C^2d + CN) \right)$, where $N$ is the number of temporal latent bottleneck state vectors. Hence, it has a much lower computational 
complexity compared to a vanila Transformer applied on the overall sequence. 
A concurrent work, Block-Recurrent Transformer (\sigla{Brect}) \cite{hutchins2022block}, handles the input sequence in a similar manner to \sigla{Transformer-XL}: a document is split into multiple segments of size $T$, processed one after the other. Each segment is processed in chunks (or blocks) using a sliding window attention with size $C$, with a causal mask that forces each token to attend solely to the previous ones. \sigla{Brect} is composed of a recurrent cell that receives as inputs $C$ token embeddings, where $C$ is the block/window size, and a set of $N$ \textit{state} vectors. Similarly to \sigla{TLB}, a two-way processing is performed in the proposed recurrent cell, consisting of the so-called vertical and horizontal ``directions'': the \textit{vertical} direction (i.e., corresponding to the fast-stream in \sigla{TLB}) is implemented by a Transformer layer that performs self-attention over the input tokens of the current block and cross-attends to the recurrent states, producing output token embeddings for the current block. The \textit{horizontal} direction (i.e., the slow-stream in \sigla{TLB}) performs self-attention over the current state vectors, and cross-attends to the block input tokens, producing the next state vectors. Unlike \sigla{TLB}, cross-attention and self-attention are computed in parallel. Recurrence is integrated with the sliding window attention mechanism, since keys and values from the previous window are stored in a differentiable cache and concatenated to the ones of the current block. Residual connections are replaced by gating mechanisms (implemented either as fixed convex combinations or trainable \sigla{LSTM}-like gates) that help the model forget redundant information. For every layer of the architecture, the last recurrent states of a segment are stored in a non-differentiable cache and fed to the following segment in the document as a warm-start. This mechanism extends the sliding window to cover the entire sequence. The cache implements a form of truncated \sigla{BPTT} over long documents.  
Block-State Transformer \cite{pilault2023block} replaces the recurrent cell in \sigla{Brect} with a linear state-space model (see Section \ref{sec:ssm}), which processes the sequence and preserves long-range dependencies, while allowing also for parallel computations.

\begin{figure}
    \centering
    \includegraphics[width=\columnwidth]{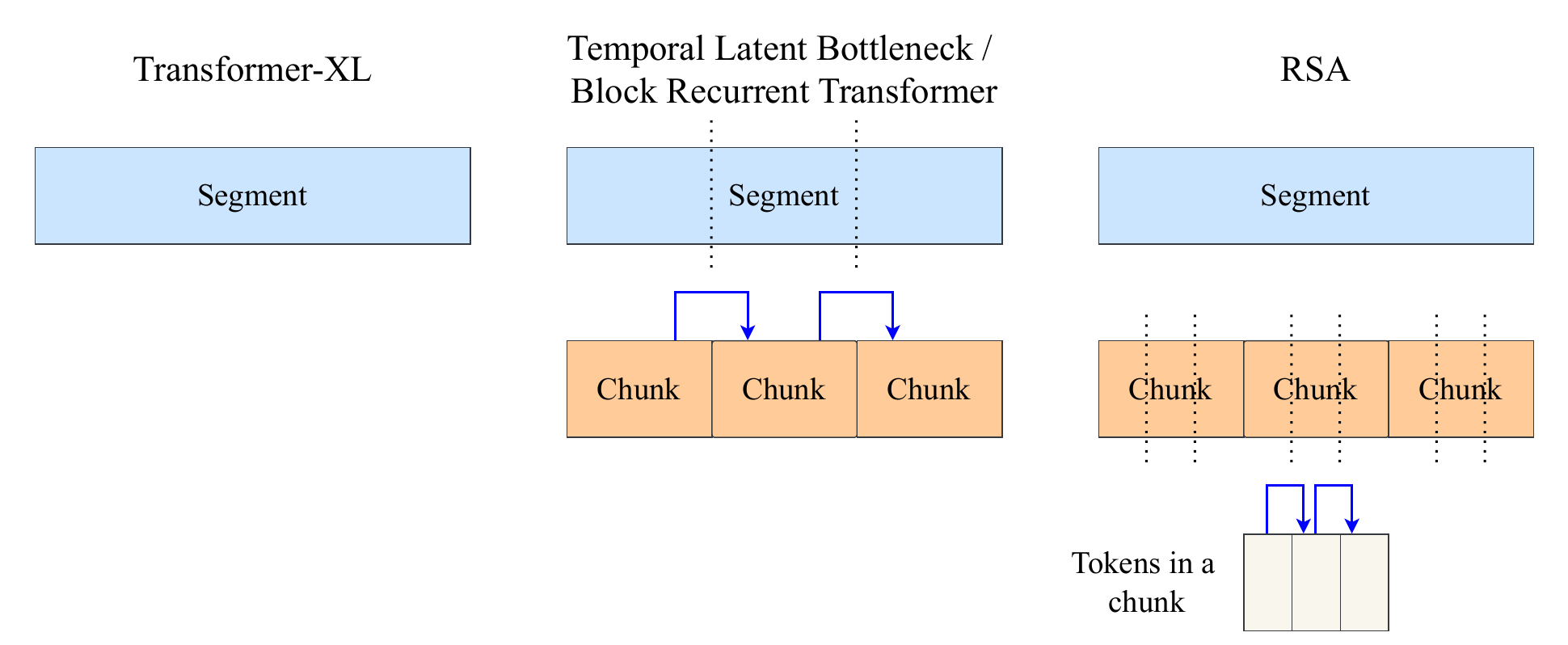}
    \caption{From segment-level to chuck-level recurrence. From left to right: Transformer-XL implements the most coarse-grained segment-level recurrence. Differently, Temporal Latent Bottleneck and Block-Recurrent Transformers include finer-grained chunk-level (or block-level) recurrence. RSA implements the most fined-grained token-level recurrence.}
    \label{fig:xl_vs_tlb_vs_rsa}
\end{figure}

{Recurrent Attention Networks (\sigla{RAN})\cite{LiLL0LZW023} iteratively process the input sequence by means of non-overlapping windows, each of them processed via multi-head self-attention. Intra-window information is propagated by a global perception cell (GPC) vector, extracted from the self-attention representation of the current window. The GPC vector is concatenated with the input tokens of the next window. A memory review mechanism cross-attends  the concatenation of the GPC vectors of all the windows to build the final output, encoding both contextual and global information. }
We report in Figure \ref{fig:xl_vs_tlb_vs_rsa} the main differences in the processing schemes of segment level and chunk-level approaches, reporting examples of some of the described models.

\subsection{Decaying Linear Attention by Gating Functions}  
\label{sec:decay}

Gating mechanisms were introduced since the dawn of \sigla{RNNs}, {being }  a key feature of the popular \sigla{LSTM} model~\cite{hochreiter1997long,qin2023hierarchically}.
In previous subsections and in the context of linear approximations to attention mechanisms, we described several works connecting Linear Transformers to Fast weight Programmers (\sigla{FWP}), which introduced decay/gating mechanisms inherited from \sigla{FWP} \cite{schlag2021linear,irie2021going,irie2022modern,mao2022fine}. 
The original gating functions in \sigla{LSTMs} consisted of units with sigmoidal activations,  responsible for gating specific signals (input, output, state) via multiplicative operations. Lately, the concept of gating has been relaxed to consider any multiplicative interaction, possibly followed by an activation function (e.g., element-wise multiplicative components that do not interact along the sequence length are referred to as gating mechanisms \cite{hua2022transformer,mehta2022long}). Nevertheless, the role of gating has become increasingly popular and now pivotal in many works.

\begin{figure}
    \centering
    \includegraphics[width=1.\columnwidth]{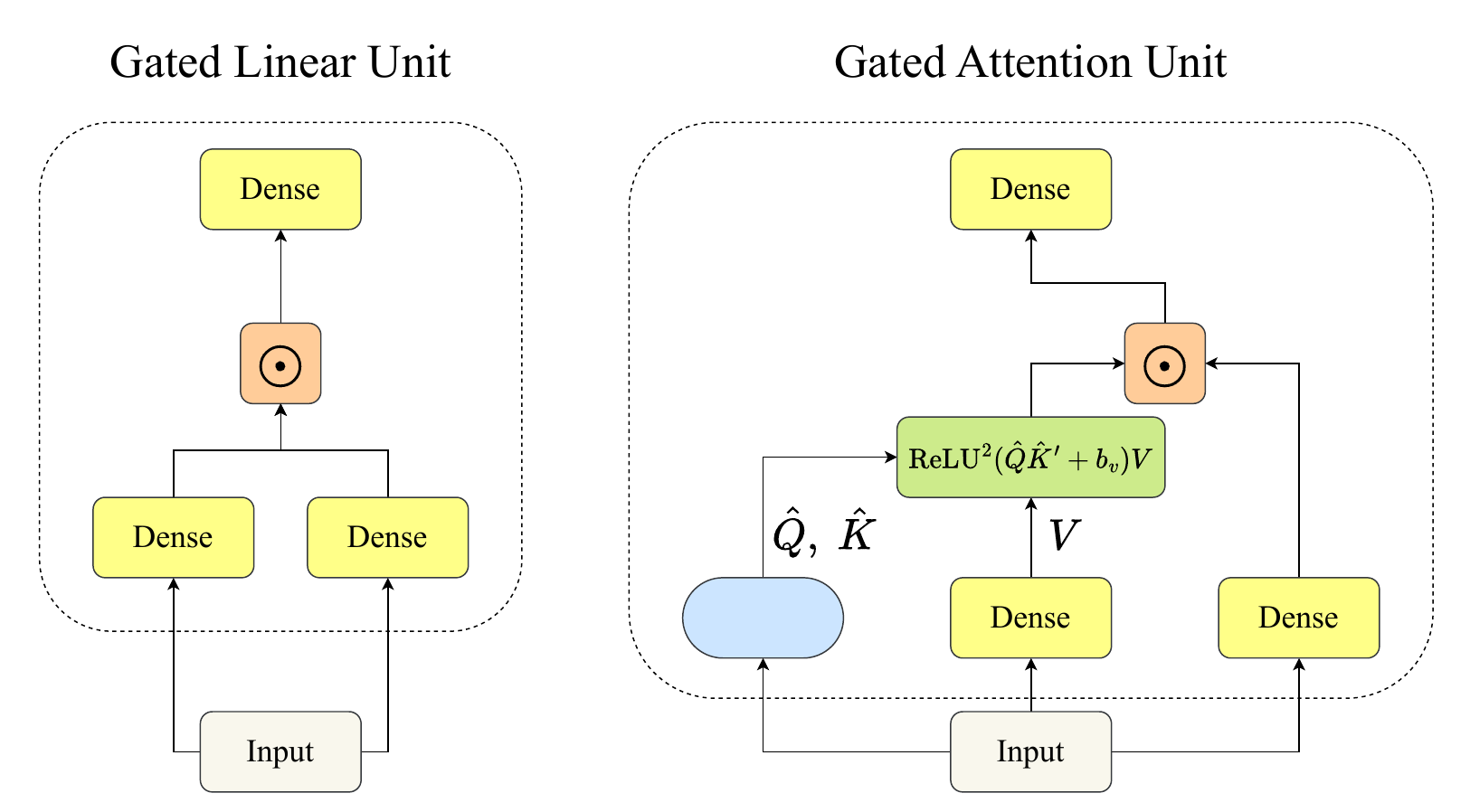}
    \caption{Gated Linear Unit (left) and Gated Attention Unit (right).}
    \label{fig:gau}
\end{figure}

\parafango{Multiplicative Interactions} Hua et al. \cite{hua2022transformer} remarked that despite the linear theoretical complexity claimed by  
linear attention methods \cite{katharopoulos2020transformers}, they have not been able to supersede vanilla Transformers as the dominant choice in state-of-the-art systems. They attributed this to ($i$) approximations needed to achieve efficiency; ($ii$) non-trivial gap between theoretical complexity and empirical speed on accelerators, due to memory re-formatting and operations (see also Section \ref{sec:flopsvsmem}); ($iii$) slow 
 training on causal autoregressive tasks, due to sequential processing of the recurrent forms. To counteract these issues, the  authors proposed \sigla{FLASH}, which is based on a novel layer, dubbed Gated Attention Unit (\sigla{GAU}), to substitute softmax multi-head self-attention. The key idea is to combine a Gated Linear Unit (\sigla{GLU})\footnote{A \sigla{GLU} \cite{dauphin2017language} is an MLP which output is modulated via a gating (i.e., a learned multiplicative interaction). In fact, the layer input $U$ is projected by two learnable matrices $W_p \in \R^{\din \times d_e}$ and $W_t\in \R^{\din \times d_e}$ into two representations $P, T \in \R^{\dseq \times d_e}$, that interact in an element-wise multiplicative manner to produce the layer output, as follows: 
$$
     R= \sigma(UW_p)  \quad V=\sigma(UW_t) \quad Y=(R \odot V)W_y.
 $$
 } \cite{dauphin2017language}
and attention in a unified layer, as depicted in Figure \ref{fig:gau}.
The parallel form of \sigla{GAU} generalizes a \sigla{GLU} as follows,
\begin{equation}
\label{eq:gau}
\begin{aligned}
    R &=\sigma(UW_r), \quad V = \sigma(UW_v), \quad  Z = \sigma(UW_z), \\ 
    \hat{Q} &= \mathcal{Q}(Z),  \quad \hat{K}= \mathcal{K}(Z), \quad \hat{V}=\text{ReLU}^2 (\hat{Q}\hat{K}' + b_v)V,  \\
    Y&=(R \odot \hat{V} W_y),
\end{aligned}    
\end{equation}
where $W_r, W_v \in \R^{\din \times d_e}$, $W_z \in \R^{\din \times d_z}$ with $d_z \ll \dk$, $\mathcal{Q}, \mathcal{K}$ are learned   transformations that apply per-dim scalars and offsets to $Z$, $W_y \in \R^{d_e \times \dout}$ denotes an output learnable matrix, and $b_v \in \R^{d_z}$ are relative position bias. Notice that this formulation substitutes the softmax with the squared ReLU\cite{so2021searching} (denoted with $\text{ReLU}^2$).  
This single-headed softmax-free formulation is capable to achieve multi-head Transformers performances without quality loss, and is cheaper in terms of computational requirements, by adding solely the learnable matrix $W_z$ with $\din \times z$ parameters on top of the \sigla{GLU}. 
Additionally, the authors analyze the causes underlying the  aforementioned memory and speed issues (issues $ii$ and $iii$). In fact, despite the huge advantages brought by the constant-inference computation and memory cost in autoregressive decoding (i.e., thanks to the stateful representation $S_t$) the rearrangement of computations in linear attention lead to an inefficiency in the case of autoregressive training.\footnote{Due to the
causal constraint for auto-regressive training, the query vector corresponds to a different cache value at each time step. This requires the model to compute
and cache $\dseq$ different values  of the incremental state and requires $\dseq$ memory accesses in the sequential loop.} To counteract this, \sigla{FLASH} provides a mixed-chunk attention formulation, where the input sequence is   
chunked into $\lfloor\dseq/C \rfloor$ chunks with fixed size $C$. From each chunk $c$, representations $R_{[c]}, V_{[c]}\in \R^{C \times d_e}$ and $Z_{[c]} \in \R^{C \times d_z}$ are obtained following Eq. \eqref{eq:gau}. Then, an attention mechanism composed by a local (quadratic) and a global (linear) component is formulated. Local attention follows the same procedure of Eq. \eqref{eq:gau} to compute a local chunk-based version of $\hat{V}$, denoted with  $\hat{V}_{[c]}^{\text{loc}}$, given by $\hat{V}^{\text{loc}}_{[c]}=\text{ReLU}^2(\hat{Q}^{\text{loc}}_{[c]}\hat{K}^{{\text{loc}}'}_{[c]})V_{[c]}$. Here $\hat{Q}^{\text{loc}}_{[c]}$ and $\hat{K}^{\text{loc}}_{[c]}$ are the outcome of two ad-hoc per-dim scaling/offsets transformations of $Z_{[c]}$. The complexity of these operations is linear in the sequence length, i.e.,  $\mathcal{O}(\dseq C \din)$. 
Differently, the global linear attention captures long-interactions across chunks, exploiting other ad-hoc per-dim scaling and offsets transformations of $Z_{[c]}$, $\hat{Q}^{\text{glob}}_{[c]}$, and $\hat{K}^{\text{glob}}_{[c]}$. The causal formulation is defined as follows,
\begin{equation}
\label{eq:flash}
\hat{V}^{\text{glob}}_{[c]} = Q_{[c]}^{\text{glob}} \left( \sum_{h=1}^{{[c]}-1} \hat{K}^{{\text{glob}}'}_{[c]}V_h \right).  
\end{equation}
Summation is performed at the chunk level, reducing the number of elements in the cumulative sum by a factor of $C$.
The local and global contributions, $\hat{V}^{\text{loc}}_{[c]},
 \hat{V}^{\text{glob}}_{[c]}$, are added together to yield the final output,  $Y_{[c]} = R_{[c]} \odot (\hat{V}^{\text{glob}}_{[c]} + \hat{V}^{\text{loc}}_{[c]}) ] W_y$. 
 For autoregressive training, thanks to chunking, the sequential dependency in the auto-regressive case reduces from $\dseq$ steps
in the standard linear attention to  $\dseq /C$ steps in the
chunked version in Eq. \eqref{eq:flash}.
Another work focussing on gating functions with multiplicative interactions is the so-called Moving-average Equipped Gated Attention mechanism (\sigla{MEGA}) \cite{ma2023mega}, which injects a temporal locality inductive bias into
the attention mechanism by leveraging a multidimensional exponential
moving average (EMA) \cite{hunter1986exponentially}. The EMA captures local dependencies that exponentially decay over time, and is then integrated with a variant of the single-head \sigla{GAU}.
The multidimensional damped EMA firstly expands each dimension of the input sequence $U$ into $\dstate$ dimensions via an expansion matrix $\beta \in \R^{\din \times \dstate}$, to increase the expressiveness of the model, i.e. $\hat{U} = U\beta$.  Then, the EMA update with state $x_t \in \R^\dstate$ is carried on as follows,
\begin{equation}
\label{eq:mega}
    \begin{aligned}
        x_t &= \alpha \odot \hat{u}_t + (1- \alpha \odot \delta) \odot x_{t-1}, \\
y_t &= \eta'x_t,
    \end{aligned}
\end{equation}
where $\delta \in (0,1)^\din$ is a damping factor,  $\alpha, \beta, \nu \in \R^\dstate$, while $x_t \in \R^\dstate$ is the EMA state at timestep $t$; $\hat{u}_t \in \R^\dstate$ is the expanded input vector at time $t$ (i.e., a column vector that is extracted from a row of $\hat{U}$), and $\eta$ is a projection vector to map the $\dstate$-dimensional hidden state back to 1-dimensional output. 
Despite the recurrent formulation in Eq. \eqref{eq:mega},  computation of EMA can be represented as $t$ individual convolutions, which can be computed efficiently using fast Fourier Transforms (FFTs) (see Section \ref{sec:ssm}).
As we will describe in the next Section, the multi-dimensional damped EMA can be seen as a simplified variant of a state-space
model, and \sigla{MEGA} is closely related to \sigla{S4} \cite{gu2021efficiently}, a state-space model with structured state matrices. The EMA sub-layer in \sigla{MEGA} applies diagonalization on the state matrix and restricts the diagonal
elements in the range of $(0, 1)$. Thus, the convolution kernel would be a Vandermonde product, which can be computed in an efficient and numerically stable way.
The output from Eq. \eqref{eq:mega} is collected into a matrix $\hat{Y}$, which is propagated into a mixed \sigla{GAU}-\sigla{GRU} architecture. The former follows Eq. \eqref{eq:gau} to transform $\hat{Y}$ (the authors leverage a SiLU activation function \cite{ramachandran2017searching} instead of the $\text{ReLU}^2$). Then, the multiplicative interaction in the output inherited from \sigla{GAU} is combined with reset and update gates from \sigla{GRUs}. %
 The authors additionally propose \sigla{MEGA-chunk}, a model variant  with linear complexity due to its chunked-form, where the EMA component takes care of extending the effective context being exploited
by chunk-wise attention. 
 
\begin{figure}
    \centering
    \includegraphics[width=0.8\columnwidth]{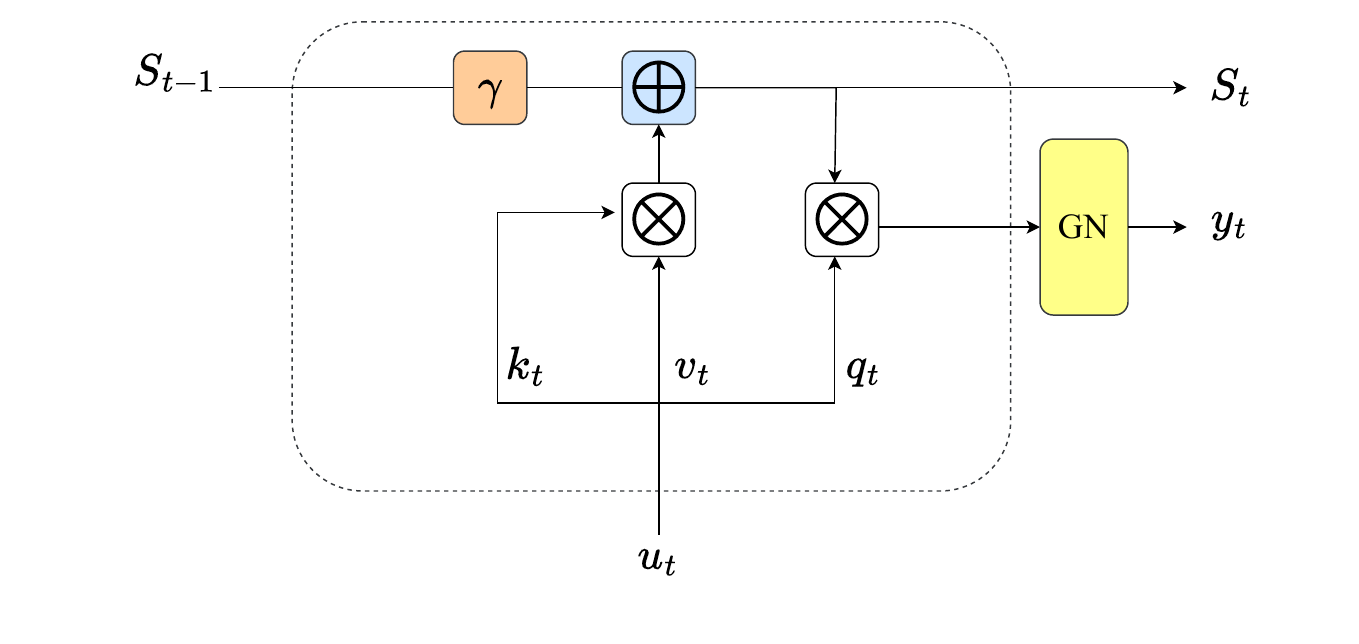}
    \vskip -4mm
    \caption{\sigla{RetNet}, recurrent form (\sigla{GN} is short for GroupNorm).}
    \label{fig:retnet}
\end{figure}

\parafango{Decaying Interactions} Sun et al.~\cite{sun2023retentive} propose a {\it retention} mechanism in place of attention, based on an explicit decay matrix, that controls the ability of each token to pool information from its surrounding tokens based on distance priors. The proposed \sigla{RetNet} encodes the sequence autoregressively. Retention implements a sequence modeling problem in a recurrent fashion, by leveraging a linear recurrence with state $s_t \in \R^{\dstate}$ and a scalar projection of the input, $v_t \in \R$,  obtained via $v_t = w_v'u_t $, as follows, 
\begin{equation}
\label{eq:ret_recurr}
\begin{aligned}
    s_t &= As_{t-1} + k_t \cdot v_t, \\
y_t &= q_t's_t = \sum_{m=1}^t q_t' A^{t-m} k_m v_m,
\end{aligned}
\end{equation}
where $q_., k_.$ are the usual queries and keys computed following Eq. \eqref{eq:general-attn} with projection matrices $W_q, W_k$. Eq.~\eqref{eq:ret_recurr} (top) maps $v_t$ onto the state vector $s_t$, and then implements a linear transformation to encode the sequence information recurrently (bottom).
Matrix $A$ is diagonalized into $A=\Lambda(\gamma e^{i\theta})\Lambda^{-1}$, with $\gamma, \theta \in \R^\dk$. Similarly to \sigla{RSA} \cite{huang2022encoding},  the exponentiation yields  $A^{t-m}=\Lambda(\gamma e^{i\theta})^{t-m}\Lambda^{-1}$. By  simplifying $\gamma$ as a scalar and absorbing $\Lambda$ into the projection matrices $W_q, W_k$, it is possible to simplify Eq. \eqref{eq:ret_recurr} as,
\begin{equation*}
  y_t =  \sum_{m=1}^t \gamma^{t-m} (q_t' e^{it\theta})(k_m e^{im\theta})^\dag v_m,
\end{equation*}
where $\dag$ denotes the conjugate transpose. {Notice the resemblance between the multiplying factors and the \sigla{xPOS} \cite{SunDPMHBCSW23} positional encodings.}
Starting from this formulation, it is easy to obtain the \sigla{RetNet} \textit{parallel form} (Figure \ref{fig:retnet}, left), which is defined  as follows when considering a vector mapping instead of the scalar one of Eq. \eqref{eq:ret_recurr},
\begin{equation*}
\begin{aligned}
\label{eq:ret:parallel}
Q &= (U W_q) \odot \Theta, \quad K = (U W_k) \odot \overline{\Theta}, \quad V = U W_v, \\
 &\quad\quad \Theta_t = e^{it\theta}, \quad\quad\quad
D_{tm} =
\left\{
\begin{aligned}
& \gamma^{t-m}, &t\ge m \\
& 0, &t < m \\
\end{aligned},
\right.
\\
&\hskip -0.35cm \text{Retention}(U) = (QK' \odot D)V,
\end{aligned}
\end{equation*}
where $\overline{\Theta}$ is the complex conjugate of $\Theta$, and $D \in \mathbb{R}^{\dseq \times \dseq}$ contains both causal masking and exponential decay, encoding the prior temporal knowledge as a relative distance in the one-dimensional sequence. This form is particularly advantageous for parallel training.
The retentive mechanism can be directly written into a recurrent form by means of 2D-states, $S_t$, (Figure \ref{fig:retnet}, right),
\begin{equation}
\label{eq:ret:element}
\begin{aligned}
    S_t &= \gamma S_{t-1} + k_t \otimes v_t, \\
y_t &= S_{t}q_t.
\end{aligned}
\end{equation}
It is easy to discern that Eq.~\ref{eq:ret:element} corresponds to  Eq.~\ref{eq:simple_linear_attention} with the addition of the fixed decay factor $\gamma$, which is usually selected as $\gamma=1-2^{-5-b}$, being $b$ a constant.
In order to accelerate training, the authors also propose a chunk-wise recurrent paradigm inspired by the aforementioned inter/intra-level recurrent approaches. 
\textit{Inter-chunk recurrence} propagates the hidden states at chunk-level, followed by \textit{intra-chunk parallel computation} that directly computes the output $Y$ based on the chunk-level hidden states. This approach allows to parallelize computations within a chunk without explicitly \textit{materializing} the intermediate hidden states in the high bandwidth memory (\sigla{HBM}) of the GPU \cite{dao2022flashattention} (see Section \ref{sec:flopsvsmem} for further details on hardware efficiency). 
Formally, the input $U$ is split into non-overlapping chunks, where each chunk is of length $C$. Let $S_{[i]} \in \R^{\dk \times \dk}$ be the chunk-level hidden state after processing $i$ chunks, i.e., $S_{[i]}:=S_{iC}$. The query vector corresponding to the $i$-th chunk is defined as $Q_{[i]}:= Q_{iC+1:(i+1)C} \in \R^{C\times \dk}$, and $K_{[i]}$, $V_{[i]}$, $O_{[i]}$ are similarly defined. 
Then, for $i \in [0, 1, \dots \frac{L}{C} - 1]$, the inter-chunk recurrence is defined as,
\begin{equation*}
\begin{aligned}
    S_{[i+1]} &= 
 \gamma^{C} S_{[i]} + K_{[i+1]}' (V_{[i+1]}\odot\Gamma),
\end{aligned}
\end{equation*}
where $\Gamma_{ij}=\gamma^{C-i}$ for all $j$. The sum of all \sigla{RNN} inputs from a chunk (i.e., $K'_{[i]} V_{[i]}$) can be computed \emph{before} the recurrence {in parallel} in $\mathcal{O}(C^2\din)$. 
The intra-chunk parallel computation is given by,
\begin{equation*}
\begin{aligned}
Y_{[i]} &= \underbrace{\big(Q_{[i]} S_{[i-1]} \big)\odot \zeta }_{\text{cross-chunk}} + \underbrace{\left(Q_{[i]} K'_{[i]} \odot D\right) V_{[i]}}_{\text{intra-chunk}},\\
\zeta_{ij}&=\gamma^{i+1},\ \forall j.
\end{aligned}
\end{equation*}
The \textit{intra-chunk} component is a linear attention performed on the chunk, and thus takes $\mathcal{O}(C^2\dk + C\dk^2)$, while the \textit{cross-chunk} integrates the previous chunk component for the contribution from the hidden state from the previous chunk, and takes $\mathcal{O}(C\dk^2)$. 
Overall, training complexity is  $\mathcal{O}\left(\frac{L}{C}(C^2\dk + C\dk^2) \right)=\mathcal{O}(LC\dk+L\dk^2)$. The chunk size $C$ can be controlled for a trade-off between \sigla{FLOPs} and wall-clock speed. 
Overall, the decay factor introduced by \sigla{RetNet} puts more weight on recently processed inputs and is independent on the processed data. Finally, \sigla{RetNet} exploits multiple heads equipped with retention and a different $\gamma$ for each head, resulting in different variance statistics. GroupNorm \cite{wu2018group} normalizes the output of each head, and a swish gate \cite{hendrycks2016gaussian} increases the non-linearity of retention layers.
\sigla{RetNet} is not the only approach belonging to this category. \sigla{TransNormerLLM} \cite{qin2023scaling} is claimed to be the first linear attention-based Large Language Model (\sigla{LLM}) (up to 175 billion parameters) that outperforms conventional softmax attention-based models in terms of both accuracy and efficiency. It builds upon \sigla{TransNormer}\cite{qin2022devil} (see Section \ref{sec:linear}), replacing diagonal attention with linear attention. addresses  the issue of attention dilution by adding linearized relative positional encodings with exponential decay \cite{qin2023linearized}, linear attention acceleration (by leveraging the recomputation technique from FlashAttention \cite{dao2022flashattention} to avoid the materialization of the 2D hidden state $S_t$ -- see Section \ref{sec:flopsvsmem}), tensor normalization from \cite{qin2022devil}, and a gating mechanism with a decay factor applied to the additive recurrent update. 
\sigla{GateLoop} \cite{katsch2023gateloop} incorporates a data-controlled gating mechanism which is applied on inputs, hidden states and outputs, replacing the fixed decay rate exploited in \sigla{RetNet} with a time-varying data-dependant diagonal state transition $A_t \in \mathbb{C}^{\dstate \times \dstate}$, defined in polar form,
\begin{equation}
    \begin{aligned}
        A_t &= \text{diag}(\gamma_t e^{i\theta_t}) \defeq \text{diag}(\sigma (\alpha_t) e^{i\beta_t}), \\ 
        \label{eq:gatelooprec}
        X_t &= A_tS_{t-1}  + k_t \otimes v_t, \\
        y_t &= S_t q_t,
    \end{aligned}
\end{equation}
where $S_t \in \mathbb{C}^{\dstate \times \dstate}$, $\alpha_t, \beta_t$ are learned linear projections of the input $x_t$ and $\sigma$ is the sigmoid activation. 
Indeed, similarly to \sigla{RetNet} and other recent works (i.e., \sigla{LRU}\cite{orvieto2023resurrecting}, see  Section \ref{sec:ssm}), the magnitude and phase of the state transition $A_t$ are controlled separately. Interestingly, the authors remark how $q_t$ and $k_t$ act
as input and output gates, respectively, and $A_t$ can be interpreted as a forget/retain gate on the linear recurrence.
Unfolding the recurrence in Eq. \eqref{eq:gatelooprec} yields $ y_t =\sum_{m=1}^t  q_t' (k_m \otimes v_m) \prod_{j=m+1}^t A_j$, that equals the \sigla{RetNet} output computation (Eq. \eqref{eq:ret_recurr}) if we fix the state transition gate, $y_t =\sum_{m=1}^t  q_t' (k_m \otimes v_m) A^{t-m}$. Additionally, \sigla{GateLoop} leverages an efficient associative scan computation for efficient parallelized computation of the linear recurrence (see Section \ref{sec:ssm}). 
A concurrent work, \sigla{ReLiT} \cite{pramanik2023recurrent}, investigates index-wise outer-product based gating functions instead of the scalar one  available in previous works \cite{schlag2021linear,irie2022modern} (e.g., $g_t$ and $\beta_t$), as long as an approximation based on trigonometric functions, referred to as \sigla{AReLiT}. The model is a kernel-based Linear Transformer where the authors propose learnable feature maps $\phi$ instead of fixed ones.
Gated Linear Attention (\sigla{GLA})\cite{yang2023gated} explores a data-dependent\footnote{This differs, for instance, from the gating mechanism implemented by \sigla{RetNet}, which decays over time independently with respect to the input.} gating mechanism for linear Transformers, and propose both parallel and block-parallel forms that can take advantage of tensor core computations on modern accelerators (GPUs, TPUs). The recurrent form updates the recurrent state $S_t$ by computing a gating matrix produced by means of an outer-product (similar to \cite{mao2022fine} and \cite{pramanik2023recurrent}), i.e., $G_t =\alpha_t \otimes \beta_t \in \R^{\dk \times \dk}$, where  $\alpha_t, \beta_t \in \R^\dk$. A possible instance of this form is the following one, where $\alpha_t$ is a 
low-rank re-parametrization of the input and  
$\beta_t$ is a column vector filled with ones,
\begin{equation}
   \begin{aligned}
    \alpha_t &= \sigma\big(W_\alpha^2 W_\alpha^1 u_t + b_\alpha \big)^\tau,   \\
    \beta_t &= \mathbb{1}, \\     
    S_{t} &= G_t \odot S_{t-1} + k_t \otimes v_t,
    \label{eq:gla_recurrence} 
\end{aligned}
\end{equation}
where $\sigma$ is the sigmoid function, $W_\alpha^1 \in \R^{16 \times \dk}, W_\alpha^2 \in \R^{\dk \times 16}$ implement a low-rank parametrization, $\tau \in \R$ is a temperature term to encourage the model to have a slower forgetting rate. 
Overall, the output of the recurrent form of the \sigla{GLA} layer is,
\begin{equation}
   \begin{aligned}
& o_{t} = S_tq_t,  \\
& r_t = \text{Swish}(W_ru_t + b_r),  \\ 
& y_t = (r_t \odot \text{LayerNorm}(o_t)) W_o,  
\end{aligned}
\end{equation}
where the $\text{LayerNorm}$ after $o_t$ follows prior work (also referred to as $\text{NormAttention}$)~\cite{qin2022devil,qin2023scaling,sun2023retentive}. The final output $y_{t}$ is obtained by following the structure of a \sigla{GAU} layer from the \sigla{FLASH} model~\cite{hua2022transformer}, where an additional output gate $r_t$  with the $\text{Swish}$ activation \cite{hendrycks2016gaussian} is used.
The \sigla{GLA} parallel form is computed as follows,
$$
O = \left(\left((Q \odot B)\left(\frac{K}{B}\right)'\right)\odot M\right)V, $$
where $B \in (0,1)^{\dseq \times \dk}$ is the matrix obtained by stacking $b_t \defeq \prod_{j=1}^t \alpha_j$ and $M$ denoted the causal mask.\footnote{For stability, it is computed in log space (see the referenced paper for further details).} 
 \sigla{GLA} also provide a parallel and two-level chunk-wise block-parallel forms. See Section \ref{sec:flopsvsmem} for further details on their hardware-aware solutions.

{This section shows that the once-sharp boundary between self-attention and recurrence is rapidly eroding. Modern Transformer research first adopts kernel or linear attention schemes (e.g., \sigla{Linear Transformer}, \sigla{Performer}) that rewrite the softmax dot product as a feature-kernel product. This reframes attention as an additive update of a constant-size matrix state and cuts the quadratic $\mathcal{O}(L^2)$ cost of vanilla attention to linear $\mathcal{O}(L)$ while preserving full parallel training. Extending that idea, fast-weight and delta-rule methods such as \sigla{DeltaNet} and \sigla{SRWM} treat the key–value cache as a programmable associative memory, using outer-product updates and data-controlled forgetting gates to balance capacity and recency. A parallel line of work eliminates dot-products entirely: low-rank or channel-wise interactions in Attention-Free Transformers and the \sigla{RWKV} family compute gated exponential averages that naturally down-weight distant tokens and enable linear-time decoding. Finally, segment and chunk recurrent architectures such as \sigla{Transformer-XL}, \sigla{RetNet} and their variants propagate compressed hidden states across windows, extending effective context length without incurring quadratic growth. Together these threads reveal a unifying trend: contemporary Transformers increasingly resemble specialised recurrent networks, compressing history into compact learned states, applying explicit decay or memory tokens to regulate long-range influence, and thereby reclaiming much of the efficiency once unique to classic \sigla{RNNs}.}

{While this section demonstrated that many Transformer variants now reuse recurrent ideas, an orthogonal line of work pursues the same goal from first principles: \emph{State-Space Models} (SSMs). By discretising continuous-time dynamics, SSMs encode sequence history in compact linear states and thus inherit the favourable $\mathcal{O}(L)$ scaling of classical RNNs without relying on attention. Section~\ref{sec:ssm} reviews this rapidly growing family and its role in long-sequence processing.}


\section{Deep State-Space Models}
\label{sec:ssm}

\begin{table}
\caption{
Complexities for sequences of length $L$, comparing representatives of the categories of Deep State-Space models discussed in Section~\ref{sec:ssm}, vanilla Transformers and Recurrent Models (Section~\ref{sec:related}). In order to bridge the discussion in the main text of paper, here we assumed $d_k=\din$, for the sake of simplicity.}
\centering
\setlength{\tabcolsep}{2pt}
\small
\begin{tabular}{r|lccc}
\toprule
 & \textsc{Model} & \textsc{Recur.} & \textsc{Time} & \textsc{Space}\\
\midrule
Sec.~\ref{sec:related} & Rec. Net & Yes & $\mathcal{O}(\dseq\din^2)$ & $\mathcal{O}(\dseq\din)$  \\
Sec.~\ref{sec:transformers} & Transf. & No &  $\mathcal{O}(\dseq^2\din)$ & $\mathcal{O}(\dseq^2 + \dseq\din)$ \\
\midrule
\multirow{1}{*}{Sec.~\ref{sec:linear}} & H3  & Yes & $\mathcal{O}(\dseq \din(\text{log}\dseq + \din))$ & $\mathcal{O}(\dseq\din)$ \\
\multirow{1}{*}{Sec.~\ref{sec:alt}} & S4  & Yes & $\mathcal{O}({\dseq\din^2})$ & $\mathcal{O}({\dseq\din})$ \\
\multirow{1}{*}{Sec.~\ref{sec:subrec}} & Hyena  & Yes & $\mathcal{O}(\dseq \din(\text{log}\dseq + \din))$ & $\mathcal{O}(\dseq(\text{log}\dseq \cdot \din))$ \\
\multirow{1}{*}{Sec.~\ref{sec:decay}} & Mamba  & Yes & $\mathcal{O}({\dseq\din})$ & $\mathcal{O}({\din})$ \\
\bottomrule
\end{tabular}
\label{tab:complexity_comparison}
\end{table}

\begin{figure*}[h]
    \centering    \includegraphics[width=0.9\linewidth]{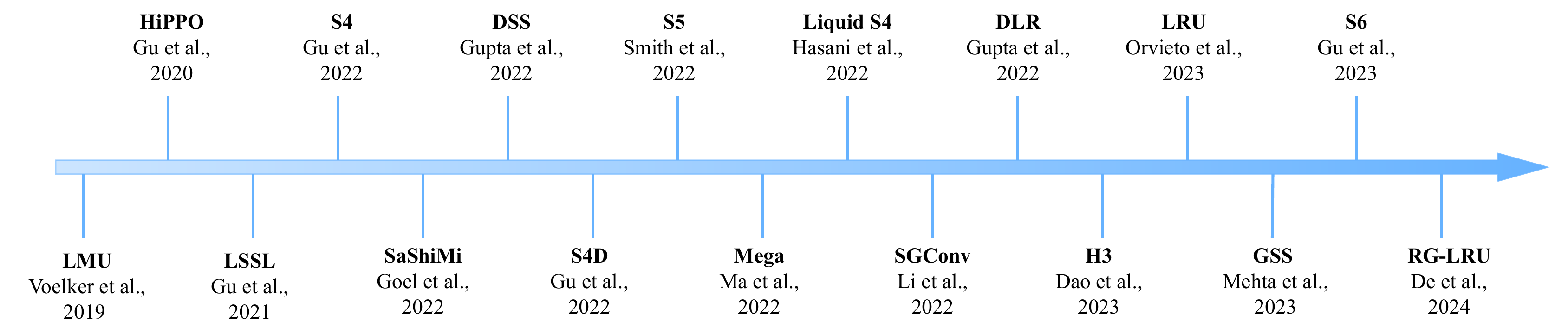}
    \caption{Timeline showing the chronological development of deep state-space models---Overview of the organization of Section \ref{sec:ssm}.}
    \label{fig:timeline_ssm}
\end{figure*}

Recent works on models that are intrinsically based on recurrent computations particularly emphasize the notion of (deep) State-space Models \cite{gu2021combining}. In particular, there exists a growing interest in exploiting the computational advantages of using multiple stacked instances of
linear recurrences, whose dynamics are appropriately conditioned
to avoid trivial explosive/vanishing dynamics.
The development of such lines of works can be traced back to the seminal work of~\cite{voelker2019} and~\cite{gu2020hippo}, in which the authors propose methods to perform \emph{online function approximation}. Then, the scientific literature of the last years transitioned from focusing on online function approximation to specifically designed (deep) State-Space Models \cite{gu2021combining,gu2021efficiently} and more advanced architectures exploiting them \cite{gu2023mamba,de2024griffin}, as we anticipate in Figure~\ref{fig:timeline_ssm}.

\parafango{Online Function Approximation}
The basics of online function approximation, with regard to the first works on this novel wave of state-space models, can be formalized as follows.
Given a function of one variable  defined on the
half line $u\colon [0,+\infty)\to\R$, the problem of online
approximation of such function is twofold: ($i$)
for time instant $t\in [0,+\infty)$ 
find an approximation of $u$ until $t$,  i.e., $u^t:=u|_{I_t}$ 
with $I_t:=(0,t)$  and ($ii$)
have a method to update online such approximation. 
In order to formalize the concept of approximation, we need to
have some notion of closeness, and hence we  assume that the function we want to approximate lies in some normed space.
Moreover, the measure with respect to which we define
the notion of integrability plays a rather important role in computing an
online approximation. In the following descriptions, we will mostly refer to the seminal works
in \cite{gu2020hippo}, where the \sigla{HiPPO} model/theory is introduced, and \cite{voelker2019}, based on Legendre Memory Units (\sigla{LMUs}). In~\cite{gu2020hippo}
the authors find that working with a normalized Lebesgue
measure on $I_t$ (which is the standard choice in $\R^n$) has
several advantages. A different choice is explored in \sigla{LMU} \cite{voelker2019}
that, in light of the theoretical formulation of the problem presented
in~\cite{gu2020hippo}, correspond to choosing a measure of density that is
constant on a window $[t-\theta,t]$ of size $\theta$ just before the end
point $t$ of the considered temporal instant.
The other basic ingredient to consider in function approximation is the class of basis functions with which 
we want to perform such an approximation.
In~\cite{gu2020hippo} the authors consider the case of translated and rescaled Legendre polynomials, $v^t_n$ for 
$n=1,2,\dots$, defined in $[0,t]$ by,
\begin{equation}\label{eq:legendre-rescaled}
v^t_n(x)=\sqrt{2} e_n\Bigl(\frac{2x}{t}-1\Bigr)\quad \hbox{$\forall x\in
[0,t]$},\quad n=0,1,\dots,
\end{equation}
where $e_n$ are normalized Legendre polynomials
(see~\cite{ciarlet2013linear}). A similar choice has been also done
in~\cite{voelker2019}.
Then, the wanted approximation $v^t$ of
the function $u^t$
can be expressed (as explained in~\cite{gu2020hippo}) by,
\begin{equation}\label{eq:coeff_def}
v^t=\sum_{n=0}^{N-1} c_n(t) v^t_n\quad \hbox{where}\quad
c_n(t):=(u^t,v^t_n)_t,
\end{equation}
where  $(u^t,v_n^t)_t:=\int_{I_t} u^t v_n^t\, dx/t$ is the standard 
scalar product in $L^2((0,t);\R)$ rescaled by a factor $1/t$.
More precisely, since the goal is to solve an approximation problem on
$I_t$, in order to define integrability we will consider the
Lebesgue measure $\leb^1$ on $\R$ restricted to $I_t$ and we will
define $\forall t>0$ the rescaled measure  $\leb^1_t$ such that
$\leb^1_t(A)=\leb^1(A)/t$ for all
$A\subset I_t$.\footnote{Notice that $\leb^1_t$ is a probability measure on
$I_t$ since beside being a well defined measure we also have that
$\leb^1_t(I_t)=1$.}
One we have this measure we can define the Hilbert
space $L^2_{\leb^1_t}(I_t;
\R)$ which is exactly the space of square $\leb^1_t$-integrable,
real-valued functions. 
So it is natural to require that the method that we will develop works on
functions $u\colon\overline\R_+\to\R$ such that for all $t>0$
$u|_{I_t}\in L^2_{\leb^1_t}(I_t;\R)$.
The approximation problem then can be stated as the problem of
finding a solution to the following minimization problem,\footnote{
This problem has always a unique solution since the subspace
$V^t_N$ is finite dimensional and hence it is closed}
\begin{equation}\label{eq:approximation-problem}
\min_{v\in V^t_N} \Vert v-u^t \Vert_{L^2_{\leb^1_t}(I_t;\R)},
\end{equation}
where $V^t_N\subset L^2_{\leb^1_t}(I_t;\R)$ is a finite,
$N$-dimensional subspace that we assume to be
spanned by $N$ \emph{orthonormal} basis functions $v^t_0,\dots, v^t_{N-1}$;
i.e., $V^t_N:=\spn\{v^t_0,\dots, v^t_{N-1}\}$. Here othornormality
as usual means that $(v^t_i, v^t_j)_t=\delta_{ij}$ for all
$i,j=0,\dots, N-1$ where $\delta_{ij}$ is the usual Kronecker delta
and 
$(\cdot, \cdot)_t$ is the standard scalar product in
$L^2_{\leb^1_t}(I_t;\R)$, that is
$(f,g)_t:=\int_{I_t} f g\, d\leb^1_t\equiv (\int_{I_t} f g\, dx)/t$
being $dx$ the usual notation for the Lebesgue measure.
In general the  solution to the problem in
Eq.~\eqref{eq:approximation-problem}
(see~\cite{ciarlet2013linear}) is given by 
\begin{equation}\label{eq:coeff_def}
v^t=\sum_{n=0}^{N-1} c_n(t) v^t_n\quad \hbox{where}\quad
c_n(t):=(u^t,v^t_n)_t.
\end{equation}
The crucial result presented in~\cite{gu2020hippo} and~\cite{voelker2019}
is that the computation of the coefficients $c_n$ defined above
can be done using system of ordinary differential equations
with a Cauchy initialization so that they can be estimated in
an online manner. In particular if we denote with $c:=
(c_0,\dots, c_{N-1})$ then $c$ can be computed as a solution of 
\begin{equation}\label{eq:hippo-eq}
\dot c(t)=A(t) c(t)+B(t) u(t),
\end{equation}
where the matrix $A(t)$ and the vector $B(t)$ can be explicitly computed.
In particular in the \sigla{HiPPO} setting these matrices turns out to be
(see Appendix~E of~\cite{gu2020hippo}),
\begin{equation}\label{eq:hippo-matrices}
\begin{aligned}
&A_{ij}(t)=-\frac{1}{t}\begin{cases}
\sqrt{(1+2i)(1+2j)}& \hbox{for $i>j$}\\
1+i &\hbox{for $i=j$}\\
0& \hbox{for $i<j$}\\
\end{cases},\\
&B_i(t)=\frac{1}{t}\sqrt{1+2i},
\end{aligned}
\end{equation}
where the temporal dependence takes the form of a rescaling $1/t$
that, in turn, comes from the choice of the measure $\mathcal{L}^1_t$
defined above.

\begin{figure}[h]
\centering
\includegraphics[width=0.8\columnwidth]{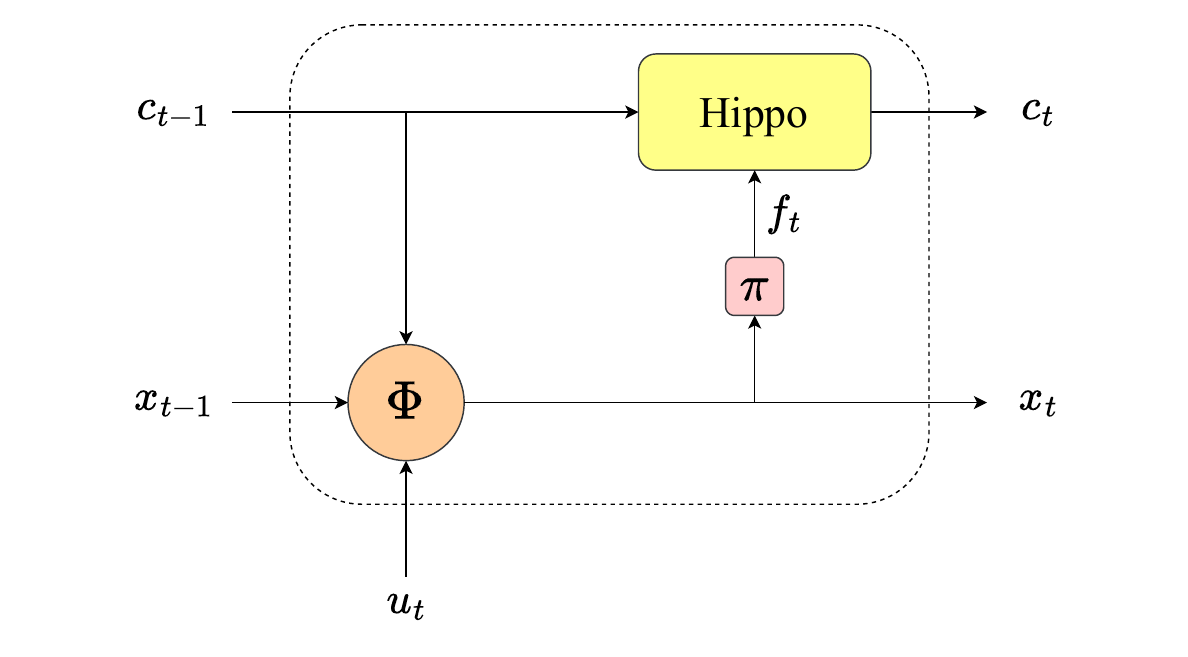}
\caption{Online function approximation methods based in ODE,
such as \sigla{HiPPO} \cite{gu2020hippo}, can be
seamlessly added to a recurrent computation of a standard RNN.
We indicate with $\Phi$ the state transition function, $\pi$ is the
``one dimensional'' projection described in the main text of the paper, and the
\sigla{HiPPO} module performs the integration of the \sigla{HiPPO}
equations, Eq.~\eqref{eq:hippo-eq}, with $u$ replaced by $f$.}
\label{fig:OFA}
\end{figure}

\begin{figure}[h]
\centering
\includegraphics[width=\columnwidth]{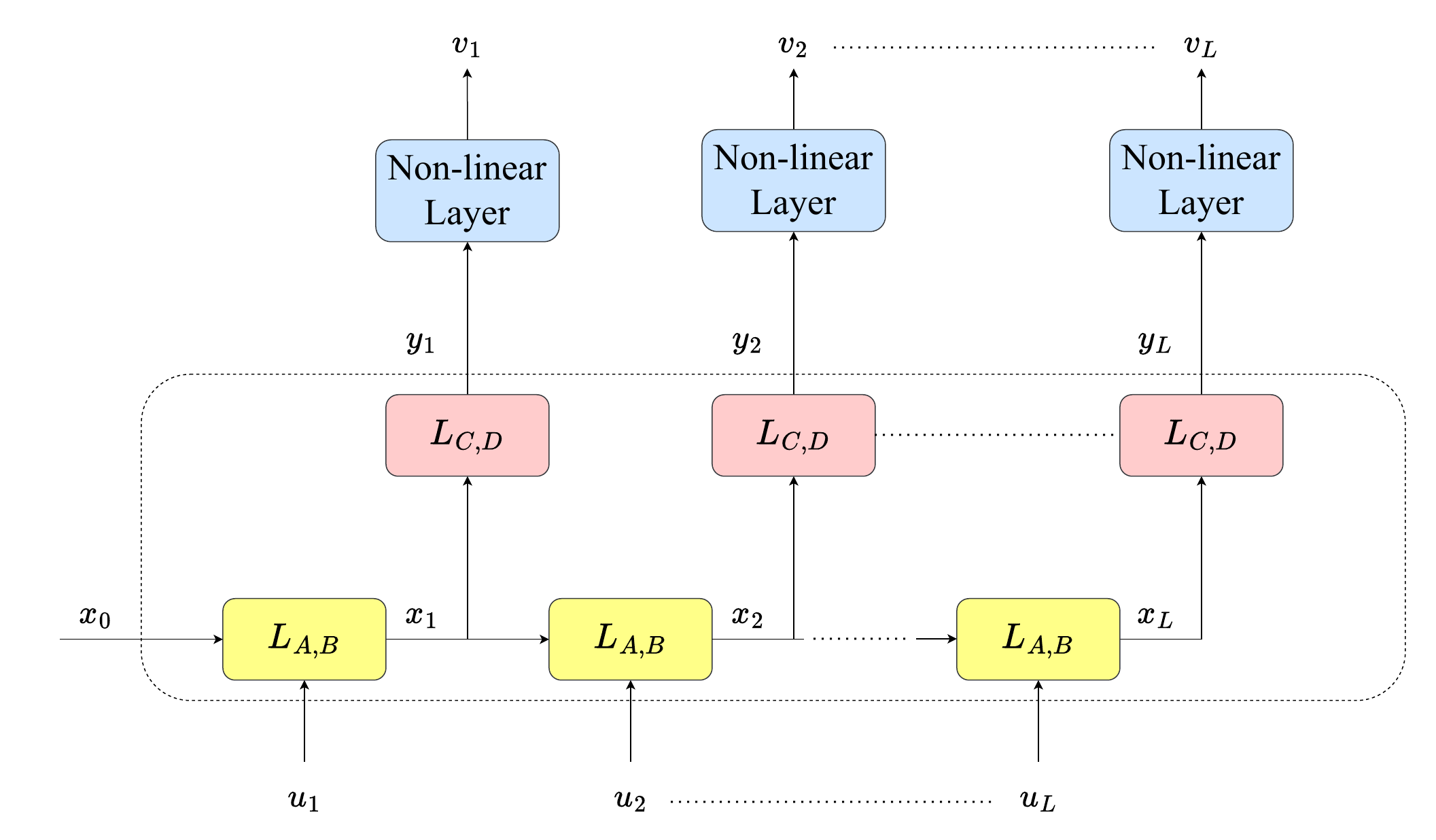}
\caption{A Linear State Space Layer, \sigla{LSSL}, is a \emph{nonlinear}
transformation that maps a sequence of length $L$, $(u_1,\dots,
u_L)$ into another sequence $(v_1,\dots, v_L)$ through a
\emph{linear} recurrence and a non-linear feed-forward layer.
In this figure we indicate with $L_{A,B}$ the linear transformation
that defines the transition function of the recurrence
with parameters A and B (as in Eq.~\eqref{eq:linear-state-model}), while
$L_{C,D}$ is the linear function that define the
static mapping $x\mapsto Cx + Du$.}
\label{fig:LSS}
\end{figure}

\parafango{From Online Approximation to Deep State-Space Models}
Online function approximation \emph{is not} a learning problem;
however the results of~\cite{voelker2019} and~\cite{gu2020hippo}
discussed above show that the coefficients of the online approximation can be efficiently updated using a linear
recurrence relation. Hence, in both works the authors propose a direct application
of this idea to learning, using this online approximation
mechanism inside a recurrent network to maintain a compact representation
of a projection of the state over the whole past times (Figure~\ref{fig:OFA}). More precisely,
the state of the \sigla{RNN} at time $t$ is updated using the following update rule,
\begin{equation}
\label{eq:general_RNN_form}
\state_t=\Phi(\state_{t-1}, c_{t-1}, \inp_t),
\end{equation}
where $\Phi$ is the transition function that depends on the precise
recurrent architecture, $\inp_t$ is the input of the net at time $t$ and
$c_{t-1}$ are the coefficients of the online approximation of the function
$f_t=\pi(\state_t)$, where  $\pi\colon\R^\dstate\to\R$ is a projection of the state onto
the real line (which is necessary, since these online approximation
methods work on scalar functions).
\def\<#1>{\includegraphics{./figures/mpost/dss-#1.mps}}
The leap from this hybrid model, where the state of the recurrence
is enriched with an online approximation of the state itself, to
Deep State-Space Models has been proposed in~\cite{gu2021combining}, which analyzes Linear State Space Layers (\sigla{LSSLs}) in comparison to other deep
learning models that are used to process sequences,
and in~\cite{gu2021efficiently}, which refines such models to address
computational limitations of the former model.
In Linear State Space Layers (Figure~\ref{fig:LSS}), the main idea is to use a
linear continuous-time expression to model the update of the state itself,

\begin{equation}
\label{eq:linear-state-model}
\dot\state(t)= A \state(t) +B \inp(t),
\end{equation}
where the input signal $t\mapsto\inp(t)\in\R$ is one dimensional,
and processing of higher dimensional signals of features is 
achieved by learning independent models for each input dimension. The matrix $A\in\R^{\dstate\times\dstate}$
while $B\in\R^{\dstate\times1}$. As it is customary in state space models,
the output trajectory of the model, that we will denote as $t\mapsto \out(t)\in\R$, is then computed via another ``static''
(i.e., not involved in a recurrence) linear map,
\begin{equation}\label{eq:output-map}
y(t)= C x(t)+Du(t),
\end{equation}
where $C\in\R^{1\times\dstate}$ and $D\in\R$.

\begin{figure}[h]
    \centering
    \includegraphics[width=\columnwidth]{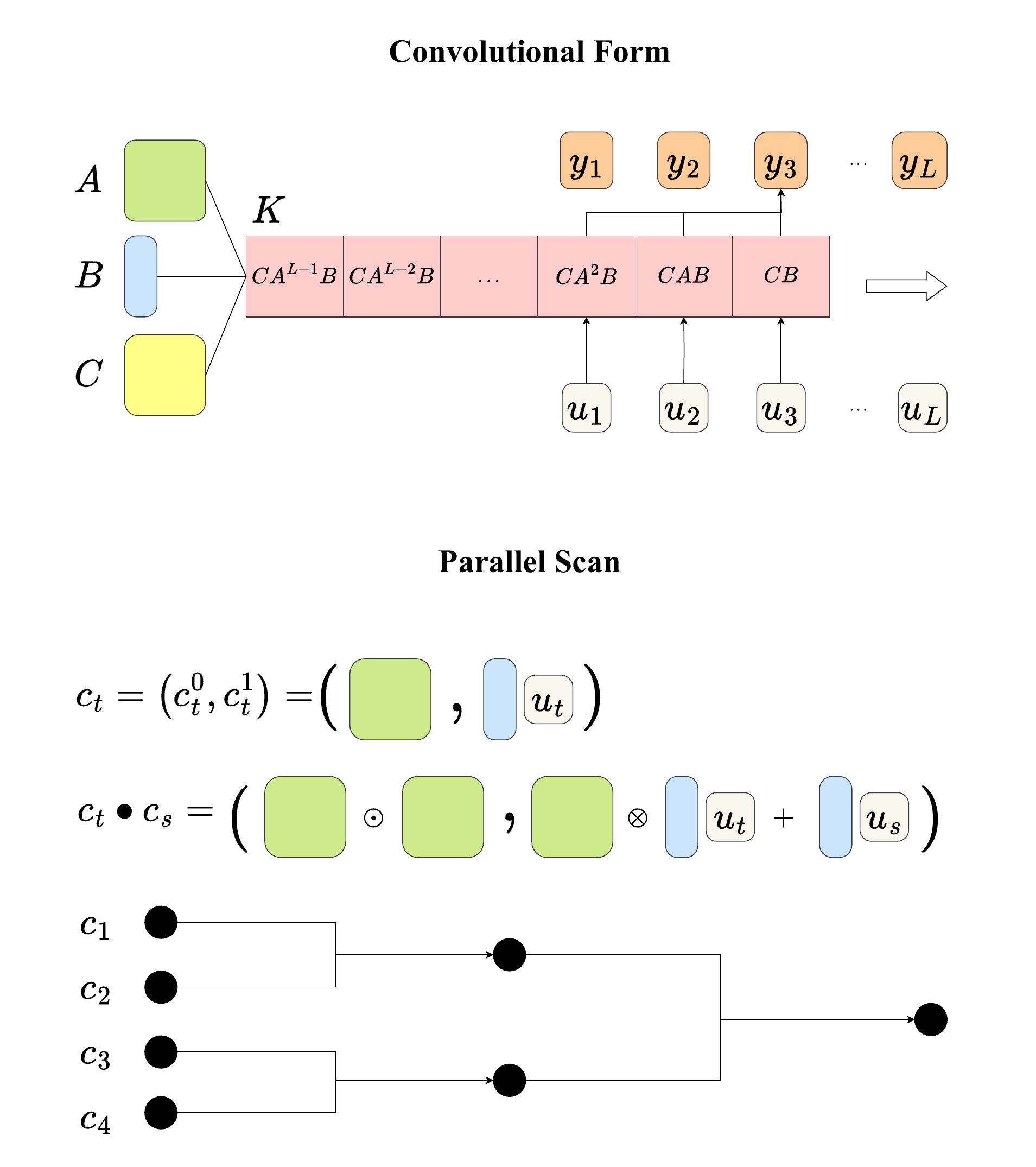} 
    \caption{Methods to parallelize computations over sequence length in Linear RNNs: Convolutional form, computed with FFT transform \cite{gu2021efficiently}, and Associative Parallel Scans \cite{martin2018parallelizing,smith2022simplified}}
    \label{fig:parallel-scan}
\end{figure}

The continuous time model described by Eq.~\eqref{eq:linear-state-model}
is typically discretized in order to be numerically implemented. Different discretization techniques can be applied, the more direct being the
explicit (or forward) Euler method,\footnote{ Technically this is a mixed
scheme since the input is computed at the step $t+1$, however we call it
explicit since it is so with respect to the state variable.}
\begin{equation}\label{eq:euler-forward}
\state_{t+1}= \state_t +\tau\bigl(A\state_k+B\inp_{t+1}
\bigr).
\end{equation}
Where, $(\state_t)$ and $(\inp_t)$ are sequences. More generally, a typical
discrete approximation of Eq.~\eqref{eq:linear-state-model}
will have the form
\begin{equation}\label{eq:discrete-SSM}
\state_{t+1}= A^\tau \state_t +B^\tau\inp_{t+1},
\end{equation} 
where for instance $A^\tau=\Id+\tau A$ and $B^\tau=\tau B$ for the forward
Euler scheme described in Eq.~\eqref{eq:euler-forward}, being $\Id$ the identity matrix.
Another very common discretization scheme for
Eq.~\eqref{eq:linear-state-model} used in the context of Deep State-Space Models
(see~\cite{gu2021combining}) is the \emph{bilinear method}, that is equivalent to the choice $A^\tau=(\Id-(\tau/2)
A)^{-1}(\Id+(\tau/2)A)$ and $B^\tau=(\Id-(\tau/2) A)^{-1}\tau B$.
On the other hand, the output map described in Eq.~\eqref{eq:output-map} remains exactly the same and defines,
in the discrete setting, the sequence of outputs $(\out_t)_{t>0}$
defined in terms of the state sequence as $\out_t=
C\state_t+D\inp_t$.
Assuming for definiteness that $x_t\equiv 0$ if  $t<0$, the recursion relation in Eq.~\eqref{eq:discrete-SSM}
can be unfolded to obtain a closed expression for the
$t$-th element of the sequence of the state in terms
of the inputs $\inp_0,\dots,\inp_t$. Indeed, it is immediate
(by repeated use of Eq.~\eqref{eq:discrete-SSM}) that,
\[\begin{aligned}
\out_t &= CB^\tau \inp_t+CA^\tau B^\tau \inp_{t-1}+ C(A^\tau)^2 B^\tau \inp_{t-2}\\
&\qquad {}+\dots+ C(A^\tau)^t B^\tau \inp_0 + Du_t\\
&=\sum_{j=0}^t C(A^\tau)^j B^\tau \inp_{t-j} +Du_t.
\end{aligned}
\]
Now, if we define the sequence of real numbers $(p^\tau_t)_{t\ge0}$ as $p^\tau_t:= C(A^\tau)^t B^\tau\in\R$, the outputs can
be expressed as a convolution of the input $\inp$ with the
sequence $(p^\tau_t)_{t\ge0}$,
\begin{equation}\label{eq:convolution-view}
\out_t =\sum_{j=0}^t p^\tau_j \inp_{t-j} +Du_t.
\end{equation}
This is what is commonly referred to as the convolutional form
of the linear state space model in Eq.~\eqref{eq:discrete-SSM} -- see Figure~\ref{fig:parallel-scan}-top.
Now, going back to \sigla{LSSL} \cite{gu2021combining}, the matrix $A$ is represented with a suitable matrix factorization,
i.e., $A=P(D+T^{-1}Q)$, with $D$, $P$ and $Q$ diagonal matrices and 
$T$ tridiagonal. The \sigla{HiPPO} matrix $A(t)$ defined in Eq.~\eqref{eq:hippo-matrices}
admits such factorization (see Appendix E.2 of~\cite{gu2021combining}). In this way, matrix $A$ is guaranteed to be quasiseparable, a property
that is presented as desirable both for handling long 
dependencies and for enabling efficient matrix-vector multiplication. 
As a common practice in deep learning, several \sigla{LSSLs} can be stacked together, each layer receiving as input the output of the previous one. This is possible since the input and output have the same dimensionality. The main problem with this architecture is the computational
cost{;}
 indeed (see~\cite{gu2021efficiently}), it
requires $\mathcal{O}(\dstate^2\dseq)$ operations and it scales as $\mathcal{O}(\dstate
\dseq)$ for what concerns memory in order to compute the input-output 
mapping in Eq.~\eqref{eq:discrete-SSM}.
In order to overcome this precise limitation
the \sigla{S4} model~\cite{gu2021efficiently} has been introduced 
to condition the structure of the matrix $A$. There is a large set of works that were published in the last few years along this line of research, and that, starting from \sigla{S4}, we describe in the following. Refer to Figure~\ref{fig:timeline_ssm} for an overview.

\parafango{S4}
The {Structured State Space Sequence Model} (\sigla{S4}) \cite{gu2021efficiently} is based on the continuous-time linear system in Eq.~\eqref{eq:linear-state-model} and Eq.~\eqref{eq:output-map}. The matrix $A$ is imposed to have the following form, 
\begin{equation}\label{s4:parametrization}
    A = \text{diag}(\lambda_1,\dots,\lambda_\dstate) + PQ^\dag,
\end{equation} 
where $\text{diag}(\lambda_1,\dots,\lambda_{\dstate})$ is a diagonal matrix, $\lambda_i\in\mathbb{C}$ for every $i$ and $P,Q\in\C^{\dstate\times 1}$. In Eq.~\eqref{s4:parametrization}, the $\dag$ operation denotes the conjugate transpose and the term $PQ^\dag$ is usually referred to as \textit{low-rank correction}. With this particular choice for $A$, the computation 
of the recursion in Eq.~\eqref{eq:linear-state-model} has complexity
$\tilde{\mathcal{O}}(\dstate+\dseq)$. For discretizing the dynamics in Eq.~\eqref{eq:linear-state-model} and Eq.~\eqref{eq:output-map}, the bilinear transform  with discretization step $\tau$ is applied, leading to the already introduced,
\begin{equation}
\begin{aligned}
    A^{\tau} &= \left(I - \frac{\tau}{2}A\right)^{-1}\left(I + \frac{\tau}{2}A\right), \\
    B^{\tau} &= \left(I - \frac{\tau}{2}A\right)^{-1}\tau B.
\end{aligned}
\label{eqn:s4_discretization}
\end{equation}
The model follows a single input single output (SISO) structure, meaning each component of the input (called \emph{input channel}) $u_i$, for $i=1,\dots,\din$, is processed by a distinct discretized system, each generating a scalar output $y_j$, for $j=1,\dots,\din$ (notice that $\dout = \din$). The dynamics matrix $A$ for each of the $\din$ SISO subsystems is initialized according to \sigla{HiPPO} theory. While the original \sigla{S4} does not inherently favor initialization towards marginal stability to maintain long-range memory, the subsequent work \sigla{SaShiMi} \cite{goel2022s} ensures stability by enforcing the real part of $\lambda_i$ to be negative, $\text{Re}(\lambda_i) \in \mathbb{R}^{-}$, for every $i$. For training \sigla{S4}, the convolutional representation in Eq.~\eqref{eq:convolution-view} of the output is used and the structure of $A^{\tau}$ in Eq.~\eqref{s4:parametrization} is exploited for efficiently computing its inverse. At inference time, a recurrent representation of the model,
\begin{equation}
\begin{aligned}
    x_{t+1} &= A^{\tau}x_t + B^{\tau}u_t, \\
    y_t &= Cx_t + Du_t,
\end{aligned}
\label{eq:discrete-dyn}
\end{equation}
is directly used. Subsequent works show that it is possible to match the performances of \sigla{S4} even without the low rank correction, but
still retaining the initialization of the diagonal part to be consistent with
the diagonal part of the \sigla{HiPPO} matrix. The diagonal structure of the matrix $A$ leads to the \emph{Diagonal State Space} (\sigla{DSS}) model \cite{gupta2022diagonal} and this work is theoretically expanded in the infinite width setting in \cite{gu2022parameterization}, leading to \sigla{S4D}.

\parafango{S4D}
The {Diagonal Structured State Space Sequence Model} (\sigla{S4D}) \cite{gu2022parameterization} builds upon \sigla{S4}, and it assumes that the matrix $A$ has a diagonal structure,
\begin{equation}\label{eqn:s4d_parametrization}
    A = \text{diag}(\lambda_1, \dots, \lambda_{\dstate}),
\end{equation}
which yields computational improvements. In order to get its discrete-time version, exact discretization is applied to the dynamics of Eq.~\eqref{eq:linear-state-model} and Eq.~\eqref{eq:output-map}, with discretization step $\tau$, leading to,
\begin{equation}\label{eqn:s4d_discretization}
\begin{aligned}
    A^{\tau} &= e^{\tau A},\\
    B^{\tau} &= (\tau A)^{-1} (A^{\tau} - I) \tau B.
\end{aligned}
\end{equation}
\sigla{S4D} retains the SISO structure from \sigla{S4} and its initialization is still based on \sigla{HiPPO} theory. Similar to \sigla{SaShiMi}, the eigenvalues of $A$ used for initialization lie in $\mathbb{R}^{-}$. Again, convolutional representation of Eq.~\eqref{eq:convolution-view} is used in training, and the recurrent representation in Eq.~\eqref{eq:discrete-dyn} is used during inference. The diagonal structure of the matrix $A^{\tau}$ allows for efficient computation of the discretization in Eq.~\eqref{eqn:s4d_discretization}. 
The \sigla{SSMs} described so far, as showed in \cite{dao2022hungry}, struggle with tasks like recalling earlier tokens and comparing tokens across a sequence when applied to language modeling tasks (see also Section~\ref{sec:expressivitygap}). The \sigla{H3} model is explicitly designed to tackle these challenges and will be described in the following, right after having extended the SISO family of models to the more advanced MIMO.

\parafango{From SISO to MIMO: S5}
The {Simplified Structured State Space Sequence Model} (\sigla{S5}) \cite{smith2022simplified} is the first Deep \sigla{SSM} to be parameterized leveraging the concept of multiple input multiple output (MIMO) systems, for simplifying the architectural components and enhancing computations. This means that the full input vector $u \in \mathbb{R}^\din$ is fed into a single bigger MIMO system of Eq.~\eqref{eq:discrete-dyn}, instead of $\din$ SISO scalar smaller subsystems, by stacking the matrices $A^{\tau}, \, B^{\tau}, \, C$, used in \sigla{S4} and \sigla{S4D}. \sigla{S5} inherits the \sigla{S4D} parametrization of the matrix $A$ (i.e., it is a diagonal matrix), while it can be discretized applying both bilinear (as done in \sigla{S4}, see Eq.~\eqref{eqn:s4_discretization}) and exact (as done in \sigla{S4D}, see Eq.~\eqref{eqn:s4d_discretization}) discretizations. The MIMO structure and the diagonal parameterization allows for parallel computation of the individual output components via a \emph{parallel scan} algorithm (see Appendix~H of~\cite{smith2022simplified}).  The parallel scan
algorithm (see Figure~\ref{fig:parallel-scan}-bottom) offers a way to parallelize the computations of a sequence of
elements of a \emph{semigroup} $(S,\bullet)$ generated by a recurrence
relation of the the form $s_{i+1}=s_i\bullet c_i$, where $(c_i)_{i=1}^\dseq$
is a given sequence of elements of $S$.\footnote{ We recall that a semigroup
consist of a set $S$ together with an associative operation $\bullet$.}  This
approach can be directly applied to the computation of a linear recurrence of
the form in Eq.~\eqref{eq:discrete-SSM} with the following choices,
\begin{enumerate}
\item $S=\{\, (M,v): M\in\R^{\dstate\times\dstate}\quad\hbox{and}\quad
v\in\R^\dstate\,\}$;
\item $(M,v)\bullet(N,u):= (NM,Nv+u)$ for all $(M,v)\in S$ and $(N,u)\in S$;
\item $c_i= (A^\tau, B^\tau\inp_i)$.
\end{enumerate}
Indeed one can show (see Appendix~H of~\cite{smith2022simplified}) that the
sequence
\[\begin{cases}
s_0=(\Id, 0)\in S\\
s_{i+1}=s_i\bullet (A^\tau, B^\tau\inp_i)
\end{cases}
\]
has the following  representation in terms of the solution $\state_k$
of Eq.~\eqref{eq:discrete-SSM} with zero initialization $\state_k\equiv 0$
for $k\le0$,
\[
s_k=((A^\tau)^{k-1}, \state_k)\quad k\ge0.
\] Therefore, computations at training and inference time are made efficiently in the recurrent representation of Eq.~\eqref{eq:discrete-dyn}. \sigla{HiPPO} theory is again used for initializing the matrix $A$, obtaining the same starting eigenvalues of \sigla{S4D}. Together with \sigla{S5}, some novel variants of \sigla{S4} are introduced. Recent literature describes also \sigla{Mega} \cite{ma2023mega} (see Section \ref{sec:transformers}) as a \sigla{SSM}. Indeed, it can be interpreted as a simplification of \sigla{S4} to a diagonal \sigla{SSM} where the values in the diagonal of the matrix $A$ are restricted to be real numbers, interpreting it as an exponential moving average (EMA). 
\sigla{Liquid S4} \cite{hasani2021liquid} exploits the original \sigla{S4} formulation (with low-rank correction) combined with liquid time-constant networks (please refer to Section \ref{sec:odernn} for further details on liquid time-constant networks).  \sigla{SGConv} model \cite{li2022makes} leverages the convolutional form of Eq.~\eqref{eq:convolution-view}
to obtain a filter-based version of \sigla{S4}. Up to this point, \sigla{SSMs} rely on discretizing the continuous time dynamics in Eq.~\eqref{eq:linear-state-model}. The authors of \cite{gupta2022simplifying} eliminate the discretization step and introduce a model based on vanilla \emph{Diagonal Linear RNNs} (\sigla{DLR}), closely related to \sigla{DSS} and \sigla{S4D}, in which each input is processed independently at each layer. Here, the discretization step is directly absorbed into the continuous-time transition matrix $A$. The authors show that, after numerical integration, diagonal state-space models and linear RNNs share the same function approximation class. 

\parafango{SSMs in Language Modeling: H3}
The {Hungry Hungry Hippo} (\sigla{H3}) \cite{dao2022hungry} model is a novel approach to leverage \sigla{SSMs} for language modeling, aiming to address the limitations of previous \sigla{SSMs} in tasks like Associative Recall and Induction Heads (see Section~\ref{sec:benchmarks}) compared to attention-based models. \sigla{H3} draws inspiration from linear attention, which assumes a specific form for the similarity metric used in attention calculations. It stacks two discrete \sigla{SSMs}: one with a shift matrix (i.e., a local convolution) to remember past tokens and one with a diagonal matrix to retain state over the entire sequence. The key innovation lies in introducing \emph{gates} (i.e., multiplicative interactions) between the outputs of these \sigla{SSMs} and projections of the input {. Combined with the shift matrix, these gates enable \sigla{H3} to compare tokens across the sequence}. The shift \sigla{SSM} identifies specific events (like the presence of a key token), while the diagonal \sigla{SSM} stores and retrieves associated information (like the corresponding value token). \sigla{H3} has a time complexity of $\mathcal{O}(\dseq \log(\dseq))$ for a sequence of length $\dseq$, making it asymptotically more efficient than traditional attention, which has a complexity of $\mathcal{O}(\dseq^2)$.

\parafango{LRU}
Another model belonging to the family of MIMO systems, as \sigla{S5}, is the {Linear Recurrent Unit} (\sigla{LRU}) \cite{orvieto2023resurrecting}. The pre-processing of the input and post-processing of the output are also identical to those in \sigla{S5}. Instead, \sigla{LRU} is the first of the \sigla{SSMs} that does not come from a discretization of a continuous-time model, since a discrete parametrization of $A^{\tau}$ and $B^{\tau}$ is directly used. Indeed, it parameterizes the discrete-time dynamics in Eq.~\eqref{eq:discrete-dyn} as,
\begin{equation}\label{eq:lru_parameterization}
    A^{\tau} = e^{-e^{\, diag\left(\lambda_1,\dots,\lambda_{\dstate}\right)} + i\, diag\left(\theta_1,\dots,\theta_{\dstate}\right)}, \quad\quad B^{\tau} = e^{\gamma} \Gamma,
\end{equation}
where $i$ is the complex unit, $\lambda_j,\theta_j\in\mathbb{R}$ for every $j=1,\dots,\dstate$, $\Gamma \in \mathbb{C}^{\dstate \times \din}$ is a dense complex-valued matrix, and $\gamma \in \mathbb{R}$. Given this parameterization, the eigenvalues of $A^{\tau}$ in polar coordinates (i.e., $a_j =  r_j + i\ \theta_j$ where $r_j = e^{-e^{\lambda_j}}$) are constrained to lie in the unit-disk, by construction. The initialization is then directly performed in polar coordinates by defining a range for $r$ and $\theta$ in which they are uniformly sampled. This provides an alternative to the \sigla{HiPPO} theory for initialization (the \sigla{HiPPO} theory is instead used in \sigla{S4}, \sigla{S4D} and \sigla{S5}). Moreover, it is also the first formalization where $A^{\tau}$ and $B^{\tau}$ do not share parameters. As for \sigla{S5}, the model is implemented using a parallel scan algorithm for both training and inference.

\parafango{S6: Mamba}
Unlike previous models, the {Scan Selective Structured State Space Sequence Model} (\sigla{S6}) \cite{gu2023mamba} introduces a linear time-varying representation of the dynamics in Eq.~\eqref{eq:linear-state-model} and Eq.~\eqref{eq:output-map}, which is referred to as a \text{``selection mechanism''}. This is achieved by letting the parameters that affect interactions along
 the sequence (e.g., the recurrent dynamics of the RNN) be input-dependent.
Indeed, the matrices $B$ and $C$ are now functions of the input $u_t$, at every time-step $t$, parametrized as,
\begin{equation}
\begin{aligned}
    B_t &= W_B u_t \\
    C_t &= W_C u_t,
\end{aligned}
\label{eq:s6_parametrization}
\end{equation}
where $W_B$ and $W_C$ are linear projection matrices of appropriate dimensions.
Similar to \sigla{S4D}, the matrix  $A$ is a time-invariant diagonal matrix, as in Eq.~\eqref{eqn:s4d_parametrization}, and it uses exact discretization to compute the discrete-time dynamics of Eq.~\eqref{eq:discrete-dyn}. However, in \sigla{S6}, also $\tau$ is time-varying and function of the input, leading to the discretization,
\begin{equation}
\begin{aligned}
    \tau_t &= \text{softplus}(W_\tau u_t), \\ 
    A^{\tau}_t &= e^{\tau_t A}, \\ 
    B^{\tau}_t &= (\tau_t A)^{-1} ( A^{\tau}_t - I) \tau_t B_t,
\end{aligned}
\label{eq:s6_discretization}
\end{equation}
where $C^{\tau}_t = C_t$, $D^{\tau}_t = D_t$ and $W_\tau \in \mathbb R^{1\times \din}$. The model is structured in a MIMO manner (as \sigla{S5}, \sigla{LRU} and \sigla{S6}) and the dynamic matrix $A$ is initialized with $\lambda_j = -j$, for every $j = 1, \dots, \dstate$, ensuring that the eigenvalues lie in the negative halfplane. Since $\tau_t$ is time-varying, the eigenvalues of $A^{\tau}_t$ have an initialization that is input-dependent. 
The time-varying representation presents computational challenges, despite being more expressive. The authors provide an efficient implementation of the time-varying dynamics in Eq.~\eqref{eq:discrete-dyn}, presenting a variation of the parallel scan and exploiting it both at inference and training time. \sigla{S6} introduces an innovative way of pre-processing the input, called \sigla{Mamba}, which relies on both linear and non-linear maps. The input enters the recurrence through a linear projection, followed by a causal convolution. Additionally, it passes through a linear projection followed by a SiLU nonlinearity before entering the gating function for post-processing. The gating function is inspired by previous models, i.e., \sigla{H3} and \sigla{GAU}. An architecture close to  \sigla{Mamba} {is the \emph{Gated State Space} (\sigla{GSS}) layer} \cite{mehta2022long}, again inspired by \sigla{GAU}. 
\sigla{GSS} resembles \sigla{Mamba} but includes additional projections. The key difference is that \sigla{GSS}'s projection reduces the model dimension to decrease the state size of the \sigla{SSM}, whereas \sigla{Mamba} expands the model dimension to increase the state size.

\parafango{RG-LRU: Hawk and Griffin}
The {Real-Gated Linear Recurrent Unit} (\sigla{RG-LRU}) \cite{de2024griffin} fuses ideas from \sigla{LSTMs}, \sigla{LRU}, and \sigla{S6}. As in \sigla{LRU}, the \sigla{RG-LRU} model is structured by means of a MIMO system and, as in \sigla{S6}, the parametrization of the linear dynamics is time-varying. Unlike all previous \sigla{SSMs}, the matrices $C$ and $D$ are not present here. \sigla{RG-LRU} does not rely on a continuous-time representation (the same thing happens in \sigla{LRU}) and directly parametrizes the discrete matrices $A^{\tau}_t,\ B^{\tau}_t$ as,
\begin{equation}
\begin{aligned}
    A^{\tau}_t &= e^{-c \cdot \text{softplus}(W_A) \sigma(W_\tau u_t)},\\
    B^{\tau}_t &= \sqrt{1 - A_t^2}\sigma(W_B u_t),
\end{aligned}
\label{eq:rglru_parameterization}
\end{equation}
where $\sigma(\cdot)$ is the sigmoid function, $W_\tau,\ W_A,\ W_B$ are linear projection matrices (of appropriate dimensions) initialized with standard initialization methods, e.g., Glorot, and $c \in \mathbb{R}$ is a scalar constant. The square root operation is computed element-wise. Given this parameterization of $A^{\tau}_t$, its eigenvalues are restricted to the unit disk by construction. The implementation of \sigla{RG-LRU} assumes that the state and input dimensions coincide, i.e., $\dstate=\din$. Since the parametrization in Eq.~\eqref{eq:rglru_parameterization} is time varying, \sigla{RG-LRU} exploits a customized variation of the parallel scan algorithm in both training and inference. 
The authors introduce two additional task-specific pre-post-processing operations close to \sigla{Mamba} that are tailored to language modelling: \sigla{Hawk} and \sigla{Griffin}.
\sigla{Griffin} blends gated linear recurrences with local attention, aiming for both performance and efficiency. 
\sigla{Griffin} employs \sigla{RG-LRU} to efficiently process sequences by compressing information into a fixed-size hidden state that is iteratively updated. The gating mechanism in \sigla{RG-LRU} enables it to retain important information from the past while filtering out less relevant inputs, enabling the model to potentially learn long-range dependencies in the sequence. In addition to \sigla{RG-LRU}, \sigla{Griffin} incorporates local multi-query attention \cite{beltagy2020longformer} to focus on a limited window of nearby tokens, while processing each part of the sequence.
\sigla{Hawk} still uses the \sigla{RG-LRU} layer, but relies solely on gated linear recurrences for temporal mixing, making it a pure \sigla{RNN}-based model.  Please refer to Sections \ref{sec:expressivitygap} and \ref{sec:flopsvsmem} for further considerations on the \sigla{Griffin} model expressivity and efficiency.

\parafango{Hyena}
\sigla{Hyena} \cite{polihyena} is a novel approach designed as a more efficient alternative to the attention mechanism prevalent in large language models. It is based on a recurrent structure, where each step involves two key components: ($i$) a long convolution operation, implicitly parameterized using feed-forward neural networks for efficiency, and ($ii$) element-wise multiplicative gating, which selectively modulates the information flow. More precisely, given three linear projections $q,k,v$ of the input $u$, each of length $\dseq$ in $\mathbb{R}^{\din}$, \sigla{Hyena} maps the input $u_t$ in $(\mathcal{H}u)_t$ through,
\begin{equation} \label{eq:multisiso_hyena}
    \begin{aligned}
        {(\mathcal{H} u)}_t^i &= u_t^i + \sum_{j=0}^{\din-1}\sum_{m=0}^t R^{ij}q_t^j h^j_{t-m} k_m^j v_m^j,
    \end{aligned}
\end{equation}
for $i = 0, \dots, \din - 1$, where $h^j_t$ are implicit long convolution filters learned by shallow feed-forward neural networks and $R\in\R^{\din\times \din}$ is an output projection that mixes channels across the sequence length.  This approach decouples filter length from parameter cost, providing advantages over explicit parameterization. The number of recurrent steps determines the complexity of the operator , and they can be represented as a decomposition of data-controlled matrices. These matrices dynamically adapt based on the input data, similar to how attention mechanisms compute a weighted sum over input elements. Instead of explicitly computing the full data-controlled matrix, \sigla{Hyena} leverages fast convolution algorithms, particularly Fast Fourier Transform (FFT)-based convolutions, to achieve subquadratic time complexity. Moreover, unlike some models that restrict the receptive field, it allows for interactions between any elements in the sequence through its long convolutions, enabling it to capture long-range dependencies effectively.

In subsequent work \cite{massaroli2023laughing}, some improvements to further enhance the efficiency of Long Convolution Sequence Models (\sigla{LCSMs}), including \sigla{Hyena}, have been introduced {: the \sigla{LaughingHyena} distilling. It focuses specifically on improving}  the inference stage of these models, particularly in auto-regressive generation tasks. \sigla{LaughingHyena} distillation {aims to represent each convolutional filter of a pre-trained \sigla{LCSM} as an \sigla{SSM}}, with the smallest state dimension such that it approximates the original filter without significant loss of accuracy. To achieve this goal, it utilizes a method called \emph{modal interpolation}, which provides coefficients for the numerator and denominator of a rational function that minimizes the difference between the original filter and the approximating \sigla{SSM} transfer functions. Through these coefficients, it is possible to define the matrices $A$, $B$, and $C$ which characterize the \sigla{SSM}. This distillation procedure is then followed by two steps: ($i$) \emph{pre-filling} and ($ii$) a \emph{recurrent update} rule. Pre-filling involves computing the state to start generating new tokens when a length-$\dseq$ prompt is fed to the model during auto-regressive generation, exploiting the denominator of the approximate transfer function. The recurrent update rule for the complex state is defined as follows

\begin{equation}
\begin{aligned}
    x_{t+1} &= A x_t + B u_t, \\
    y_t &= \text{Re} (C x_t) + h_0 u_t.
\end{aligned}
\end{equation}
Here $h_0$ denotes the value of the original filter at initial time and $\text{Re}(\cdot)$ is the real part operator (since a real-valued output is usually required). Beyond language processing, \sigla{Hyena} has also been employed for time series forecasting \cite{zhang2023effectively} and for DNA sequence analysis \cite{nguyen2024hyenadna}.

\parafango{Theoretical Foundations of SSMs}Performances achieved by \sigla{SSMs} are remarkable, thus inspiring several research efforts to understand both their expressive capabilities and the connections to existing popular technologies (such as attention), with which they share many features but have been commonly developed in isolation. Orvieto et al. \cite{orvietoUniversalityLinearRecurrences2024} theoretically show that combining MLPs with either real or complex linear diagonal recurrences (such as in \sigla{S4}, \sigla{Mamba}, etc.) enables highly precise approximation of regular causal sequence-to-sequence maps. The proof is based on the fact that the linear RNN provides a lossless encoding of the input sequence, and the MLP conducts nonlinear processing on this encoding. While real diagonal linear recurrences are sufficient for achieving universality, employing complex eigenvalues near the unit disk, a strategy that has shown empirical success in \sigla{S4}, significantly improves the ability of recurrent models to store information. Cirone et al. \cite{cirone2024theoretical} leverage tools from Rough Path Theory, and provide theoretical grounding for the fact that, when random linear recurrences are enhanced with simple input-controlled transitions (selectivity mechanism), the hidden state is demonstrably a low-dimensional projection of a mathematical construct known as the signature of the input. This signature captures nonlinear interactions between tokens across different timescales. Other recent works focus on the connections and differences between \sigla{SSMs} and other sequence processing models~\cite{he2023unified,sieber2024understanding}, as long as to their links with control theory \cite{alonso2024state}.  
{Deep \sigla{SSMs} such as \sigla{S4}, \sigla{Mamba} and \sigla{H3} demonstrate that stacking linear recurrences with carefully chosen spectra can match or exceed attention on long contexts while retaining $\mathcal{O}(L)$ time and memory. Key themes include \textit{(i)} diagonal, structured or low-rank transition matrices; \textit{(ii)} convolutional or parallel-scan implementations for hardware efficiency; and \textit{(iii)} theoretical guarantees that a linear core plus shallow non-linearity is a universal causal operator. These results confirm that recurrence—when stabilised—remains a first-class tool for sequence modelling.}

{The \sigla{SSMs} literature reveals that careful structuring of linear recurrence can recover scalability. Building on that insight, the community has revisited traditional \sigla{RNNs} themselves, refining gating, constraining weight geometry, or casting the dynamics in continuous time, to push their performance further. The following Section~\ref{sec:other_rnns} reviews these architectural enhancements to “plain” recurrent networks.}

\section{{Enhancing Recurrent Neural Networks}}
\label{sec:other_rnns}

\begin{figure}
    \centering
    \includegraphics[width=0.9\linewidth]{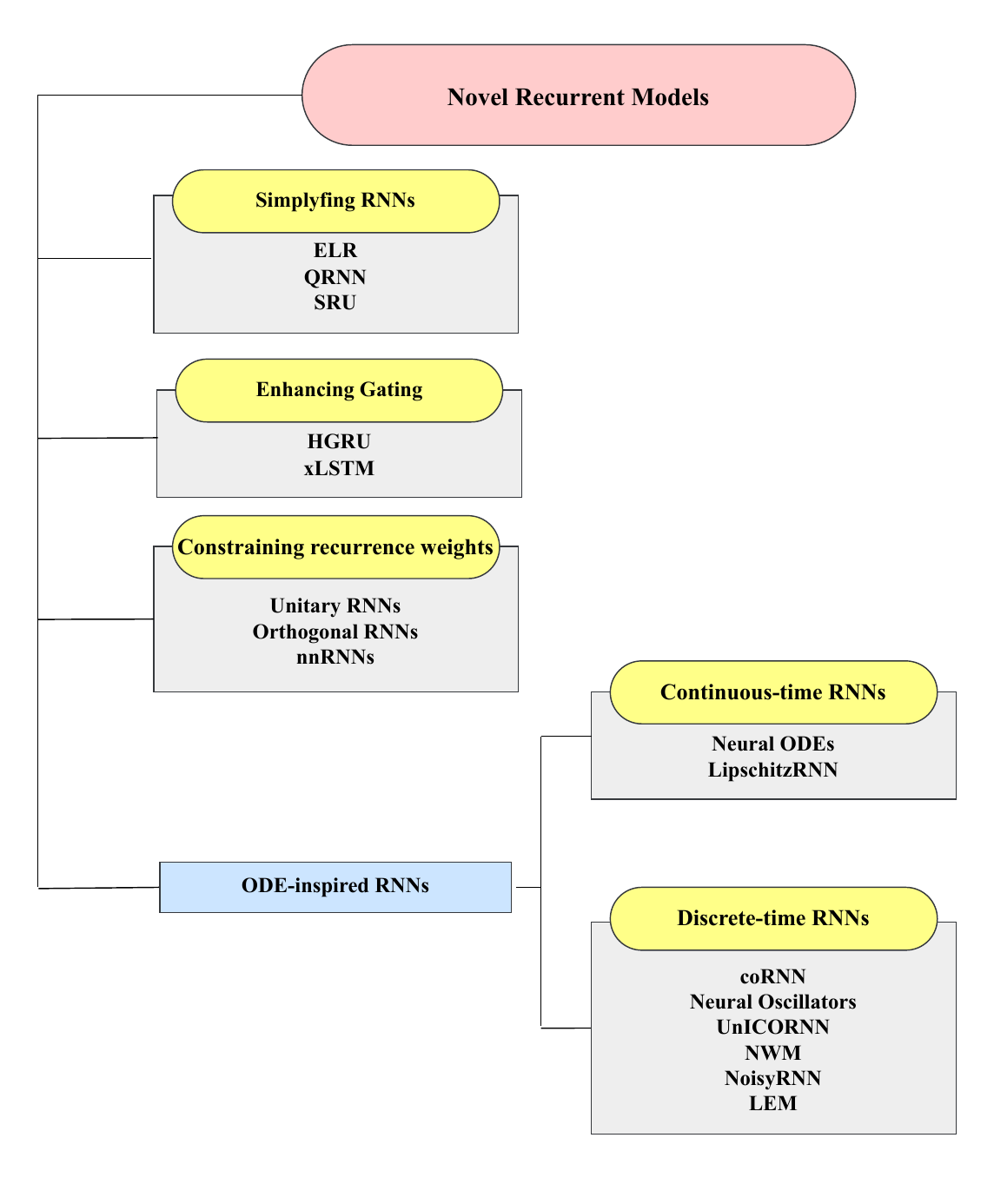}
    \caption{Conceptual overview of the organization of Section \ref{sec:other_rnns}     , which categorizes novel recurrent models based on their architectural innovations and modeling strategies. Blue boxes indicate subsections, yellow boxes denote thematic categories, and representative models are listed in gray.    
    }
    \label{fig:Sec5-guide}
\end{figure}

This section gathers recent approaches that are not directly related to the previous macro-categories of Transformer architectures and State-Space Models, but still focus on improving recurrent models. It turns out that several of the architectural trends described in the previous sections (i.e., element-wise linear recurrence, novel gating mechanisms, etc.), and other new ones, are also significantly explored in the scientific literature that aims to address two of the main drawbacks of \sigla{RNN} , namely slow sequential training and limited capability in modeling long-term dependencies. In Figure~\ref{fig:Sec5-guide}, we report an overview of the main topics/approaches covered by this section.

\parafango{Simplifying RNNs to Gain Speed} \sigla{RNNs} are based on sequential processing of the input data, which does not directly allow for the development of efficient implementations that process the input tokens in parallel or update the components of the hidden state in parallel. It turns out that this limit is mostly due to ($i$) the non-linearity applied to recurrent layers, and ($ii$) the fact that updates in the hidden state are performed by full matrix multiplication, due to a dependency on all components of the hidden state from the previous time step.
In detail, the standard update scheme of \sigla{RNNs} (Eq. \ref{eq:RNN}) assumes that all the neurons in one layer contribute to the state computation of every other neuron (i.e., through the $Ax_{t-1}$ term). Each element of the state vector $x_{t}$ depends on all the entries of $x_{t-1}$.
Early works, such as Independently RNN (\sigla{IndRNN}) \cite{li2018independently}, propose layers composed of ``independent'' neurons, achieved by modifying Eq.
\eqref{eq:RNN} as,
\[
x_t = \sigma(a \odot x_{t-1} + Bu_t),
\]
where the recurrent weight $a \in \R^{\dstate}$ is a vector instead of a matrix, and $\odot$ is the Hadamard (element-wise) product. Notably, the gradient computed by means of \sigla{BPTT}, whose original form is described by Eq. \eqref{eq:product}, factorizes as $ \prod_{s=j}^t a'  \big(\sigma'(x_{s-1})\big)$, thus no matrix multiplications are involved. 
The authors of \sigla{IndRNN} derive upper/lower bound for the recurrent weight values such that \sigla{IndRNN} can tune the preservation or forgetting of long-term memories. 
Neurons in the same \sigla{IndRNN} layer are independent of each other, but cross-channel information over time can be propagated through multiple layers. Remarkably, assuming linear activation, a vanilla \sigla{RNN} with a diagonalizable recurrent weight is a special case of a two-layer \sigla{IndRNN}. In recent literature, recurrent layers with independent neurons and linear activation are referred to as {\it element-wise recurrent} (\sigla{ELR}) layers. 
Quasi-Recurrent neural network (\sigla{QRNN}) \cite{bradbury2016quasi} deal with the inability to parallelize computation in \sigla{RNNs} over the temporal dimension by proposing a mixed architectures which alternates convolutional layers, working simultaneously across different time steps, and  recurrent pooling functions that works in parallel across different channels. 
\sigla{QRNN} alters a classical gated architectures \cite{hochreiter1997long}, replacing the previous hidden state $x_{t-1}$ with the previous input $u_{t-1}$ in the forget gate $f_t$ computation,
$$
f_t = \sigma(W_f^1u_t + W_f^2 u_{t-1}).
$$
This equation can be interpreted as a convolution with kernel-size 2 on the input sequence, an operation that can be computed in parallel along both the temporal and mini-batch dimensions. When considering larger kernel sizes, \sigla{QRNN} performs convolutions over the temporal dimension with a bank of filters,
$$
Z=\text{tanh}(W_z * U), \quad F=\sigma(W_f * U), \quad O= \sigma(W_o * U),
$$
where $W_z, W_f, W_o \in \R^{\dk \times \din \times \dstate}$ are the convolutional filter banks with kernel size $\dk$, and $*$ denotes a causal masked convolution performed along the temporal dimension.
Subsequently, a recurrent pooling operation computes the state, e.g., the \textit{dynamic average pooling} with a single forget gate from \cite{balduzzi2016strongly}, $x_t = f_t \odot x_{t-1} + (1-f_t) \odot z_t$. 
Simple Recurrent Units (\sigla{SRUs}) \cite{lei2018simple} follow the path of element-wise recurrence by substituting all the matrix-multiplications in gates with point-wise multiplications, similarly to \sigla{IndRNN} \cite{li2018independently}. Formally, an \sigla{SRU} makes the cell state $c_t$ independent and parallizable by,
\begin{eqnarray}
\label{eq:forget_sru}
 f_t  &=& \sigma(a_f \odot c_{t-1} + B_f u_t + b_f), \\
 \label{eq:celstate_sru}
 c_t  &=& f_t \odot c_{t-1} + (1-f_t) \odot Bu_t, \\
 \label{eq:reset_sru}
 r_t  &=& \sigma(a_r \odot c_t + B_r u_t +b_r), \\
 \label{eq:state_sru}
 x_t  &=& r_t \odot c_t + (1-r_t) \odot u_t,
\end{eqnarray}
where Eqs. \eqref{eq:forget_sru}-\eqref{eq:celstate_sru} represent the  proposed ``lightweight'' recurrence, where the reset gate $r_t$ adaptively combines the input and the cell state $c_t$, with learnable vectors $a_f, a_r, b_r, b_f$ and learnable matrices $B_f, B_r$. The skip connection in Eq.~\eqref{eq:state_sru} favours gradient propagation. Independence between distinct hidden states enables efficient element-wise product instead of full matrix multiplication (i.e., in classical forget gates dense matrices products inject a dependencies on all neurons previous states), as the authors show when the (nonlinear) recurrence is fused within a single CUDA kernel. This allows to reduce the  complexity to $\mathcal{O}(\dseq b \din)$, where $b$ denotes here the batch dimension, while a standard \sigla{LSTM} takes $\mathcal{O}(\dseq b \din^2)$.
The seminal work by Martin \& Cundy  \cite{martin2018parallelizing} 
highlights that linear recurrences in the form of  $x_t = \Lambda_t x_{t-1} + u_t$ are specific instances of the \textit{scan} operation, a computation involving the repeated application of a binary operator over an array of data. This allows for a highly parallelizable unrolling of the recurrence using parallel scans \cite{blelloch1990prefix}, resulting in substantial improvements in training speeds. When $\Lambda_t$ is diagonal, the cost of a \sigla{ELR} with parallel scan and $p$ processors is $\mathcal{O}\left(6\dstate(\dseq/p + \text{log} p)\right)$, while the cost of serial scan is $\mathcal{O}(2\dstate\dseq)$.  This reduction  becomes important when the sequence length $\dseq$ is large since, given sufficient processors $p$, the parallel time scales logarithmically with the sequence length. Please refer to Appendix H of \cite{smith2022simplified} for a detailed overview of the parallel scan operation. Moreover, several existing models (\sigla{QRNN}\cite{bradbury2016quasi}, \sigla{SRU} \cite{lei2018simple}) fall under this class of approaches, and the authors of \cite{martin2018parallelizing} provide an efficient CUDA kernel that speed up their training. This work laid the foundations for the efficient parallel implementation of \sigla{SSMs} such as \sigla{S5} \cite{smith2022simplified} and others \cite{orvieto2023resurrecting} (Section~\ref{sec:ssm}). Additionally, while typical forget gate values depend on both the previous hidden state and the current input, the authors suggest that forget gate values should depend solely on the current inputs to enable parallel training.
Lately, Lim et al. \cite{lim2023parallelizing} built on top of \cite{martin2018parallelizing} and have shown that it is also possible to parallelize the evaluation and training of \textit{non-linear} sequential models like classic \sigla{RNN} and \sigla{NeuralODE} \cite{chen2018neural}, by introducing a general framework to solve non-linear differential equations, which are restated as fixed-point iteration problems with quadratic convergence, equivalent to Newton’s method. Each fixed-point iteration involves parallelizable operations and an inverse linear operator that can be evaluated in parallel even for the aforementioned sequential models, resulting in improvements of up to $3$ orders of magnitude in terms of speed in the case of long sequences.

\parafango{Enhancing Gating Mechanisms} Learning long-term dependencies requires the ability to modulate the effect of the incoming information. Several recent studies have incorporated gating mechanisms inspired by \sigla{LSTMs} \cite{hochreiter1997long} (or related intuitions) into \sigla{SSMs}, which are characterized by gates acting on linear recurrence layers \cite{gu2020improving,gu2023mamba,smith2022simplified}, yielding impressive performance gains.

Gu et al. \cite{gu2020improving} investigated the saturation regime of gates, remarking the fact that capturing  long-term dependencies in gated \sigla{RNNs} requires forget gate values close to one. Unfortunately, learning with gates in their saturated regimes (i.e., values close to $0$ or $1$) is difficult, since they suffer from vanishing gradients. Moreover, if all forget gate values are close to one, the model’s ability to forget irrelevant information is hindered. To overcome such issues, the authors of \cite{gu2020improving} propose to tweak the forget gate $f_t$ with an independent \textit{refine gate} $r_t$ that is exploited to produce an input-dependent bounded additive term $\phi(f_t, u_t)$ that modulates $f_t$, allowing much higher/lower activations,

\begin{equation*}
\begin{aligned}
    r_t &= \sigma(W_r^1u_t + W_r^2 x_{t-1}), \\
     g_t &= f_t + \phi(f_t, u) \defeq  r_t \left(1-(1-f_t)^2\right) + (1-r_t)f_t^2, \\
     c_t &= g_tc_{t-1} + (1-g_t)\hat{c}_t,
\end{aligned}
\end{equation*}
where $W_r^*$ denote projection matrices and $\hat{c}_t$ is the cell input activation vector from \sigla{LSTMs}.
The form of the additive update $\phi(f_t, u)$ emerges from the consideration that gating mechanisms should be bounded in $[0,1]$, symmetric around $0$, and differentiable. Additionally, the authors propose to initialize the forget gate $f_t$ with values sampled from the uniform distribution $\mathcal{U}(0,1)$, instead of constant values, even allowing negative biases. This choice fosters the gate’s ability to grasp different timescales, as noticed in the \textit{chrono initialization} approach by Tallec \& Olivier \cite{tallec2018can}.

In previous Sections, we showed how several linear \sigla{RNNs} with static decay rates perform
eigendecompositions on the recurrent weight matrix to achieve element-wise linear recurrence \cite{huang2022encoding,orvieto2023resurrecting}. Notice that, if only real-valued eigenvalues are allowed, this choice restricts the range of the recurrent weight matrix to {be symmetric, limiting the expressiveness of the model. To overcome this limitation, linear
\sigla{RNNs} often employ complex-valued eigenvalues, \cite{gu2021combining,orvieto2023resurrecting}. Following such intuitions, Hierarchically Gated Recurrent Units (\sigla{HGRU}) \cite{qin2023hierarchically} exploit linear recurrence in the complex domain, and address the saturating issue pointed out by Gu et al. \cite{gu2020improving} by adding an additive learnable value $\Gamma$ to the original forget gate, with the purpose of pushing gate activations away from the saturated regimes. 
The $\Gamma$ variable, which acts as a lower bound on forget gate values, is forced to increase monotonically with the model depth, inspired from Ordered Neuron LSTM \cite{shen2018ordered}: small value in lower layers, in order to ease the forgetting of past-information (short-term dependencies); forget value close to one in top-most layers, facilitating the modeling of long-term dependencies.
In detail, \sigla{HGRU} leverages a gated linear recurrent as follows,

\begin{equation*}
\begin{aligned}
    \text{Re}(c_t) &= \text{SiLU}(u_tW_{re}), \quad  \text{Im}(c_t) = \text{SiLU}(u_tW_{im}),\\
    f_t &= \lambda_t \odot e^{i\theta}, \\
    x_t &= f_t \odot x_{t-1} + (1-\lambda_t) \odot c_t,     
\end{aligned}
\end{equation*}
where the real ($\text{Re}$) and imaginary ($\text{Im}$) part of $c_t$ are parametrized separately by means of learnable projections $W_{\cdot}$, $\text{SiLU}$ is the Sigmoid Linear Unit function \cite{hendrycks2016gaussian},\footnote{It is implemented as $\text{SiLu}(x) = x\sigma(x)$, where $\sigma$ is the sigmoid function.} and $i$ is the imaginary unit. 
Inspired by recent works that achieve \sigla{ELR} by eigendecomposition of the recurrent matrix $A$ \cite{gu2021combining,orvieto2023resurrecting}, both the state $x_t$ and the input mapping of \sigla{HGRU} 
are complex vectors, i.e., $x_t, c_t \in \mathcal{C}^{1 \times \dstate}$, to enhance the model expressive power, as previously introduced.
The magnitude $\lambda_t$ of the forget gate $f_t$ regulates the ability to retain previous information, while the phase argument $\theta$, which is shared among time steps, determines the oscillation frequencies, in a data-independent manner. The aforementioned layer-wise increment of the additive lower bound $\Gamma$ on the forget gate is achieved by acting on $\lambda_t$, as follows, assuming $l$ is the layer index ($H$ layers),
\begin{equation*}
    \begin{aligned}
     \gamma^l &=[\text{cumsum}(\text{softmax}(\Gamma))]_l,\\ 
      \mu_t &= \sigma ( B_\mu u_t), \\
    \lambda_{t}^l&= \gamma^l + (1 - \gamma^l) \odot \mu_t,
    \end{aligned}
\end{equation*}
where $\Gamma \in \R^{H \times \dstate}$ independently parametrizes the lower bounds for all hidden states, 
where $\text{cumsum}$ and $\text{softmax}$ operate over the first dimension of their tensor input. Basically, the composition of the the softmax and $\text{cumsum}$ functions forms an activation function which yields a monotonically increasing vector in $[0,1]$. The squared bracket-based notation is defined as $[\text{cumsum}(x)]_l = (\sum_{i=1}^l x_i) - x_1$. Notice that $\text{cumsum}_0$ is applied to the layer dimension across different layers, to enable upper layers to model long-range dependencies. Then, the model exploits an output projection with a learned gate, similarly to what happens in \sigla{SSMs}. 
Another perspective on the role of gating mechanisms can be appreciated by inspecting the already described connection between the update mechanisms in linear \sigla{RNNs} and linear Transformers, discussed in previous Sections. Recently, Zucchet et al. \cite{zucchet2023gated} proposed a unifying view on such architectures, driven by the role of gating functions. In particular, under a specific parametrization (that leverages \sigla{GLU} \cite{dauphin2017language} and requires a number of parameters squared with respect to attention parameters), they 
showed that \sigla{RNNs} equipped with linear recurrent layers interconnected by feed-forward paths with multiplicative gating can, through learning, encode attention-based algorithms disguised in their
weights.

\parafango{Enhancing Gating Mechanisms (variants of LSTMs)} The foundations of gating mechanisms, laid out by the seminal \sigla{LSTM} paper \cite{hochreiter1997long}, were recently revised, proposing two modern variants \cite{beck2024xlstm}, referred to as \sigla{sLSTM} and \sigla{mLSTM} that, when plugged into residual backbones, yield what is referred to as \sigla{xLSTM} architecture. The main goal is to scale \sigla{LSTMs}  to billions parameters to exploit them in large language models, by injecting some of the techniques we described in previous Sections, such as exponential gating and matrix-based states. 
For reference, we summarize in the following the cell update rules underlying vanilla \sigla{LSTM},\footnote{Notice that the original model is characterized by a scalar memory cell, i.e., $c_t \in \R$, as processing and storage unit. Later formulations \cite{greff2016lstm} combined multiple memory cells into a vector $c_t \in \R^h$, where $h$ is the number of cell units. In the main text of this paper, we exploit the latter vectorial variant.}
\begin{equation}
    \begin{aligned}
        c_t &= f_t \odot c_{t-1} + i_t \odot z_t, \\
        h_t &= o_t \odot \psi(h_t), \\ 
        z_t &= \sigma_z(W_z u_t + R_z h_{t-1}  + b_z),\\
        i_t &= \sigma_i(W_i u_t + R_i h_{t-1}  + b_i),\\
        f_t &= \sigma_f(W_f u_t + R_f h_{t-1}  + b_f),\\
        o_t &= \sigma_o(W_o u_t + R_o h_{t-1}  + b_o),
    \end{aligned}
    \label{vaffanculoalmondo}
\end{equation}
where $z$ denote the cell input, $i$ the input gate, $f$ the forget gate and $o$ output gates; $W_z$, $W_i$, $W_f$, $W_o$  denote the corresponding learnable matrices connecting input $u_t$ to the gates. $R_z$, $R_i$, $R_f$ and $R_o$ are the corresponding learnable recurrent weights on the hidden states and 
$b_z$, $b_i$, $b_f$, $b_o$ are
learnable biases; $\psi$ normalizes or squashes the cell state, and is typically a $\text{tanh}(\cdot)$, as long as the cell input activation function $\sigma_z$. Notice that the usage of recurrent weight matrices $R_z$, $R_i$, $R_f$, $R_o$ allows to mix memory cells outputs. The activations $\sigma_z$, $\sigma_i$, $\sigma_f$, $\sigma_o$ on gates $i,f,g$ 
are typically sigmoids. 
The \sigla{sLSTM} variant introduces exponential gating on input and forget gates, in order to allow the model to better revise decisions on what to ``store''. Moreover, it introduces a normalizer state $n_t$ to better stabilize the model dynamics. The first two \sigla{LSTM}-equations of Eq.~\ref{vaffanculoalmondo} are replaced by the following three ones,
\begin{equation*}
    \begin{aligned}
        c_t &= f_t \odot c_{t-1} + i_t \odot z_t, \\
        n_t &= f_t \odot n_{t-1} + i_t, \\
        h_t &= o_t \odot \big(c_t \odot n_t^{-1}).
    \end{aligned}
\end{equation*}
The \sigla{LSTM} activations in Eq.~\ref{vaffanculoalmondo} are implemented following specific choices: $\sigma_z \defeq \text{tanh} (\cdot)$ to help stabilizing the recurrence, $\sigma_i, \sigma_f \defeq \text{exp}(\cdot)$,  $\sigma_o \defeq \text{sigmoid}(\cdot)$.
Given that the presence of exponential function could led to large values and numerical issues, the authors further stabilize gates with an additional state (please refer to the referenced paper).   Additionally, \sigla{sLSTMs} leverage multiple memory heads, where memory mixing happens within each head (via the last introduced equations) but not across heads.
As previously stated, when considering each cell in an \sigla{LSTM} or \sigla{sLSTM}, quantities (cell state, gates) are scalar (i.e.,  $c_t, f_t, i_t, o_t \in \R$), and  multi-dimensionality is gained by considering $h$ cells (i.e., $c_t \in \R^h$). The second model variant of \cite{beck2024xlstm}, \sigla{mLSTM}, enhances the vanilla model's storage scalar capacity by introducing a matrix-based cell state, $C_t \in \R^{\dstate \times \dstate}$, whose update is regulated by an outer product rule akin to Linear Transformers or Fast Weight Programmers \cite{schlag2021linear} (see Section \ref{sec:transformers}). Hence, it leverages a key, query, value projections of the input as follows,
\begin{equation*}
    \begin{aligned}
        C_t &= f_t C_{t-1} + i_t (v_t \otimes k_t), \quad 
        n_t = f_t n_{t-1} + i_t k_t, \\
        h_t &= o_t \odot \frac{C_t q_t}{\text{max}({n_t^T q_t, 1})},\quad 
        \ \ \ \ i_t = \sigma_i(w_i' u_t + b_i),\\
        f_t &= \sigma_f(w_f' u_t + b_f), \quad 
        \ \ \ \ \ \ \ \ \ \ o_t = \sigma(W_o u_t + b_o),       
    \end{aligned}
\end{equation*}
where $q_t,k_t,v_t \in \R^\dk$ are linear projections of the input (such as in self-attention), while $w_i,w_f \in \R^{\din}$ are learnable weight vectors.  
In the cell-state update rule, the forget gate acts as a decay rate, the input gate as a learning rate, while the output gate scales the vector which is retrieved by the outer product. 
The normalizer state $n_t$, which keeps a record of the ``strength'' of the gates, is the weighted sum of the key vectors, where, in turn, each key is weighted by the input gate and the current 
forget gate. 
In \sigla{mLSTM}, considering multiple cells is equivalent to multiple heads, since in this case there is no memory mixing, due to the presence of matrix states. Interestingly, the absence of memory mixing (no hidden-to-hidden connections, hence each cell/head can be computed independently of the others) allows us to reformulate \sigla{mLSTM} in a parallel form, in order to speed up training when the full sequence is available in advance (see Appendix A.3 of  \cite{beck2024xlstm} for further details). 
When composed into an architecture of stacked blocks, connected via residual connections of two types (with post-up projections when considering \sigla{sLSTM}---like Transformers---or with pre-up projections when considering \sigla{mLSTM}), the model is referred to as an eXtended LSTM (\sigla{xLSTM}). Overall,  \sigla{xLSTM} have a $\mathcal{O}(\dseq)$ computational  and $\mathcal{O}(1)$ memory complexities. Additionally, \sigla{mLSTMs}, despite being computationally expensive due to the presence of matrix-based state, implements a  parallel-computation form, while \sigla{sLSTM} is not parallelizable due to memory mixing.

\parafango{Constraining the Recurrence Weights} Exploding and vanishing gradients hamper the \sigla{RNNs}' ability to learn long-term dependencies. A recent strategy to circumvent this issue and allow the stable propagation of signals over long time scales, is to constrain the hidden-to-hidden weight matrix to be orthogonal or unitary (i.e., an element of the orthogonal group---referred to as \sigla{Unitary} and \sigla{Orthogonal RNNs}---see \cite{arjovsky2016unitary} and references therein), which ensures that the eigenvalues have unit norm and the dynamics are stable. However, despite advantages in terms of long-term memory preservation, this also reduces the expressivity of the model, as orthogonal transformations are limited in variety. We draw readers' attention to \cite{salehinejad2017recent} (survey) and the extensive descriptions of several recent works, e.g., \cite{erichson2020lipschitz,lezcano2019cheap}. Some recent works have proposed alternative ways to overcome this trade-off, such as using non-normal matrices with unit norm eigenvalues without orthogonality constraints on eigenbases (\sigla{nnRNN} \cite{kerg2019non}), or formulating the recurrent units by differential equations and updating the hidden states exploiting the difference between state values \cite{kag2019rnns} These methods aim to improve the performance and flexibility of \sigla{RNNs} while preserving their long-term memory, without explicit constraints on the weight matrices.

\subsection{ODE-inspired  Recurrent Neural Networks}
\label{sec:odernn}
A recent trend involves recurrent architectures whose processing scheme is formalized by Ordinary Differential Equations (ODEs) in the context of dynamical systems.
Two main branches of scientific works have developed, the former based on continuous-time \sigla{RNNs} and the latter on discretized ODEs.

\parafango{Continuous-time RNNs} Continuous-time recurrent networks have been the subject of investigation since the dawn of neural networks and, later on, they were deeply investigated at the intersection of machine learning and other scientific fields, such as signal processing~\cite{Mandic2001}. Amongst others, more recently, the interest in continuous-time \sigla{RNNs} has been renewed by studies on  \sigla{Neural ODEs} \cite{chen2018neural} and  \sigla{ODE RNNs} \cite{rubanova2019latent}, where a continuous ODE acts as the learning model and gradients are computed from a sensitivity equation, which allows one to trade accuracy with computing time.
The state of a \sigla{Neural ODEs} is defined by the solutions of the equation $\dot{x} = f(x, u, t, \theta)$, where  $t$ represents continuous time, $u \defeq u(t) \in \R^\din$ the time-dependent input signal, $x \defeq x(t) \in \R^\dstate$ the RNN hidden state, $\dot{x}$ its first order time derivative and $f$ a neural network parametrized by $\theta$. Readers can find further details in \cite{kidger2022neural} and references therein. \sigla{Liquid Time-constant Networks} \cite{hasani2021liquid}, rather than defining the derivatives of the hidden-state directly by a neural network $f$ as in \sigla{Neural ODE}, determine a more stable continuous-time RNN in the form,
\begin{equation*}
    \dot{x} = - \big( A + B \odot f(x, u, t, \theta)  \big) \odot x + B\odot  f(x, u, t, \theta),
\end{equation*}
where $A \in \R^\dstate$ is a time-constant state-transition mechanism and $B \in \R^\dstate$ a bias vector. Thanks to this computational structure, the neural network $f$  determines both the derivative of the hidden state $x(t)$ and serves as an input-dependent varying time-step (i.e., dynamical, hence the term \textit{liquid}) for the learning system.\footnote{Time-step $\tau_{\text{sys}}$ is a parameter characterizing the speed and the coupling sensitivity of an ODE. In this case, $\tau_{\text{sys}} =\frac{\tau}{ 1 + \tau f(x,u,t,\theta)}$.}
Hasani et al. \cite{hasani2022closed}  computed an approximation of the solution of the integral appearing in liquid time-constant dynamics, relaxing the need for complex numerical solvers.
\sigla{LipschitzRNNs} \cite{erichson2020lipschitz} describe the evolution of the hidden state exploiting a functional form composed by a linear component plus a 1-Lipschitz nonlinearity (i.e., the $\text{tanh}$),
\begin{equation*}
    \begin{aligned}
        \dot{x} = \bar{A}x + \text{tanh}(\bar{W}h + Bu + b), \qquad y = Dx,
    \end{aligned}
\end{equation*}
where $\bar{A}, \bar{W}$ are tunable matrices with an ad-hoc fixed structure. 
By leveraging tools from nonlinear systems theory, the authors carried out a stability analysis on the proposed recurrent unit behaviour in the long-term, resulting in good performance and expressivity.
Recently, an in-depth analysis of approximation properties and optimization dynamics
of continuous-time \sigla{RNNs} has been carried out in \cite{li2020curse,li2022approximation}, with an interesting take on the interaction of memory and recurrent structures in the linear dynamical setting.

\parafango{Discrete-time RNNs} Coupled Oscillatory RNNs (\sigla{coRNN}) \cite{rusch2020coupled} leverage coupled networks of controlled non-linear forces and damped oscillators, underlying several physical systems and also in biological neurons, to ensure both expressive representations and the preservation of long-term dependencies, while constraining the dynamics of state variables and their gradients.  The model is formulated through implicit-explicit time-discretizations of second-order nonlinear ordinary differential equations, capturing the dynamics of coupled oscillators in continuous time,
\begin{equation*}
    \ddot{x} = \text{tanh}(Wx + \hat{W}\dot{x} + Vu + b) - \gamma x -\epsilon \dot{x},
    \label{eq:cornn}
\end{equation*}
where  $t \in [0, 1]$ is the (continuous) time variable, $u \defeq u(t) \in \R^\din$ the time-dependent input signal, $x \defeq x(t) \in \R^\dstate$ the RNN hidden state RNN, and $\dot{x}, \ddot{x}$ its first and second order time derivatives;  $W$, $\hat{W} \in \R^{\dstate \times \dstate}$, $V \in \R^{\dstate \times \din}$ are weight matrices, $b \in \R^\dstate$  is the bias vector and $\gamma, \epsilon > 0$,  are parameters representing oscillation frequency and the amount of damping (friction) in the system, respectively.  
By introducing the \textit{velocity} variable $z \defeq \dot{x}$ it is possible to obtain a first order system of two coupled networks defined as follows,
\begin{equation}
    \begin{aligned}
        \dot{x} &= z, \\ 
        \dot{z} &= \sigma(Wx + \hat{W}z + Vu + b) - \gamma x -\epsilon z.
        \label{eq:cornn_coupled}
    \end{aligned}
\end{equation}
When discretizing such system with a fixed timestep $0< \Delta t < 1$, the RNN hidden state at time $t_n = n\Delta t \in [0,1]$ evolves accordingly to the following laws,
\begin{equation*}
    \begin{aligned}
        x_n &= x_{n-1} + \Delta t z_n, \\
        z_n &= z_{n-1} +  \Delta t \sigma(Wx_{n-1} + \mathcal{W}z_{n-1} + Vu_n + b) \\ 
       &  \hskip 0.5cm - \Delta t\gamma x_{n-1} - \Delta t \epsilon z_{\bar{n}},
    \end{aligned}
\end{equation*}
with $\bar{n}$ either $\bar{n}=n$ or $\bar{n}=n-1$, depending on the fact that the damping term $\epsilon z$ is treated implicitly (the former case) or explicitly (the latter). 
In the coupled networks defined by Eq. \eqref{eq:cornn_coupled}, each neuron updates its hidden state based on input signals and information from other neurons.The diagonal entries of $W$ and the hyperparameter $\gamma$ control oscillation frequency, while the diagonal entries of $\hat{W}$ and the hyperparameter $\epsilon$ determine damping for each neuron, whereas non-diagonal entries modulate interactions between neurons. Input signals drive the generation of (superpositions of) oscillatory wave-forms, controlled by the other tunable parameters. 
This leads to rich global dynamics and the emergence of non-trivial non-oscillatory hidden states from oscillatory inputs, emphasizing the network's high expressivity in approximating outputs from complex sequential inputs. The authors derive bounded gradients and limited hidden state magnitude for the \sigla{coRNN} model, under some mild assumptions. Thus,  \sigla{coRNN} has stable dynamics which foster better performance than existing
\sigla{RNNs}, especially on tasks with very long time-dependencies. 
In this family of models, which is recently referred to as \sigla{Neural Oscillators} \cite{lanthaler2023neural}, lies \sigla{UnICORNN} \cite{rusch2021unicornn}, a multi-layer sequence model that stacks networks of independent (uncoupled) undamped oscillators as hidden layers within an \sigla{RNN}. In contrast to \sigla{coRNN}, neurons in
\sigla{UnICORNN} are independent (uncoupled) and as there is no
damping, the ODE system yielding \sigla{UnICORNN} has a Hamiltonian structure. This characteristic allows the model to avoid any assumptions on the weights,  whereas the mitigation of exploding/vanishing gradients in  \sigla{coRNN} was {dependent }  on specific restrictions imposed on the weights.
Moreover, \sigla{Neural Oscillators} have been proven to be capable of approximating
any continuous and causal operator mapping between time-varying functions, to
desired accuracy \cite{lanthaler2023neural}.
 Their performances on long-range sequences are remarkable \cite{rusch2021unicornn}. 
Locally coupled
oscillatory recurrent neural networks have been used to model the neuroscience concept of traveling waves, referred to as Neural Wave Machines (\sigla{NWMs}) \cite{keller2023neural}. 
Such waves serve as a bias towards learning structured representations, which exhibit complex spatio-temporal dynamics when modeling real data. When tasked to reconstruct the input signal, \sigla{NWMs} use traveling waves
to encode transformations in the RNN hidden state. 
Waves-like dynamics can be modeled also with simpler RNN architectures though connectivity constraints and
initialization \cite{keller2023traveling}, and can act as memory storage system on complex sequence modeling. 
\sigla{NoisyRNN} \cite{limnoisyrnn} consider discretizations of the stochastic differential equations (SDEs) obtained from ODE formulations of \sigla{RNNs} through the addition of a diffusion (noise) term, as an implicit regularization. By dropping the noisy elements at inference time, \sigla{NoisyRNN} can be considered as a stochastic learning strategy (i.e., similarly to Dropout) with several advantages such as more stable dynamics. This introduces a form of implicit regularization leading towards the development of classifiers with a large classification margin, that keep generalization error small. However, despite the stabilization properties, noise injection could negatively impact capacity for long-term memory. 
Rusch et al. \cite{rusch2021long} pointed out that real-world data could contain information arranged according to multiple scales, i.e., time, lengths etc., depending on the considered data and task. They propose Long Expressive Memory (\sigla{LEM}), based on a time-discretization of a set of multiscale ODEs.  These scales can be
learned adaptively (with respect to states) and dynamically (in time). \sigla{LEM} has bounded gradients that mitigate exploding/vanishing issues, and favour the model ability in the context of long sequence processing. 
Irie et al. \cite{irie2022neural} introduced learning rules and \sigla{Neural ODEs} to build continuous-time sequence processing nets that learn to manipulate short-term memory in rapidly changing synaptic connections of other nets. This yields continuous-time counterparts of Fast Weight Programmers and Linear Transformers \cite{irie2021going,irie2022modern}. Learning rules can be seen as the outcome of discretization procedures applied to ODEs.
Kag et al. \cite{kag2019rnns} proposed a modified differential equation for the state update, obtained by leveraging implicit discretization methods rather than  the (explicit) Euler method, to foster  system stability. The main intuition is that the hidden states are updated based on the difference between predicted and previous states. Then, the implicit equation is solved via fixed-point recursion, resulting in stable fixed points and fast convergence. 
In a subsequent work, Kag et al. \cite{kag2021time} also explored a time-adaptive discretization of the ODE where time steps are modified based on the current observation and the hidden state.

{This section catalogued architectural innovations that aim to simplify, stabilise or extend classic \sigla{RNNs}: independent/diagonal updates for parallelism, unitary or orthogonal weights for gradient preservation, enhanced gating (e.g., xLSTM), and continuous-time formulations such as \sigla{Neural ODEs}. In the next Section~\ref{sec:forward}, we summarize recently introduced efficient and forward-in-time alternatives to \sigla{BPTT} used to train recurrent models.}


\section{Learning in Recurrent Models}
\label{sec:forward}

\begin{figure}
    \centering
    \includegraphics[width=0.7\linewidth]{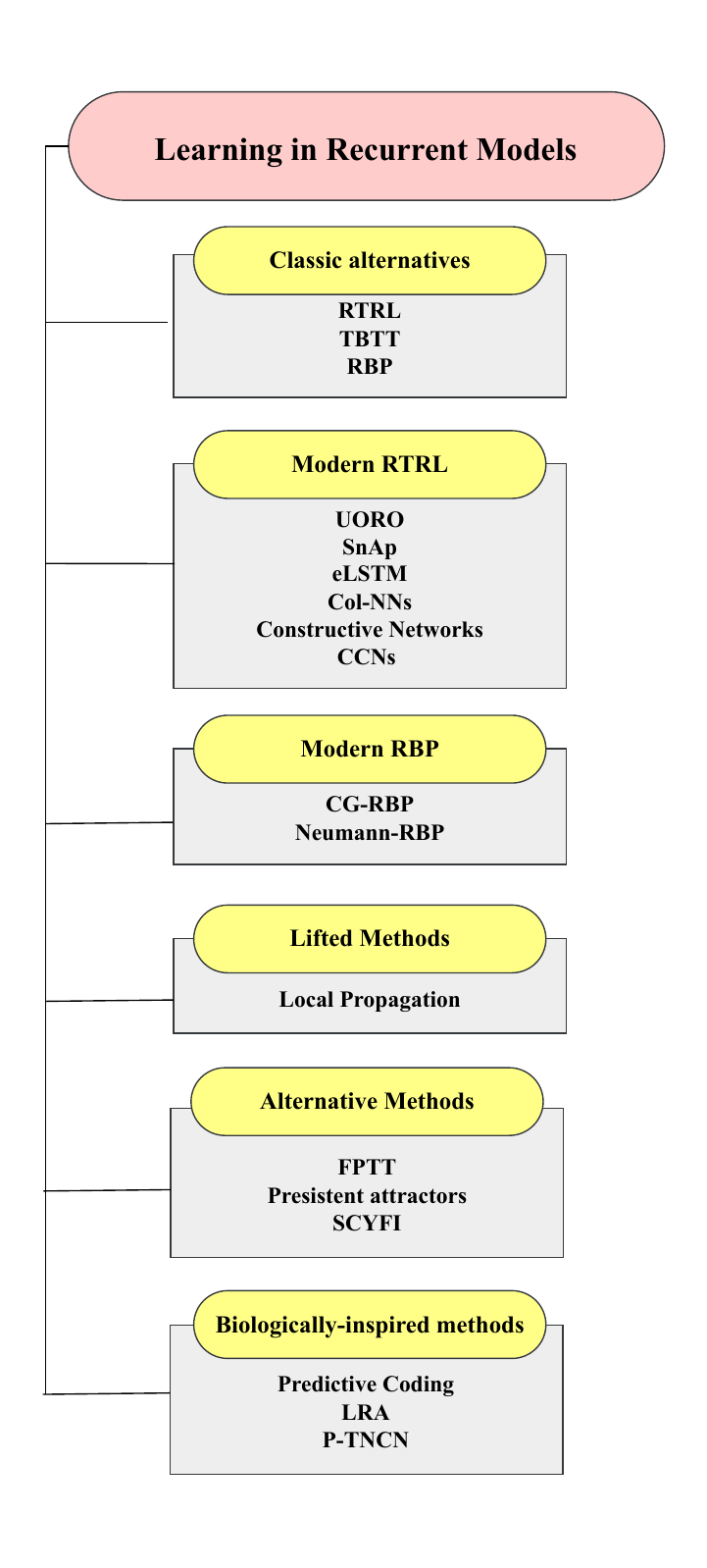}
    \caption{Conceptual overview of the organization of Section \ref{sec:forward}, which classifies learning approaches for recurrent models based on methodological principles. Yellow boxes denote thematic categories, and representative methods are listed in gray.    
    }
    \label{fig:Sec6-guide}
\end{figure}

Backpropagation Through Time (\sigla{BPTT}, see Section \ref{sec:related}) is the de-facto standard algorithm for training recurrent models. It involves unrolling (i.e., virtually replicating) a recurrent network over the whole input sequence of length $L$, sharing the same parameters $L$ times, and ``backpropagating'' the error from the $L$-th instance to the first one. In Section \ref{sec:related}, we have emphasized the advantages and drawbacks of \sigla{BPTT}, such as ($i$) learning issues due to vanishing/exploding gradients and ($ii$) the high memory and computational requirements. These requirements hinder the ability to handle long-range sequences. Indeed, whenever data come in the form of a \textit{stream} \cite{gunasekara2023survey}, the \sigla{BPTT} requirements make it a non-feasible choice for online learning on a potentially infinite sequence, given the difficulties in unrolling the network over long time horizons. Most of the models described in previous sections propose to alleviate such issues by careful architectural designs, but still retain \sigla{BPTT} as the learning procedure, and assume that the whole sequence is available beforehand.

In this section, we overview recent alternative learning mechanisms that aim to address such drawbacks ($i$ and $ii$) of \sigla{BPTT}.

We leverage a more general form of the recurrent mechanism described in Eq. \eqref{eq:RNN}, explicitly including a state transition function $F(\cdot)$ and an output readout function $G(\cdot)$, defined as follows. This general form simplifies the notation in the subsequent descriptions of the reviewed approaches.

\begin{equation}
    x_t = F(x_{t-1}, u_t, \theta^F), \qquad y_t = G(x_{t}, u_t, \theta^G),
    \label{eq:rnn_transition}
\end{equation}
where, referring to Eq. \eqref{eq:RNN},  $\theta^F := [A, B]$ and $\theta^G := [C, D]$. Before diving into the specific details of the reviewed approaches, we showcase in Figure~\ref{fig:Sec6-guide} the organization of this section.

\parafango{Classic Alternatives} Eq.~\eqref{eq:bptt} shows how \sigla{BPTT} requires to cache the neuron activations for every element within the sequence processed by the model, in order to be able to perform gradient computation in the backward stage. The amount of past activations to be stored grows linearly with the sequence length, hindering \sigla{BPTT} usage with very long sequences.
Truncated BPTT (\sigla{TBPTT})\cite{williams1990efficient} limits the gradient flow after a fixed number of time steps. While this make learning tractable in longer sequences, it inherits the structural inability to capture dependencies beyond the designated time window. 
In contrast, Real-time Recurrent Learning (\sigla{RTRL}) \cite{williams1989learning} does not require the storage of past activations, and it was proposed as an \textit{online} alternative to \sigla{BPTT}, which enables weight updates promptly after each new input is processed, provided that external error feedback to the model output is accessible for each input. For every timestep $t$, one can define the \textit{influence} (or \textit{sensitivity}) matrix $M_t \in \R^{\dstate \times |\theta_F|}$ as, 
\begin{equation}
    M_t = \frac{\partial x_t}{\partial \theta^{F}},
    \label{eq:inference_matrix}
\end{equation}
which contains the derivatives of the current state $x_t$ with respect to the parameters $\theta^F$ of the transition function $F$ in Eq. \eqref{eq:rnn_transition}. It is possible to prove the following recurrent formula to update $M_t$ over time,
\begin{equation}
    \begin{aligned}
        M_t &= \sum_{s \leq t} \frac{\partial x_t}{\partial \theta^{F}_s} = \sum_{s \leq t-1} \frac{\partial x_t}{\partial \theta^{F}_s} + \frac{\partial x_t}{\partial \theta^{F}_t} \\
        &= \sum_{s \leq t-1} \frac{\partial x_t}{\partial x_{t-1}} \frac{\partial x_{t-1}}{\partial \theta^{F}_s} + \frac{\partial x_t}{\partial \theta^{F}_t} \\
        &= \frac{\partial x_t}{\partial x_{t-1}} \frac{\partial x_{t-1}}{\partial \theta^{F}} + \frac{\partial x_t}{\partial \theta^{F}_t} \\
        &= J_t M_{t-1} + \bar M_t.
    \end{aligned}
    \label{eq:inference_matrix_rec}
\end{equation}
{We now recall that the \emph{Jacobian matrix} of a vector-valued function is a matrix of all the first-order partial derivatives of the function, w.r.t. its arguments. More precisely, let
\begin{equation}
f : \mathbb{R}^n \rightarrow \mathbb{R}^m, \quad f(x) = 
\begin{bmatrix}
f_1(x_1, x_2, \dots, x_n) \\
f_2(x_1, x_2, \dots, x_n) \\
\vdots \\
f_m(x_1, x_2, \dots, x_n)
\end{bmatrix}.
\end{equation}
Then, the Jacobian matrix of $f$ with respect to $x$ is defined as:
\begin{equation}
J_{f}(x) = 
\begin{bmatrix}
\frac{\partial f_1}{\partial x_1} & \frac{\partial f_1}{\partial x_2} & \cdots & \frac{\partial f_1}{\partial x_n} \\
\frac{\partial f_2}{\partial x_1} & \frac{\partial f_2}{\partial x_2} & \cdots & \frac{\partial f_2}{\partial x_n} \\
\vdots & \vdots & \ddots & \vdots \\
\frac{\partial f_m}{\partial x_1} & \frac{\partial f_m}{\partial x_2} & \cdots & \frac{\partial f_m}{\partial x_n}
\end{bmatrix}.
\end{equation}}
{In Eq.~\eqref{eq:inference_matrix_rec},} $J_t = \frac{\partial x_t}{\partial x_{t-1}}$ is the Jacobian of the actual state w.r.t. the previous state and $\bar M_t = \frac{\partial x_t}{\partial \theta^{F}_t}$ is called the \textit{immediate influence}. Notice the distinction between $\theta^F_s$ and $\theta^F_t$: the former is about instances of the weights in the previous time instants, while the latter is about the current weight values. It is possible to obtain the derivatives of the loss function with respect to $\theta^{F}$ by,
\begin{equation}
    \frac{\partial \ell_t}{\partial \theta^F} = \frac{\partial \ell_t}{\partial x_t} \frac{\partial x_t}{\partial \theta^F} = \bar c_t M_t,
\end{equation}
where $\bar c_t = \frac{\partial \ell_t}{\partial x_t}$ is called the \textit{immediate credit assignment vector}. Differently form the just described case of $\theta^F$, it is natural to directly learn $\theta^G$ online, because only information at present time $t$
is required to calculate the gradient $\frac{\partial \ell_t}{\partial \theta^G}$. \sigla{RTRL} suggests to propagate the partials $\frac{\partial x_t}{\partial x_{t-1}}$ and $\frac{\partial x_t}{\partial \theta^F}$ from timestep $t$ to $t + 1$. This is based on the intuition that there is significant overlap in the product term (see Eq.~\eqref{eq:bptt} and Eq.~\eqref{eq:product}) from time $t$ to $t + 1$, allowing for recursive computation. This algorithm is online and past-facing, which is defined by the fact that only previously computed quantities are used in the computations \cite{marschall2020unified}. These properties can be directly observed in the above formulas, which explicitly depend only on timesteps $t$ and $t - 1$. \sigla{RTRL} is also deterministic and provides the solution in closed form. Since \sigla{RTRL} requires, for each time step, the storage of $M_t$, which involves the gradients of each component of the state ($\dstate$ elements) with respect to all the parameters involved in the state computation (i.e., $|\theta^{F}| = \dstate \cdot \din + \dstate^2$), its memory complexity is $\mathcal{O}(\dstate(\dstate \cdot \din + \dstate^2)) \sim \mathcal{O}(\dstate^{3})$ and, for the computation of $J_t M_{t-1}$, its time complexity is $\mathcal{O}(\dstate^{4})$. Early attempts suffered large memory overhead limiting its usage, and while recent attempts \cite{tallec2017unbiased,marschall2020unified} have been more successful, these methods still fall short of \sigla{BPTT} performance, and so trainability of \sigla{RNNs} is still a significant issue.
Several other early attempts to solve the shortcomings of \sigla{BPTT} were motivated by the human way to learn from perceptual stimuli, which are intrinsically continuous over time and not pre-buffered finite-length sequences randomly shuffled to cope with stochastic gradient descent \cite{tiezzi2020focus}. Thus, from \cite{williams1990efficient}, the proposal of  {\it ``an on-line algorithm, designed to be used to train a network while it runs; no manual state resets or segmentations of the training stream is required''}. Even the \sigla{LSTMs} \cite{hochreiter1997long} were introduced with a learning algorithm that unlinke full \sigla{BPTT} is {\it ``local in space and time''}, where {\it ``there is no need to store activation values observed during sequence processing in a stack with potentially unlimited size''}.
Recurrent Backpropagation (\sigla{RBP}) \cite{almeida1990learning,pineda1987generalization,DBLP:journals/tnn/Pearlmutter95} avoids the need to unroll the entire forward pass, as required by \sigla{BPTT}, by directly computing the gradient w.r.t. the learnable parameters at a steady state of Eq. \eqref{eq:rnn_transition}, exploiting the implicit function theorem \cite{rudin1976principles} and achieving constant memory complexity w.r.t. the number of processing steps. In details, 
\sigla{RBP} assumes that the dynamics of the state transition $F(\cdot)$ reach an equilibrium with a steady-state hidden state $x^*$, i.e., $x^* = F(x^*, u, \theta^F)$ when processing an input $u$, which is fixed and not time-dependent.\footnote{\sigla{RBP} formulation assumes a fixed not time-dependent input $u$. However, in common scenarios where data is i.i.d, the sequential input data can be interpreted as sampled from a stationary distribution. As a consequences, \sigla{RBP} can be applied, since the steady state holds in expectation\cite{liao2018reviving}.} 
In this condition,  it is possible to construct a function $\Psi(x,\theta^F) = x - F(x, u, \theta^F)$  such that, when the system dynamic has reached the equilibrium (i.e., a fixed point of the state, $x^*$), $\Psi(x,\theta^F)=0$. Differentiating $\Psi(x,\theta^F)$ w.r.t the parameters $\theta^F$ at $x^*$ and rearranging the terms yields the gradient of the steady state $x^*$ w.r.t. the parameters of the stable dynamical system, 
\begin{equation}
    \frac{\partial x^*}{\partial \theta^F} = \left(I - J_{F, x^*} \right)^{-1} \frac{\partial F(x^*, u, \theta^F)}{\partial \theta^F}.
    \label{eq:steady_state}
\end{equation}
where $J_{F, x^*} = \frac{\partial F(x^*, u, \theta^F)}{\partial x^*}$ is the Jacobian matrix of $F$ evaluated at $x^*$.
This is a result from the Implicit Function Theorem \cite{rudin1976principles}, which requires ($i$) $\Psi$ to be continuously differentiable and
($ii$) $I - J_{F, x^{*}}$ to be invertible \cite{liao2018reviving}.
The term $\frac{\partial x^*}{\partial \theta^F}$ is exploited to compute the gradient of the loss function w.r.t. the learnable parameters that, by leveraging the chain rule, is,
\begin{equation} 
     \frac{\partial L}{\partial \theta^F} =  \frac{\partial L}{\partial y}\frac{\partial y}{\partial x^*}\frac{\partial x^*}{\partial \theta^F}.
     \label{eq:urra}
\end{equation}
When substituting Eq. \eqref{eq:steady_state} into Eq.~\ref{eq:urra} loss derivative, we get,
\begin{equation}
     \frac{\partial L}{\partial \theta^F} =  \frac{\partial L}{\partial y}\frac{\partial y}{\partial x^*}\left(I - J_{F, x^*} \right)^{-1} \frac{\partial F(x^*, u, \theta^F)}{\partial \theta^F}.
     \label{eq:rbp_derivative}
\end{equation}
Given that the Jacobian is non-symmetric for standard \sigla{RNNs}, directly using  solvers for linear system is impractical. Conversely, the standard \sigla{RBP} approach is to compute the term, 
\begin{equation}
    z = \left(I - J^T_{F, x^*} \right)^{-1} \left(\frac{\partial L}{\partial y}\frac{\partial y}{\partial x^*}\right)',
    \label{eq:zeta_rbp}
\end{equation}
 via fixed point iterations.
\sigla{RBP} was the learning algorithm exploited in the seminal works introducing Graph Neural Networks (\sigla{GNNs}) \cite{gori2005new,scarselli2008graph}, which can be considered  {generalizations} of
\sigla{RNNs} to handle graph-structured input data \cite{sperduti1997structuressupervised,sperduti1998graphs,micheli_graphs2009}. Indeed, the current literature refers to  \sigla{GNNs} by Scarselli et al. \cite{scarselli2008graph} as Recurrent \sigla{GNNs} (\sigla{RecGNNs}) \cite{wu2020comprehensive}. Inference in \sigla{RecGNNs} can be interpreted as a diffusion process along the graph up to the convergence to fixed points of the nodal states. \sigla{RecGNNs} ensure the \sigla{RBP} conditions ($i$) and ($ii$) by forcing $F(\cdot)$ to be a contraction map on a  Banach space, a choice thatneverthelessposes some strong limitations on the model capacity , and is difficult to satisfy.
When used with models that satisfy such assumptions, the main computational cost of \sigla{RBP} lies in solving a linear system (i.e., where the most expensive operation is the matrix-vector product $J_{F, x^*}^T z$), which has constant memory and computation time with respect to the number of unrolling steps.

\parafango{Modern RTRL} Because of the high memory and time complexity characterizing \sigla{RTRL}, researchers have recently focused on finding more efficient approximations. For example, Unbiased Online Recurrent Optimization (\sigla{UORO}) \cite{tallec2017unbiased} is a stochastic approximation of \sigla{RTRL}. If, for simplicity in the description and without any loss of generality, we consider $\theta^F$ to be a matrix of weights, indexed by pairs of indices $(i,j)$, \sigla{UORO} decomposes $M_t$ of Eq.~\ref{eq:inference_matrix} (which is a 3D tensor due to what we just stated about $\theta^F$) as the outer product of two tensors of lower rank $A_t$ and $B_t$, such that,
\begin{equation}
    M_t^{kij} = A^k_t B^{ij}_t,
\end{equation}
being $k$ the index of a unit belonging to the state. \sigla{UORO} provides stochastic approximations for $A_t$ and $B_t$ defined through a random vector $\nu \in \mathbb{R}^{\dstate}$, which satisfies $\mathbb{E}[\nu^i \nu^j] \propto \delta_{ij}$ and $\mathbb{E}[\nu^i]=0$. More precisely,
\begin{equation}
    \begin{aligned}
        A^k_t &= \rho_0 \sum_{k'} J^{kk'}_t A^{k'}_{t-1} 
 + \rho_1 \nu^k, \\
    B^{ij}_t &= \rho_0^{-1} B^{ij}_{t-1}  + \rho_1 \sum_{k'} \nu^{k'} \bar{M}^{k'ij}_t,
    \end{aligned}
\end{equation}
where $\rho_0$ and $\rho_1$ are positive constants. It is possible to prove (see \cite{marschall2020unified}) that the resulting outer product is an unbiased estimator for $M_t$. This algorithm is still online and past-facing, but the memory and time complexities are reduced to $\mathcal{O}(\dstate^2)$. Another example of \sigla{RTRL} approximation is the Sparse $n$-Step Approximation (\sigla{SnAp}) \cite{menick2020practical} algorithm, which imposes sparsity on the matrix $M_t$ to reduce the amount of computation in the product $J_t M_{t-1}$ of Eq. \eqref{eq:inference_matrix_rec}. The ``$n$-Step'' in \sigla{SnAp} refers to the fact that the algorithm considers gradients over $n$ time steps for approximating $M_t$. More precisely, if $\theta^{F}$ is flattened as a vector, the influence matrix $M$ is approximated as follows,
\begin{equation}
    M^{kz}_t \approx \begin{cases}
         M^{kz}_t, \quad \text{if ${\theta_{t}^{F}}^{z}$ influences hidden unit $x^{k}_{t+n}$} \\
         0, \quad \text{otherwise.}
    \end{cases}
\end{equation}
If we indicate with $s$ the level of sparsity of the matrix $M_t$, and we define $d$ as $d := 1-s$, the \sigla{SnAp-1} algorithm ($n=1$) for a fully connected \sigla{RNN} has a memory complexity of $\mathcal{O}(\dstate + d|\theta^{F}|)$ and a time complexity of $O(d(\dstate^2 + |\theta^{F}|))$. \sigla{RTRL} can be also made computationally tractable introducing different RNN structures and adopting specific learning processes exploiting these architectures for enabling scalable, unbiased and noise-free gradient estimation. For example, Schmidhuber et al. \cite{irie2023exploring} proposed to apply \sigla{RTRL} gradient computation to \sigla{RNNs} with element-wise recurrence (\sigla{eLSTM}). Under this architectural assumption, it is possible to obtain forward recursive formulas for the inference matrix, which have a memory complexity of $\mathcal{O}(\din\dstate)$ and a per-step time complexity of $\mathcal{O}(\dstate^2)$. Always focusing on the net structure, Sutton et al. proposed Columnar Neural networks (\sigla{Col-NNs}) \cite{javed2021scalable}, Constructive Networks and Columnar-Constructive networks (\sigla{CCNs}) \cite{javedscalable,javed2023online}. In \sigla{Col-NNs} \cite{javed2021scalable}, The network structure is restricted to be composed of independent, potentially deep columns. Each column presents a scalar recurrent state, which is not connected to the states of the other columns. Therefore, it is possible to apply \sigla{RTRL} to each column individually, reducing the computational cost to $\mathcal{O}(|\theta_{F}|)$, which means that \sigla{RTRL} for \sigla{Col-NNs} scales linearly with the size of the parameters. However, the structure of \sigla{Col-NNs} lacks hierarchical recurrent features. To introduce this hierarchy, Constructive Networks \cite{javedscalable} have been introduced, learning the recurrent network one feature at a time. The learning process prioritizes the acquisition of weights associated with the first recurrent feature, exclusively linked to the input. Subsequently, these weights are frozen, facilitating progression to the subsequent hidden unit. This unit can now establish connections with preceding recurrent features. Notably, the output weights remain dynamic, undergoing continual updates. By sequentially focusing on discrete subsets of the network during training, Constructive Networks incur even lower per-step computations than Columnar Networks. Consequently, \sigla{RTRL} can be implemented efficiently. Constructive Networks have the limitation of being unable to learn multiple features in parallel. To overcome this issue, \sigla{CCNs} \cite{javed2023online} learn multiple columns that take as input the features of all the existing frozen columns, iteratively constructing multiple columns of recurrent features in parallel. For other approximations of \sigla{RTRL}, see
\cite{marschall2020unified}. Finally,  inspired by the success of networks 
with recurrent units \cite{orvieto2023resurrecting} and leveraging the 
fact that recurrence with only self-loops greatly simplifies \sigla{RTRL}
\cite{mozer2013focused,mori1989bps}, in \cite{zucchet2023online} the authors propose
a modification of \sigla{RTRL} that is tailored to architectures with linear 
recurrent blocks interconnected through nonlinear layers. In particular, within this setting, they propose an approximation of the gradient 
in the case where more than one recurrent layer is stacked, artificially reducing the dependencies over time across layers.

\parafango{Modern RBP} Liao et al. \cite{liao2018reviving} investigated the strict requirements of \sigla{RBP} and proposed two variants based on conjugate gradient on the normal equations (\sigla{CG-RBP}) and Neumann series (\sigla{Neumann-RBP}), respectively. The former exploits an iterative solver to tackle Eq. \eqref{eq:zeta_rbp}, the conjugate gradient method on the normal equations, that however requires an expensive matrix multiplication (i.e., $J_{F, x^*}J^T_{F, x}z$) and, given that it uses normal equations, has a squared condition number leading to slower convergence times. The latter, \sigla{Neumann-RBP}, exploits a property of convergent Neumann series, $\sum_{k=0}^{\infty}A^k = (I-A)^{-1}$. A sufficient condition for the series convergence is that the largest absolute eigenvalue of A, namely $\lambda$, must be $ \lambda<1$. When  $A=J^T_{F, x}$, it is possible to replace the $(I- J^T_{F, x^*})^{-1}$ term in Eq. \eqref{eq:rbp_derivative} with the series sum. Additionally, the gradient $\frac{\partial L}{\partial \theta^F} $ can be approximated with the $K$-th order truncation of the Neumann series.  Memory complexity is constant with respect to the number of truncation steps, and the algorithm, given that it relies only on the steady state $x^*$, does not require to store the hidden states in the forward pass of the \sigla{RNN}, as done in \sigla{BPTT}. The authors remark the equivalence of \sigla{Neumann-RBP}  with \sigla{BPTT} when the Neumann series converges, and that a $K$-step \sigla{Neumann-RBP} is equivalent to a $K$-step Truncated \sigla{BPTT}.
Linsley et al. \cite{linsley2020stable} focused on recurrent vision models \cite{linsley2018learning} and remarked that when trained with standard \sigla{RBP} their training dynamics devolve into an unstable regime. The authors identified the root cause of this issue in the aforementioned condition ($ii$) of \sigla{RBP}, i.e., the fact that $I-J_{F,x^*}$ is not invertible. Forcing $F(\cdot)$ to be a contraction map, as in \cite{scarselli2008graph}, requires globally contractive model components, such as squashing non-linearities (e.g., $\text{sigmoid}$ and $\text{tanh}$), that however can be suboptimal for some computer vision tasks and hinder recurrent vision model performances. Thus, the authors proposed a soft  architecture-agnostic constraint for
learning local contraction maps, i.e., the Lipschitz Coefficient Penalty (\sigla{LCP}) $\lVert (\mathbb{1}  \cdot J_{F, x^*} - \lambda)^+\rVert_2$, where $(\cdot)^+$ denotes element-wise rectification and $\lambda \in [0,1)$ is an hand-selected Lipschitz constant which bounds $\lVert J_{F, x^*} \rVert_2$, tuning the degree of contraction in $F(\cdot)$. This choice allows to keep the largest singular value of $J_{F,x^*} < 1$, and forces  $F(\cdot)$ to be locally contractive at $x^*$. \sigla{LCP} can be combined with any task-related loss function for optimization.

\parafango{Lifted Methods} The vanishing/exploding gradient issues that arise in training \sigla{RNNs} with \sigla{BPTT} are critical due to the usage of gradient descent as the optimization technique. Alternative approaches have emerged within the family of \textit{lifted methods}. Overall, the main intuition of lifted methods is to act in an enlarged space where the neural states represent additional variables to the training problem, and the propagation dynamics of Eq. \eqref{eq:RNN} are expressed as architectural constraints as follows,
\begin{equation}
\begin{aligned}
    &\text{min}_{\theta, x} \quad \mathcal{L}(\theta) \\
    &\text{s.t.} \qquad x_{t} = F(x_{t-1}, u_t, \theta^F),  
\end{aligned}    
\end{equation}
where $\mathcal{L}$ is the loss function of Eq. \eqref{eq:loss_rnn}. Thus, such models are referred to as ``lifted'' because the usual parameter search space of $\theta$-variables is lifted to an higher dimensional space composed by  $(\theta, x)$-variables. The common approach is to transform the non-smooth constrained optimization problem into a smooth unconstrained problem in the enlarged space.   
Early approaches were proposed for feed-forward architectures (but can be easily extended to \sigla{RNNs}) \cite{carreira2014distributed}, by adding quadratic penalties to approximately enforce the equality constraints. Succeeding works \cite{taylor2016training} use Lagrange multipliers to exactly enforce equality constraints and optimize via the Alternating Direction Method of Multipliers (\sigla{ADMM}) and Bregman iteration. A clear advantage of these methods is the ability to decompose the training problem into multiple, local sub-problems, which can
be solved efficiently.
Recently, works by Askari et al. \cite{askari2018lifted,pmlrv108gu20a} proposed convex and biconvex formulations that can be optimized using Block Coordinate Descent. These methods were extended to \sigla{RNNs} in \cite{askari2018lifted}.
Marra et al. \cite{marra2020local} investigated the connections of lifted methods to Backpropagation, and proposed a hard-constraining scheme, referred to as \sigla{Local Propagation}, based on the augmented Lagrangian. Learning consists of a differential optimization problem converging towards a saddle point of the Lagrangian. 
Interestingly, this approach has been extended to devise a novel learning algorithm for \sigla{RecGNNs} \cite{tiezzi2020lagrangian}, allowing for the exploitation of deep \sigla{RecGNNs} \cite{tiezzi2021deep,maggini2024lagrangian}, by implicitly expressing the state convergence procedure via a constraint satisfaction mechanism. This removes the need for iterative procedures to be applied at each training epoch, and eliminates the network unfolding of the original model { and the harsh constraints imposed by \sigla{RBP}}.
Despite achieving good performance in classic benchmarks, the memory requirements of lifted methods, due to the introduction of additional trainable parameters, remain their main drawback.

\parafango{Alternative Methods} There have been several other attempts to propose alternatives to \sigla{BPTT}. Forward Propagation Though Time (\sigla{FPTT} \cite{kag2021training}) avoids the temporal unrolling by updating the RNN parameters at each time step towards the optimization of an instantaneous risk function: the loss at time $t$ plus a dynamically evolving regularizer, controlled by a state-vector, which summarizes the evaluation of past losses. In details, at each time $t$, a two-step update is applied, given the instantaneous loss function $\ell_t(\theta) := \ell \left( y_{t}, \hat{y}_{t}\right | \theta) $ and the RNN learnable parameters $\theta$ (see Section \ref{sec:related}):
\begin{equation}
    \begin{aligned}   
\overline{\ell}(\theta) &:= \ell_t(\theta) + \frac{\alpha}{2} \| \theta - \overline{\theta}_t - \frac{1}{2\alpha} \nabla \overline{\ell}_{t-1}(\theta_t)\|^2, \\ 
&\hskip -0.1cm \theta_{t+1} = \theta_t - \eta \nabla_{\theta} \overline{\ell}(\theta) \big|_{\theta=\theta_t}, \\ 
&\hskip -0.1cm \overline{\theta}_{t+1} = \frac{1}{2} (\overline{\theta}_t + \theta_{t+1}) - \frac{1}{2\alpha} \nabla \ell_t(\theta_{t+1}),
    \end{aligned}
    \label{eq:fptt}
\end{equation}
where $\alpha$ is a weighing factor, $\overline{\ell}(\theta)$ denotes the augmented instantaneous risk function and $\overline{\theta}$ is the state-vector that summarizes past losses. Indeed, such a ``summary'' represents a running average of past $\theta_t$ plus a correction term, that enforces stability in updates and convergence of the parameters toward a fixed point (i.e., only in this case hidden state trajectories simulate the usual static time invariant RNN parameters). 
\sigla{FPTT} requires $\mathcal{O}(\dseq)$ gradient computations for an $\dseq$-length sequence, while \sigla{BPTT} computes gradient only once, when the whole sequence has been processed. However, the constants (when evaluating the complexity) involved in taking gradient for the full-sequence are higher than computing single-step gradients. From the memory point of view, \sigla{BPTT} stores intermediate hidden state about the whole sequence, i.e., $\mathcal{O}(\dseq)$, while  \sigla{FPTT} does not require storing hidden states, i.e., it is $\mathcal{O}(1)$. 
The authors of \sigla{FPTT} additionally propose a more efficient variant  that performs updates restricted to $K$ time-steps, referred to as \sigla{FPTT-K}. Given that Eq. \eqref{eq:fptt} exploits instantaneous loss computations, when the task does not involve step-wise supervisions (e.g., terminal prediction),  \sigla{FPTT} leverages an alternative formulation that approximate the previous one \cite{kag2021training}.

Going beyond \sigla{FPTT}, another alternative method for addressing the exploding-vanishing gradient problem in processing long-sequences, is the one of Park et al. \cite{park2023persistent}. Such a method consists in a novel initialization scheme for \sigla{RNNs} enhancing learning of temporal dynamics. In order to achieve a \textit{stable limit cycle}, defined as an attracting ring manifold where the neural activations form a periodic trajectory, the authors structure the weight matrix using a scaled rotation matrix. This results in a block orthogonal matrix arrangement where, within each $2\times 2$ block, the behavior of the paired neurons exhibits spontaneous oscillations. The exploding-vanishing gradient issue can also be related to the presence of \textit{bifurcations} in the parameter space of the \sigla{RNN}, which represent qualitative shifts in network dynamics due to parameter variations. As shown by Eisenmann et al. \cite{eisenmann2023bifurcations}, bifurcations are associated with abrupt changes in stability regions and the topological structure of the  state space. The authors of \cite{eisenmann2023bifurcations} propose a heuristic algorithm (\sigla{SCYFI}, which stands for Searcher for Cycles and Fixed points) for identifying fixed points and cycles in ReLU-based \sigla{RNNs} and determining their existence and stability regions in the parameter space, along with eventual bifurcations. Once that fixed points, cycles and bifurcations have been identified, Generalized Teacher Forcing (\sigla{GTF}) (a method aimed at redirecting diverging trajectories towards their intended targets) tends to circumvent bifurcations during training. Finally, we mention the different route followed by Echo State Networks \sigla{(ESNs)} and, more generally, by instances of Reservoir Computing \sigla{(RC)}, where a large (usually non-learned) sparse connectivity layer is followed by a trainable readout function. {\sigla{ESN} offer a compelling alternative to traditional \sigla{RNNs} by separating the roles of dynamic feature extraction and learning. In an \sigla{ESN}, the core of the network, called the \emph{reservoir}, consists of a large, fixed, and sparsely connected recurrent layer, whose weights are typically randomly initialized and left untrained. This reservoir serves as a dynamic projection space that transforms input signals into a rich set of nonlinear features over time. The key innovation lies in \emph{echo state property}, which ensures that the effect of any initial state fades away, making the network response dependent primarily on the recent input history. This property allows for stable dynamics without the need to train the internal connections, significantly simplifying the learning process compared to standard \sigla{RNNs}. Learning in \sigla{ESNs} is confined to the readout layer, a typically linear function that maps the high-dimensional reservoir states to the desired output. Because only this readout is trained, usually via efficient linear regression techniques, \sigla{ESNs} are computationally efficient and less prone to issues like vanishing or exploding gradients.} The reader can find more information in \cite{micheli_guest_reservoir_2023,micheli_dynamic_graph_echo_state_2022,micheli_echo_markovian_factors2011}.

\parafango{Koopman operator view}
{Koopman theory provides a powerful mathematical framework for analyzing nonlinear dynamical systems by embedding them into a linear, though typically infinite-dimensional, function space. The core idea is to focus not on the system's state evolution itself but on how observable functions of the state evolve over time under the action of the \emph{Koopman operator}, which is linear even if the underlying system is nonlinear \cite{koopman1931hamiltonian}. This approach has found recent traction in machine learning, particularly for modeling time-series data, where traditional \sigla{RNNs} struggle with issues like vanishing gradients and lack of interpretability \cite{pascanu2013difficulty, bengio1994learning}. In this context, Azencot et al.~\cite{pmlr-v119-azencot20a} propose a \emph{Consistent Koopman Autoencoder}, a model that draws a direct link between Koopman theory and deep learning by embedding high-dimensional temporal data into a latent space where dynamics evolve linearly via learned Koopman operators. Unlike standard \sigla{RNNs}, which rely heavily on training to learn latent state transitions, this method uses physics-informed constraints, such as forward-backward consistency, to promote stability and interpretability. Their approach retains computational tractability while outperforming traditional and Koopman-based baselines, especially under noise and over longer horizons. Thus, Koopman theory not only complements RNNs by offering a principled alternative but also enriches neural forecasting architectures with physically meaningful structure and improved generalization capabilities.}

\parafango{Biologically-inspired Methods} Despite its extensive and successful exploitation in training procedures for state-of-the-art models, the Backpropagation algorithm (\sigla{BP}) has been also criticized about its lack of biological plausibility \cite{crick1989recent}. The main issue is represented by the way in which \sigla{BP} performs credit assignment (see Section \ref{sec:related}), with backpropagation-based synaptic weights adaptations that are unlikely to occur in real neuronal cells. Several other issues such as weight transport, non-locality, global feedback pathway, are summarized in  recent works \cite{lillicrap2019backpropagation,millidge2022predictive,ororbia2023brain}.
\sigla{BPTT} inherits all the plausibility-related issues from \sigla{BP}, with the addition that the need to store neural activities  for error computations is further exacerbated by the requirement to store them over time---a very different procedure w.r.t. the local message-passing structure happening in the brain \cite{ororbia2023brain}. 
Among several biologically plausible synaptic modification rules, such as Hebbian Learning \cite{kuriscak2015biological}, Millidge et al. \cite{millidge2022predictive} recently proved that Predictive Coding \cite{whittington2017approximation} approximates \sigla{BP} in arbitrary computational graphs, including \sigla{RNNs}, by relying solely local and plausible rules. Predictive Coding revolves around the perspective that the brain operates as a probabilistic generative model, consistently formulating predictions about its surroundings and adjusting internal hypotheses in accordance with sensory data, in line with the Bayesian brain hypothesis \cite{ororbia2023brain}. In \cite{millidge2022predictive}, the variational inference problem to be solved is that of inferring activation values of each node in the computational graph (i.e., activation values of neural units) given the start nodes (the input data) and target nodes. The neural computational graph is enriched by \textit{error neurons/units} that encode the difference between the activity at a given layer of the network and that which is predicted or generated from its parent nodes. In the case of \sigla{RNNs}, both the output $y_t$ and the states $x_t$ are augmented with additional error units that compute the mismatch between the actual values of the corresponding variables and the prediction obtained via variational inference. The final goal is the optimization of a functional known as the variational free energy, aiming at improving the generative model accuracy and minimizing model complexity.

Inspired from the same notions, Local Representation Alignment (\sigla{LRA}) \cite{ororbia2019biologically} aims at optimizing a related variational free energy by minimizing the discrepancy between neural activations and ad-hoc local targets.  \sigla{LRA} does not assume an underlying probabilistic generative model, and associates each layer with a target activity vector, so that synaptic weights are modified to guide the layer activity towards these targets. \sigla{LRA} ensures that realistic targets are chosen, considering the possible representation of a layer. Mismatch signals are computed as the first derivative of a distance function, and the target representation for each layer is determined based on accumulated mismatch signals from other layers. The synaptic plasticity rules involve adjustments to connections between layers and error message passing pathways.
The effectiveness of training recurrent models with \sigla{LRA} has been explored in \cite{ororbia2020continual}, where the authors propose the Parallel Temporal Neural Coding Network (\sigla{P-TNCN}), a layerwise parallelizable recurrent model which avoids unfoldings over the temporal dimension. 
The reader can refer to the very recent survey \cite{ororbia2023brain} for a comprehensive overview on biologically plausible learning rules.

{In summary, beyond architecture, recent work revisits training itself: modern approximations to \sigla{RTRL}, local credit-assignment schemes, lifted optimisation, and biologically inspired predictive-coding variants all aim to circumvent \sigla{BPTT}’s memory footprint and latency. These algorithmic developments complement the architectural trends surveyed earlier, completing the picture of next-generation recurrent sequence processing. \added{Table~\ref{tab:rnn_learning_summary} reports a comparative summary of the algorithms mentioned in this section, highlighting their key properties and complexities.} In the next Section~\ref{sec:benchmarks}, we review the most popular benchmarks for evaluating the models surveyed so far.

\begin{table}
\caption{\added{Comparative summary of two key features of training algorithms for recurrent models. 
We indicate if an algorithm is able to learn online on single components of each sequence, if the overall complexity (either spatial or temporal or both) is \(\mathcal{O}(L)\) (assuming all other parameters fixed), and if it
scales worse than  $\mathcal{O}(d_s^2)$ in term of its dependence from the hidden state size. 
See the paper text for more details.}}
\centering
\setlength{\tabcolsep}{4pt}
\small
\begin{tabular}{lcccc}
\toprule
\textsc{Algorithm} & \textsc{Online} & \textsc{$\mathcal{O}(L)$?} & \textsc{ $\mathcal{O}(d_s^k)$, $k>2$?} \\
\midrule
\sigla{BPTT} \cite{rumelhart1985learning} & No & Yes & No \\
\sigla{RBP} \cite{almeida1990learning,pineda1987generalization,DBLP:journals/tnn/Pearlmutter95} & No & Yes & No\\
\sigla{RTRL} \cite{williams1989learning} & Yes & No & Yes \\
\sigla{UORO} \cite{tallec2017unbiased} & Yes & No & No\\
\sigla{SnAp-1} \cite{menick2020practical} & Yes  & No & No \\
\sigla{FPTT} \cite{kag2021training} & Yes & Yes & No \\
\sigla{Local Prop.} \cite{marra2020local} & Yes & No & No \\
\bottomrule
\end{tabular}
\label{tab:rnn_learning_summary}
\end{table}}

\section{Benchmarks}
\label{sec:benchmarks}
Devising appropriate benchmarks for evaluating the wide spectrum of architectures for processing long sequences is a fundamental step. Indeed, looking back at the whole history of sequence processing, up to the present day, there is a clear lack of a well-established and consistent consensus on which benchmarks to use when comparing different algorithms. Ideally, benchmarks  should create conditions to investigate a model's ability to preserve long-term dependencies, providing long sequences and tasks in which old information might play an important role. However, the meaning of \textit{long sequence} has changed over the years. While in the early era of vanilla \sigla{RNNs} and \sigla{LSTMs}, a sequence was considered to be ``long'' when it consisted  a number of steps in the order of a few tens or hundreds, current models (Recurrent Transformers, \sigla{SSMs}) are capable of processing sequences composed of thousands to millions of tokens \cite{bulatov2023scaling}.

In this section , we describe the most common benchmarks that have been tackled since the dawn of \sigla{RNNs} up to the more recent Long Range Arena benchmark \cite{tay2020long}, and our analysis is aimed at providing a comprehensive view of benchmarking with long sequences. We mostly focus on the benchmarks and datasets used in the large set of papers described in this survey: Table \ref{tab:benchs} summarizes the benchmarks (first column) and the scientific papers in which such benchmarks have been used (subsequent columns, each column containing the papers described in a different section of this survey). We categorized the various benchmarks according to their application field, distinguishing among synthetic benchmarks, computer vision, natural language, time series, audio/speech, and the recent Long Range Arena \cite{tay2020long}, which includes multiple sub-benchmarks.

\subsection{Synthetic Benchmarks}

\parafango{Adding Task} The \sigla{Adding} task \cite{hochreiter1997long} is a sequence-based regression problem which serves as a benchmark to evaluate the capability of \sigla{RNNs} in learning long-term dependencies. At every time step, a 2-dimensional input vector is provided, $u(t) = (u_1(t), u_2(t))$, consisting of a real-valued signal, sampled uniformly in range $(0, 1)$, and a binary value, respectively. In particular, given a sequence of $\dseq$ elements, $u_2(t) \neq 0$ only in two specific time instants of the $\dseq$ ones that compose input sequence. Let us assume such two time instants are $t_a$ and $t_b$. The goal is to predict the sum of the real-valued entries of the inputs proved at $t_a$ and $t_b$, i.e., $u_1(t_a) + u_1(t_b)$ (see Figure \ref{fig:copy-heads}).
The challenge of this task consists in capturing the possibly long-term dependencies between the pairs of relevant elements of the sequences. Sequences have different lengths, usually $\dseq \in (100, 500, 1000)$. The quality of a model in approaching this task is evaluated in function of its ability to accurately predict the aforementioned sum, measured by the Mean Squared Error (MSE). The baseline performance is set by a simple model always predicting a sum value of $1$, resulting in an MSE of $\approx 0.1767$. 
This task has been exploited in several works \cite{hochreiter1997long,le2015simple,arjovsky2016unitary,li2018independently,kag2021training} to assess the capability of the proposed \sigla{RNNs} in handling long-range dependencies. 


\parafango{Copy Task} The \sigla{Copy} task \cite{hochreiter1997long} is a largely known benchmarks to evaluate the network ability to recall information observed several time steps in the past, asking the network to reproduce the ``initial'' tokens of a long sequence. Data consist of sequences of $\dseq$ + 20 digits. The first ten tokens $(a_0,a_1,...,a_9)$ are randomly chosen from $\{1,...,8\}$ and represent the sequence to be ``remembered''. The subsequent $\dseq$ tokens are populated by noise (usually simply modeled by the integer $0$), and the last ten
tokens are all equal to the \textit{repeat marker} (usually the integer $9$), that serves as an trigger to tell the model that it is time to reproduce the initial 10 tokens. In fact, the goal of the recurrent model is to output the starting ten tokens $(a_0,...,a_9)$ in the correct order during the last 10 time steps,
whenever the \textit{repeat marker} 9 is presented.  The random
guessing baseline has loss $\text{log}(8) \simeq 2.08$.  Training and test examples are both generated this way.
Intuitively, temporal credit assignment is critical in this task.
Several relevant works tackled this task \cite{hochreiter1997long,arjovsky2016unitary,menick2020practical,ba2016layer,gu2020hippo,zucchet2023online,park2023persistent}. 
Gu et al. \cite{gu2023mamba} observed that Linear Time Invariant \sigla{SSMs} (Section \ref{sec:ssm}), encompassing linear recurrence and global convolutions, can efficiently tackle this task by focusing solely on tracking time rather than comprehensively understanding the data. For instance, this could involve constructing a convolution kernel of the right length, as also pointed out by Romero et al. \cite{romero2021ckconv}. For this reason, it was proposed to exploit the \sigla{Selective Copying} task (originally introduced by Jing et al. \cite{jing2019gated} as the \sigla{Denoise} task). This task disrupts the straightforward approach of \sigla{Copy} by introducing random spacing between relevant tokens (i.e., relevant tokens are interleaved by a varying number of noise-like tokens, see Figure \ref{fig:copy-heads}), eliminating trivial solutions. Consequently, this adjustment necessitates content-aware reasoning for the network to effectively memorize and filter out relevant tokens while disregarding irrelevant ones. Lately, the \sigla{Selective Copy} task has been used to asses Large Language models capability of retraining long range dependencies \cite{gu2023mamba,de2024griffin}, with sequences of $\dseq=4096$, and a vocabulary of $16$ tokens. 
Some other variants, such as the \sigla{Capacity} task, have been proposed to evaluated the case of sliding windows \cite{voelker2019}. 
\begin{figure}
    \centering
    \includegraphics[width=0.8\columnwidth]{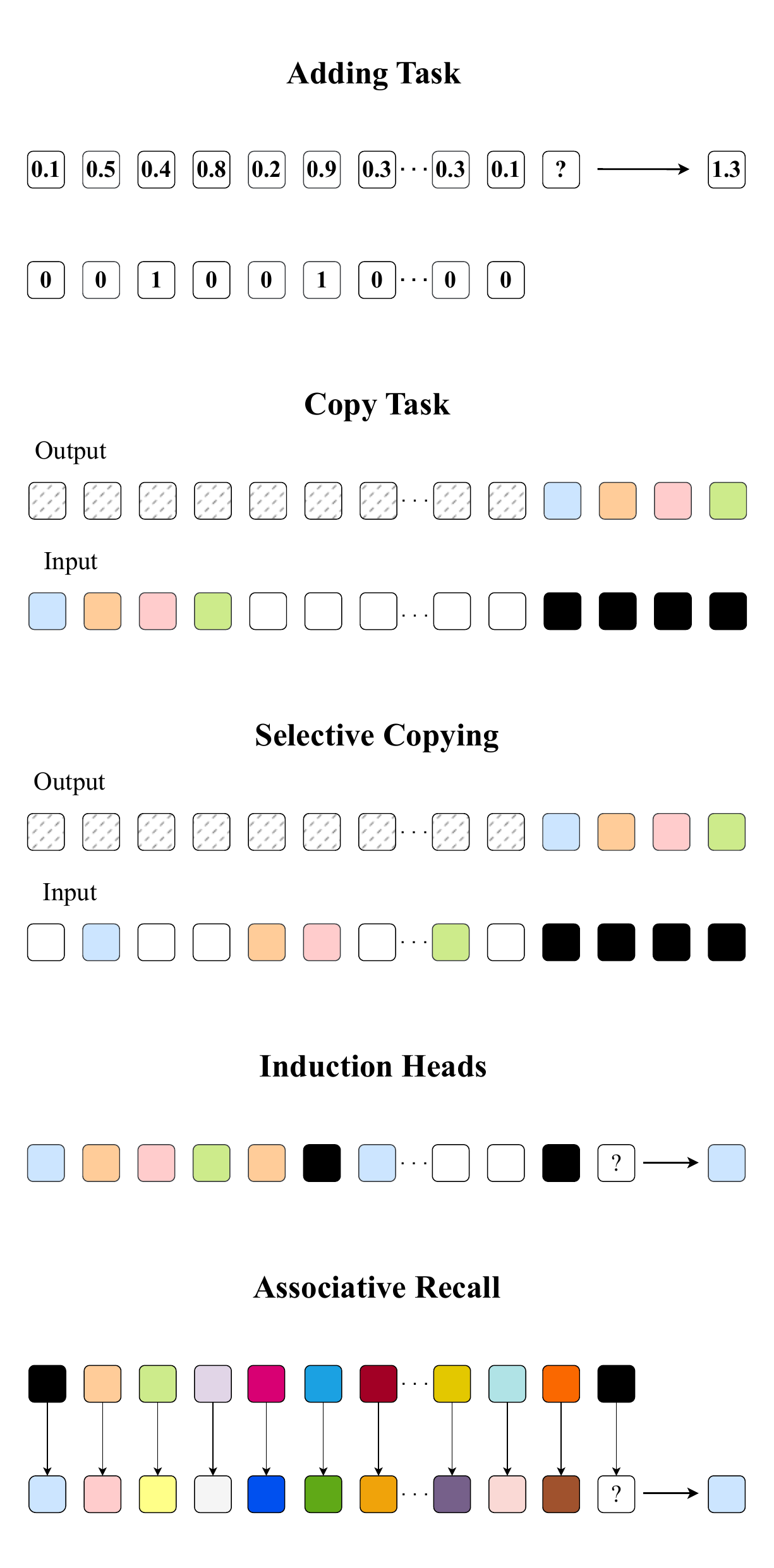}
    \caption{
    Synthetic benchmarks. The standard version of the \sigla{Copy} task involves constant spacing between input and output elements and is easily solved by time-invariant models exploiting linear recurrences and global convolutions. Crossed-out tokens in the output denote tokens that are masked out in the loss. The \sigla{Selective Copying} task has random spacing in between inputs and requires time-varying models that can selectively remember or ignore inputs depending
on their content. The \sigla{Induction Heads} task is an example of associative recall that requires retrieving an answer based on context, a key ability for Large Language Models. See the main text for further details.}
    \label{fig:copy-heads}
\end{figure}


\parafango{Induction Heads}  In-context learning abilities of Large Language  Models (LLMs)\footnote{In-context learning is usually defined as the ability to learn directly from examples within the context (the prompt) of the input sequence at inference time, and use information from the input to generate the right answer for the output \cite{olsson2022context}.} have been recently hypothesized to emerge from the so-called induction head mechanism, that was tested by means of the synthetic \sigla{Induction Heads} task, proposed in  \cite{olsson2022context} (Figure \ref{fig:copy-heads}). 
In this task, models are required to perform copy and \textit{associative recall} operations. Specifically, the models are tasked with recalling the token immediately following a designated special token (referring to Figure \ref{fig:copy-heads}: the black token is the special token, the token to be retrieved is the blue one). When the special token is presented once again, the model must be capable to subsequently retrieve and output the token following it within the context (the blue token in Figure \ref{fig:copy-heads}). 
The successful execution of this task implies that the model possesses the ability to generalize to longer sequences than those it was trained on. During training, a vocabulary size of $t$ tokens is exploited ($t=16$ in \cite{gu2023mamba,de2024griffin}, various $t$ are tested in \cite{polihyena}), with sequences of length $\dseq=256$, composed by tokens randomly sampled from the vocabulary. Additionally, the location of the special token within the sequence is also randomly determined.
The model generalization and extrapolation abilities are assessed at test time on sequences with increasing lengths, ranging from $2^6 = 64$ to $2^{20} = 1,048,576$.
\sigla{Induction Heads} is exploited in several recent works \cite{gu2023mamba,de2024griffin}.


\parafango{Associative Recall} Proposed in \cite{ba2016using}, this task is similar to \sigla{Induction heads} but introduces an additional challenge. The model processes multiple key-value pairs in a sequence (keys are characters, values are digits---the sequence is composed by keys followed by values, see Figure \ref{fig:copy-heads}). At the end of the sequence, one of the keys is presented as query, and the model must recall the specific value associated to it. It has been used in \cite{dao2022hungry,gu2023mamba,polihyena,yang2023gated,bulatov2022recurrent}. Arora et al. \cite{arora2023zoology} proposed an harder \sigla{Multi-query Associative Recall} task where multiple queries must be recalled.   

\parafango{Phonebook Look-Up} In the recently introduced \sigla{PhoneBook}    task \cite{jelassi2024repeat}, the model is asked to process a synthetic phone-book (e.g., each line  looks like “John Powell: 609-323
7777”) and to return the phone number when given a name. Various sizes of the phone-books are tested. The main intuition behind this task is to prove that despite the efficiency at inference-time, \sigla{SSMs} are somewhat limited when compared to Transformers on tasks that require copying from the input context, as assessed by \cite{jelassi2024repeat,de2024griffin}.

\parafango{Other Synthetic Tasks} The seminal paper proposing \sigla{LSTMs} by Hochreiter \& Schmidhuber \cite{hochreiter1997long} contains several interesting tasks to test long-range dependencies. Among the others, in \sigla{Task 2b} the input consists of  sequences of length $\dseq$, composed of vectors $u \in R^\din$. The input at $t=0$ is either $u_0 = (1, 0, . . . , 0)$ or $u_0 = (-1, 0, . . . , 0)$. For $t>0$, $u_t$ is a randomly chosen one-hot vector. 
The evaluated model must process the entire sequence and then output the sign of the first component of $u_0$. Recent works exploited these benchmarks \cite{li2018independently}. 
Bulatov et al. \cite{bulatov2022recurrent} proposed some other variants. For instance, in the \sigla{Reverse Task}, a certain input sequence should be generated at test time in the reversed order.

\subsection{Computer Vision Benchmarks}
\label{sec:compvis}

\parafango{MNIST} The MNIST dataset \cite{lecun1998gradient} has been  widely exploited to asses performances of sequence processing models. There are two common alternative setups/tasks: sequence classification and generative autoregressive image modeling.
In the former, each MNIST digit ($28 \times 28$ pixels) is decomposed into a sequential pattern and classification is performed once the whole sequence has been processed. In the scientific literature, different modalities have been explored\footnote{This also depends on the model architecture that is being tested. An \sigla{RNN} processes vectorial patterns (pixel-by-pixel or row-by-row), while Transformer-based models generally follow the Visual Transformer (ViT) \cite{dosovitskiy2020image} approach of processing a sequence of image patches.}, i.e., processing each MNIST digit following the order of scanlines, one pixel at a time, starting at the top-left corner of the image, and ending at the bottom-right corner (resulting in a sequence with $\dseq= 784$)---or alternatively, row-by-row, where each row represents the model input at a certain time $t$, resulting in $\dseq=28$).  We refer to both these dataset variations as Sequential MNIST (\sigla{sMNIST}) \cite{le2015simple}.
An harder variant is permuted sequential MNIST (\sigla{psMNIST}), where the same fixed random pixel permutation is applied to each digit of the dataset, resulting in an increased time-delay between interdependent pixels. The datasets \sigla{sMNIST}  and \sigla{psMNIST} are exploited in \cite{li2018independently,rusch2021unicornn,le2015simple,gu2021combining,smith2022simplified,kag2021training}, amongst others.
Differently, sequential image generation is performed by predicting images in pixel-by-pixel  manner \cite{katharopoulos2020transformers}. Image completions and unconditional samples are also usually considered.
The former takes as input a portion of an image (that can be considered as a prompt, and it is usually approached by Transformers architectures \cite{katharopoulos2020transformers}) and proceeds in generating the remainder of the image. The latter comprises the task of generating images with no condition in any context (such as a prompt text or another image).
Ororbia et al. \cite{ororbia2020continual} proposed a video-based generative modeling assessment, exploiting \sigla{BouncingMNIST}, \sigla{BouncingNotMNIST} and \sigla{BouncingFashionMNIST} in sequence prediction, zero-shot adaption and online continual learning setting. In \sigla{BouncingMNIST}, each sequence is a $20$-frame video consisting of two randomly chosen digits from MNIST, moving and bouncing around in a $64 \times 64$ frame. Digits are placed at random initial locations within the overall patch, are assigned a velocity and move and bounce off edges of the overall frame. When digits occupy the same location, they occlude each other and overlap. In \sigla{BouncingNotMNIST}, fonts/glyphs from the NotMNIST dataset\footnote{\scriptsize\url{https://yaroslavvb.blogspot.com/2011/09/notmnist-dataset.html}} are exploited, with the same characteristics (e.g., dimensions, sequence length, etc.) of \sigla{BouncingNotMNIST}. The same holds for \sigla{BouncingFashionMNIST}, with exemplars taken from the FashionMNIST dataset \cite{xiao2017fashion}.
Sequence prediction is tackled both with an encoder/decoder approach (first $10$ frames are provided as input to the encoder, while the decoder has to generate the next $10$ frames) and by training on input sequences of $10$ frames to reconstruct those $10$ frames as well as predict $10$ other future frames \cite{srivastava2015unsupervised} (a recurrent architecture processes image features extracted by a convolutional model). In zero-shot adaptation, a model that is trained on one dataset (e.g., \sigla{BouncingMNIST}) is tested on its ability to generate samples from another set (e.g., \sigla{BouncingNotMNIST}). In the continual online learning scenario \cite{mai2022online}, as discussed in \cite{ororbia2020continual}, models are trained on a stream composed of three of the aforementioned datasets, with a single pass over the , using mini-batches of size $1$ (i.e., pure online learning), which is a much more realistic setting when faced with infinite streams of patterns, where data arrives from different tasks at different times. The models are trained in a one-shot manner on a concatenation of the three datasets. We frame these approaches under the \sigla{genMNIST} identifier in Table \ref{tab:benchs}.

\parafango{CIFAR}  Similarly to the case of MNIST, CIFAR-10/100 \cite{krizhevsky2009learning} are also  exploited as datasets for sequence-oriented models. A common approach is to flatten each CIFAR image ($32 \times 32$ pixels with three channels) along height and width and process at each time step the three channels, resulting in $L=1024$.  Lim et al. \cite{lim2023parallelizing} apply random horizontal flipping before transforming it to a sequence with three channels by simply reshaping the signal. 
\sigla{TLB} \cite{didolkar2022temporal} (see Section \ref{sec:transformers}) tests the generalization abilities of the
model by comparing its performance on images of
higher resolution ($128 \times 128$ pixels) with respect to the ones experienced during training. A \sigla{ViT} \cite{dosovitskiy2020image} is exploited as the perceptual module and a Temporal
Latent Bottleneck module is added to it. 
The input image is split into patches of $4 \times 4$ pixels and fed
in raster order to the model. To predict the classification scores, the mean across the final Temporal Latent Bottleneck state vectors is computed and the resulting representation is processed through an MLP.
Zhai et al. \cite{zhai2021attention} consider the problem of image autoregressive modeling, by minimizing the negative log likelihood. In this case,  images are processed in scanline order resulting in an unrolled sequence length of $\dseq=3072$. Each sub-pixel
is represented as a $256$-way discrete variable. 
A different scenario involves the noise-padded CIFAR dataset (\sigla{npCIFAR}) as described in \cite{chang2018antisymmetricrnn}. In this setup, images undergo row-wise processing and are flattened across channels, creating sequences with a length of $\dseq = 32$. To introduce a distraction factor,  the sequence is extended by additional uniform random numbers (for $968$ steps) up to a sequence length of $1000$.
CIFAR10/100 dataset are considered in \cite{rusch2020coupled,rusch2021unicornn,gu2021combining,smith2022simplified,zhai2021attention,lim2023parallelizing}.

\parafango{ImageNet} Following similar ways of building sequences from static images, also the popular ImageNet  dataset \cite{dosovitskiy2020image} has been exploited in the literature of sequence processing. \sigla{RFA} \cite{choromanski2020rethinking} exploits a $64 \times 64$ ImageNet variant, proposed by Parmar et al. \cite{parmar2018image}, processed pixel-by-pixel and considering each channel as independent, resulting in $\dseq = 12288$. 
Other models simply integrate their approach into pre-existing architectures, such as \sigla{HRGN} \cite{qin2023hierarchically}, which is injected into a \sigla{DeiT} \cite{touvron2021training} structure and tested on the ImageNet-1k dataset for image classification. \sigla{SGConv} \cite{li2022makes} replaces the $7\times 7$ 2D convolutional
kernels of  \sigla{ConvNeXt} with \sigla{SGConvs}, which  treats 2D features as sequences.  
Additionally, Irie et al.  \cite{irie2022modern} leverage the Mini-ImageNet dataset \cite{vinyals2016matching}, composed of $100$ classes with $600$ examples each,  typically resized to $84 \times 84$ pixels \cite{ravi2016optimization}. 
Works exploiting ImageNet are the following:  \cite{choromanski2020rethinking,qin2023hierarchically,liu2022ecoformer,li2022makes,polihyena,irie2022modern}

\parafango{Action Recognition} Video Activity Detection focuses on predicting actions at each time step within a video. Thus, unlike video classification, this task deals with long videos containing multiple overlapping activities, introducing a challenging problem due to the need to strongly exploit and build an expressive temporal context. 
The recent \sigla{TTM} \cite{ryoo2023token} (see Section \ref{sec:transformers}) evaluated causally masked inference, providing activity predictions for incoming frames without considering future steps, in the Charades dataset \cite{sigurdsson2016hollywood}, which contains $9.8$k videos showcasing $157$ daily household activities, with $7.9$k training and $1.8$k validation clips. Videos can include multiple overlapping activities, requiring models to predict various activity classes per frame, taking into account interactions and longer temporal contexts. The average video length is $30$ seconds, creating a challenging scenario for temporal activity detection. The AVA v2.2 \cite{gu2018ava} dataset is also employed to test spatio-temporal activity detection, and is composed of bounding box annotations of $80$ atomic visual actions in $430$ $15$-minute movie clips. Li et al. \cite{li2018independently} leverage the NTU RGB+D dataset \cite{shahroudy2016ntu}, which is a skeleton-based action recognition task, comprising $56880$ sequences belonging to $60$ action classes and each containing $20$ frames.

\subsection{Natural Language Benchmarks} Several architectures (both in the context of Linear Transformers and \sigla{SSMs}, e.g., \sigla{TransNormer} \cite{qin2023scaling}, \sigla{GLA }\cite{yang2023gated}, \sigla{RWKV}\cite{peng2023rwkv}, \sigla{Hyena} \cite{polihyena}, \sigla{Mamba} \cite{polihyena}, \sigla{Griffin} \cite{de2024griffin}, and others) are being exploited to setup LLMs, with the final goal of optimizing performances at scale. Thus, the tasks that are tackled to test such solutions are the ones characterizing the world of Natural Language Processing. 

\parafango{Language Modeling} One of the peculiar characteristic behind the success of LLMs is their ability to model the generative likelihood of word sequences, i.e., they are able to predict the probabilities of future tokens. In other words, they can build powerful Language Models for a target language, being able to capture the syntax, grammar, and semantic properties.
There exist several Language Modeling dataset, and many of the works described in this survey are evaluated in Language Modeling tasks.
PennTreebank (\sigla{PTB}) \cite{marcus1993building}, \sigla{WikiText-103} \cite{merity2016pointer} and \sigla{The Pile} \cite{gao2020pile} are commonly used for evaluating model performances terms of \textit{perplexity}. 
\sigla{PTB} is a language model corpus containing 1 million words. Generally, models are trained to predict the next word (a variant denoted with \sigla{PTB-w}, evaluated with the perplexity metric) or character  (referred to as \sigla{PTB-c},  evaluated with \textit{bits-per-character} metric).
The corpus contains $42,068$ sentences ($971,657$ words, average word-length of about $4.727$ characters) of varying length (the range is from $3$ to $84$ words). The vocabulary for the character level models includes $49$ unique symbols (including one for spaces). The standard train/valid/test split consists of $929$k training words, $73$k validation words, and
$82$k test words, with lower-case words, numbers  replaced with token $N$, newlines replaced with $<$eos$>$, all other punctuation removed. The vocabulary is composed by 
the most frequent $10$k words, with the rest of the tokens replaced by an {$<$unk$>$} token \cite{mikolov2012context}. In literature, various segmentations of the corpus into training sequences of different lengths are being used ($\dseq=50$ for \sigla{PTB-w} in \cite{li2018independently,rusch2021long}, $\dseq \in (70,300)$ for \sigla{PTB-w} and $\dseq=150$  for \sigla{PTB-c} in \cite{kag2021training}). This task is considered in \cite{li2018independently,kag2021training,dai2019transformer,bradbury2016quasi,kerg2019non,kag2019rnns,rusch2021long,ororbia2020continual}.
\sigla{WikiText-103} \cite{merity2016pointer} is a long-range world-level language modeling benchmark. It contains $28$k articles from English Wikipedia, with an average length of $3.6$k tokens per article. In total, it contains $103$M words. The preprocessing is more realistic with respect to \sigla{PTB} (e.g., vocabulary is not limited to $10$k words, only words with a count below $3$ are discarded). It is used in \cite{gu2020improving,gu2021efficiently,ma2023mega,li2022makes,polihyena,peng2021random,irie2021going,schlag2021linear,qin2022devil,huang2022encoding,dai2019transformer,bulatov2022recurrent,LiLL0LZW023,kasai2021finetuning,mao2022fine,QinSDLWLYKZ22,wu2022memformer,katsch2023gateloop,qin2023hierarchically,menick2020practical}.
\sigla{The Pile} dataset \cite{gao2020pile}  is an $825$ GiB English text corpus designed for training large-scale language models. It is is composed of $22$ diverse and high-quality datasets, including  established Natural Language Processing datasets (\sigla{PG-19}, Books3, English Wikipedia, etc.)  and $14$ new high-quality curated language modeling datasets derived from heterogeneous sources (PubMed, ArXiv, GitHub, Stack Exchange, etc.--see Table 1 in \cite{gao2020pile} for a full description)). 
Besides its impact in training large language models, \sigla{The Pile} can serve as a benchmark for testing cross-domain knowledge and generalization ability of language models. It is exploited in \cite{polihyena,gu2023mamba,sun2023retentive,bulatov2023scaling,qin2023hierarchically,peng2023rwkv}.
Among the datasets contained in \sigla{The Pile}, 
the Project Gutenberg (\sigla{PG-19}) \cite{rae2019compressive}
consists of $28,602$ full-length books from the Project Gutenberg, published prior to $1919$, hence representing distinct writing styles with respect to other datasets based on modern text. It contains $6,966,499$ English language words.
When tokenized, each \sigla{PG-19} book has between $50$k-$100$k tokens. This datasets has been employed to test long range modeling capabilities of the approaches proposed in \cite{pilault2023block,hutchins2022block,wu2022memformer}. 
Datasets from Arxiv and GitHub were already been introduced in \cite{wu2022memorizing}, prior to \sigla{The Pile}. The ArXiv dataset is a corpus of \LaTeX{} sources of technical papers from the ``Mathematics'' section,  downloaded via the ArXiv Bulk Data Access. The GitHub dataset consists of source code extracted from diverse GitHub repositories featuring open-source licenses. In this dataset, all files within each GitHub repository are combined to form one long document. The evaluation metric commonly used with these datasets is bits-per-token (i.e., $\text{log}_2 (\text{perplexity})$, the lower the better). Used by \cite{hutchins2022block,pilault2023block,LiLL0LZW023}.
\sigla{Enwik8} \cite{mahoney2011large} is a character-level dataset consisting of $100$M unprocessed bytes of the XML text dump of the English Wikipedia. Its vocabulary contains $204$ characters, and it is exploited in \cite{zhai2021attention,huang2022encoding,dai2019transformer,bulatov2022recurrent,lei2018simple}.

\parafango{Commonsense Reasoning} A common experimental setting for evaluating LLMs consists in pre-training the model on \sigla{The Pile} and then testing the model capabilities on downstream zero-shot evaluation tasks. In particular, recent works leverage the {Language Model Evaluation Harness} (\sigla{LM-Eval-Harness})\footnote{\scriptsize\url{https://github.com/EleutherAI/lm-evaluation-harness}} from EleutherAI \cite{leogao}, which mainly evaluates on tasks/datasets that measure common sense reasoning---in which the model should use ``common sense'' or world knowledge to perform inference. It is composed by several tasks: \sigla{LAMBADA}, \sigla{HellaSwag}, \sigla{PIQA}, \sigla{ARC}, \sigla{WinoGrande}. 
\sigla{LAMBADA} \cite{paperno2016lambada} tests the modeling capacity on long-range contextual reasoning and language comprehension abilities. The model is given a certain context sentence, and it is asked to predict the last word of a target sentence based on the  
context. Accuracy and perplexity of the predicted last words are measured to assess the performance on language modeling tasks. The scaling law  property \cite{kaplan2020scaling,hoffmann2022empirical} is often measured, meaning that 
scaling language models would improve the accuracy
and reduce the perplexity. \sigla{LAMBADA} contains novels from the Book Corpus \cite{zhu2015aligning}, and comprises 
 $10,022$ passages, divided into $4,869$ development (that can be used to finetune models) and $5,153$
test passages (extracted from $1,331$ and $1,332$ disjoint novels, respectively). The average passage
consists of $4.6$ sentences in the context plus $1$ target sentence, for a total length of $\approx 75$ tokens. The dataset provides also training data, which include the full text of $2,662$
novels (disjoint from those in dev/test sets), comprising $203$ million words.
\sigla{HellaSwag} \cite{zellers2019hellaswag}\footnote{Short for {Harder Endings, Longer contexts, and Lowshot Activities for Situations With Adversarial Generations.}} expands the SWAG dataset \cite{zellers2018swag}, which introduced the task of \textit{commonsense natural language inference}: the model receives context from a video caption, along with four potential outcomes for the subsequent events. Among the provided choices, only one is correct, representing the actual succeeding caption in the video. Negatives are created using Adversarial Filtering, a method where a set of discriminators is employed to choose a tricky collection of incorrectly generated answers. According to the \sigla{HellaSwag} paper, models like BERT struggle with robust commonsense reasoning and tend to learn specific biases from the dataset. If there's a slight shift in language distribution, their performance drops significantly, even if the domain remains the same. \sigla{HellaSwag} introduces $70,000$ problems that are easy for humans ($95.6$\% accuracy) but challenging for machines ($50$\%). These problems consist of video captions from the ActivityNet Captions dataset and  context and follow-up paragraphs from WikiHow. To assess how a model can adapt to novel scenarios, category labels sourced from WikiHow and ActivityNet are sampled to build 'zero-shot' evaluation sets. For each set, being it used for validation or testing, two subsets are generated. The first one includes $5,000$ 'in-domain' examples from categories encountered during training. The second subset comprises $5,000$ 'zero-shot' examples randomly selected from categories held out during training. In total, there are $70,000$ examples in the dataset.
\sigla{PIQA} (Physical Interaction: Question
Answering) \cite{bisk2020piqa} benchmarks
are about physical commonsense understanding. 
Given a ``physical'' goal (``\textit{to separate egg whites using a water bottle...}'') expressed in natural language and two possible solutions (($i$) ``\textit{Squeeze the water bottle and press it 
against the yolk. 
Release, which creates suction and lifts the yolk;}'' ($ii$) \textit{``Place the water bottle and press it against the  yolk. Keep pushing, which creates suction and lifts the yolk.''}), a model must choose the most appropriate solution. The dataset assesses the capacity of natural language understanding models to connect text with a robust intuitive-physics model of the world. Humans effortlessly choose answer ($i$) because separating the egg involves pulling out the yolk, a task that machines find easily misleading. The dataset consists of more than $16,000$ training QA pairs with an additional $\approx 2$k and $\approx 3$k QA held out for development and testing, respectively. Goal sentences have an average length of $7.8$ words, while  correct/incorrect solutions have an average length of $21.3$ words. This results in over $3.7$ million lexical tokens within the training data.
The AI2 Reasoning Challenge (\sigla{ARC})   \cite{clark2018think}  consists of a collection of $7787$ natural grade-school science questions ( typically $4$-way multiple
choice), authored for human standardized tests. It is partitioned into two sets: Challenge Set (\sigla{ARC-c}),  which contains $2590$ questions which are incorrectly answered by both a retrieval-based algorithm and a word co-occurrence algorithm. An  Easy Set (\sigla{ARC-easy}) contrains the remainder of the questions. 
\sigla{WinoGrande} \cite{sakaguchi2021winogrande} 
is a large-scale dataset of $44$k pronoun resolution problems.  The task is to identify to which of two subject a \textit{pronoun}\footnote{In practice it is formatted as a fill-in-the-blank
problem, where the blank corresponds to the mention of one of the two names in the context.} is referring to (e.g., ``\textit{Robert woke up at 9:00am while Samuel woke up at 6:00am, so \_ had less time to get ready for school.}''; Options: {Robert} / Samuel). Generally, there is a twin sentence with nearly identical structure but where the answer is the opposite. It has been designed to be unsolvable for statistical models that rely on basic word associations. 
Going back to the main container of the just described datasets, the \sigla{LM-Eval-Harness}, we mentioned that it has been used in several recent works  \cite{gu2023mamba,de2024griffin,yang2023gated,peng2023rwkv,sun2023retentive,qin2023scaling}. 
Apart from \sigla{LM-Eval-Harness}, other comprehensive benchmarks for commons sense reasoning have been tested. 
Measuring Massive Multitask Language Understanding (\sigla{MMLU}) \cite{hendrycks2020measuring}  is a versatile benchmark for large-scale
evaluation of multi-task knowledge understanding. It consists of $57$ tasks including elementary mathematics, US history, computer science,
law, etc. It collects $15,908$ questions in total, split into a few-shot development set, a validation set, and a test set.
As shown in existing work, LLMs mostly outperform small models by a substantial margin on this benchmark, also showing  scaling laws in model size.  
We point the reader attention towards a recent survey on LLMs \cite{zhao2023survey} for further details on other commonly used datasets, such as   
 CMMLU \cite{li2023cmmlu}, C-Eval \cite{huang2024c} BoolQ \cite{clark2019boolq}, OpenBookQA \cite{mihaylov2018can} and others.

\parafango{Machine Translation} Another sequence-oriented benchmark in the context of language generation is conditional text generation \cite{li2022pretrained}, which focuses on generating texts satisfying specific task demands based on the given conditions. A common task is machine translation \cite{bahdanau2014neural}, whose goal is the translation of text or speech from a source language to another.  
Among the most common benchmarks, the \sigla{WMT} (several editions---2014 \cite{bojar2014findings} up to 2022 \cite{kocmi2022findings})
contains translation tasks from/to English to/from
several languages, such as Czech, French, German, Hindi, and Russian. A common testbed is to use the \sigla{WMT’14} English-to-German, which contains 4.6M sentence pairs, and the \sigla{WMT’17} the Chinese-to-English split which is composed by $20.6$M sentence pairs. 
The case-sensitive NIST BLEU score \cite{papineni2002bleu} is commonly adopted as the evaluation metric. 
It has been tackled in \cite{ma2023mega,peng2021random,schlag2021linear,hao2019modeling,chen2019recurrent,kasai2021finetuning,chen2018best}.
A similar benchmark is the \sigla{IWSLT} German–English spoken-domain translation \cite{cettolo2014report}. It consists of 
$209,772$ sentence pairs from transcribed TED and TEDx presentations, with a mean sentence length of $103$ characters for German and $93$ for English. It has been considered in \cite{peng2021random,bradbury2016quasi}.

\parafango{Other Language-oriented Benchmarks} Apart from the generation of high-quality natural language text, several models show strong abilities to generate regular and formal languages, such as computer programs (i.e., code synthesis) \cite{huang2022encoding,zhao2023survey}. Another popular direction for evaluation is the one is sentiment classification. The principal testbed is the \sigla{IMDb} movie review dataset \cite{maas2011}, a collection of $50$k movie reviews, equally balanced into positive and negative reviews and processed into equal-size train and test sets. Sequence length range from hundreds to thousands, with an average document length of $231$ words.  The aim of this binary sentiment classification task is to decide
whether a movie review is positive or negative. Several works considered this task \cite{rusch2020coupled,rusch2021unicornn,gu2020hippo,bradbury2016quasi,QinSDLWLYKZ22}.

\parafango{DNA Modeling}DNA can be easily connected to language, given that it consists of a sequence of discrete tokens (i.e., nucleotides) belonging to a limited vocabulary. Indeed, multiple LLM-based solutions for genomics have been proposed. Additionally, recent works have highlighted the advantages of by considering long-range interaction when predicting gene expression \cite{avsec2021effective}. Recent works in the context of this survey \cite{nguyen2024hyenadna,gu2023mamba} explore scaling laws across model size and sequence length as well as downstream classification. Pretraining is performed via causal language modeling (next token prediction) on the Human Genome \sigla{HG38} dataset, which consists of a single human genome containing $4.5$ billion tokens (DNA base pairs) in the training split. Please refer to the cited papers for further details.

\subsection{Speech/Audio Benchmarks} 

Processing sequences of ``raw'' audio data can be challenging for neural models, due to high-frequency sampling, resulting in very long sequences. Some traditional approaches involve complex pipelines that require extracting mixed-and-matched
hand-crafted features.  
The \sigla{Speech Commands} \cite{warden2018speech} dataset provides 1-second raw audio waveforms sampled at $16000$Hz (i.e., a 1-D sequence of $\dseq=16000$), consisting of $105,829$ recordings from $2,618$ speakers -- both background noise and a vocabulary of $35$ spoken words (such as ‘left’, ‘right’, etc.). 
The task is to classify which word was spoken. While several works \cite{kidger2020neural,kag2019rnns} exploit a pre-processed variant with standard mel-frequency cepstrum coefficients (MFCC), recent models tackle the problem directly processing the raw signal \cite{gu2021combining,gu2021efficiently,gupta2022diagonal,gu2022parameterization,smith2022simplified}. 
Other recent works \cite{gu2023mamba} explits a subset of the dataset (referred to as \sigla{SC09}) consisting of spoken digits ``zero'' through ``nine'' ($31,158$ training utterances---$8.7$ hours in total----by $2,032$ speakers). This is usually tackled as an autoregressive speech generation problem.      
Other similar dataset are: \sigla{TIMIT} \cite{garofolo1993darpa}, leveraged in  \cite{erichson2020lipschitz};  \sigla{Beethoven} (with the setting described by \cite{mehri2016samplernn}) consists of recordings of Beethoven’s 32 piano sonata; \sigla{YouTubeMix} is a 4 hour dataset of piano music. The latter two dataset are considered in \cite{goel2022s}.

\subsection{Time Series Benchmarks}
Time series consists of sequential samples paired with time-stamps, generally (but not necessarily) sampled at equally spaced points in time.
Such data is often characterized by very long range recurring patterns, that can be exploited to extract meaningful statistics. 
Several problems can be faced with these data, such as ($i$) the classification of a time series based on its characteristics; ($ii$) regression/prediction analysis  to test relationships between one or more different time series; ($iii$) time series forecasting, searching for a model capable of predicting future values of the series based on previously observed ones. 

\parafango{Classification} The \sigla{Eigenworms} dataset \cite{brown2013dictionary} is a collection of $259$ very long sequences, i.e., $\dseq=17,984$, describing the motion of a particular worm species. 
The kind of movement performed by these worms is a useful indicator for understanding behavioural genetics. In particular, these worms adopt certain shapes when placed on an agar plate, that can be represented by combinations of six base shapes, referred to as \textit{eigenworms}. The time series is constituted at each time step by six scalars representing the amplitudes along each dimension when the shape is projected onto the six eigenworms. The goal is to classify a worm as either wild-type or one among four mutant types, and has been tackled in  \cite{rusch2021unicornn,lim2023parallelizing,irie2022neural}.   
The \sigla{PhysioNet Sepsis} \cite{reyna2020early} is an  healthcare task, composed by an irregularly sampled time series with partially missing features. 
It consists in a binary prediction problem of sepsis, using a time series comprising $34$ medical features recorded during patients' stays in an ICU. Each sequence is accompanied by five static patient features (e.g., age). The sequences are short, with less than $72$ frames, and the data points are irregularly sampled with many missing entries, posing a challenge. There are two versions of the task: one with observation intensity information (OI) and one without (no-OI), which includes an extra input feature indicating the time stamp of each observation to provide information on measurement frequency. Tested in \cite{irie2022neural,kidger2020neural,morrill2021neural}.
Human activity recognition is performed in the \sigla{HAR-2} \cite{anguita2012human} dataset, a collection of tracked human activities measured by an accelerometer and gyroscope on a Samsung Galaxy S3 smartphone.  Six activities were binarized to obtain two merged classes (Sitting, Laying, Walking\_Upstairs) and (Standing, Walking, Walking\_Downstairs). \sigla{HAR-2} contains $\approx 7$k training sequence each of them composed by $128$ time steps represented with $\approx 1$k features.  
Exploited in \cite{kag2019rnns,rusch2020coupled}. 

\parafango{Regression/Prediction} The \sigla{BDIMC} healthcare datasets is composed by $7,949$ sequences, and it is aimed at predicting three vital signs of a patient. It is part of the TSR archive \cite{tan2020monash}, presenting clinical data from the Beth Israel Deaconess Medical Center. PPG and ECG signals were sampled with a frequency of $125$Hz for $8$ minutes each. The resulting two-dimensional sequences have a length of $\dseq=4000$. The goal is to predict a patient’s respiratory rate (RR), heart rate (HR), and oxygen saturation (SpO2), based on PPG and ECG signals. RMSE is exploited to measure performances \cite{rusch2021unicornn,gu2021combining}. 

The \sigla{Pendulum} benchmark 
\cite{becker2019recurrent,schirmer2022modeling} has been recently used \cite{smith2022simplified} to investigate models ability to handle observations received at irregular intervals. It is composed by a $\dseq=50$ long sequence of images having a resolution of $24 \times 24$ pixels, corrupted with a correlated noise process and sampled at irregular intervals from a continuous trajectory. 
The model is tasked to predict the sine and cosine of the angle of the pendulum, which follows a nonlinear dynamical system, without any explicit information on the  velocity. 

\parafango{Forecasting} The \sigla{Mackey-Glass} (MG) dataset \cite{mackey1977oscillation} tests the ability of a network to model chaotic dynamical systems. A sequence of one-dimensional observations—generated by solving the MG differential equations—are streamed as input, and the network is tasked with predicting the next $15$ values in the sequence, with an MG time-constant of $17$ steps. It is used in \cite{voelker2019,gu2020hippo}.
In the existing literature, the  prediction of long
sequence time-series, such as electricity consumption planning, is usually referred to as Long sequence time-series forecasting (\sigla{LSTF}). 
Among the most known datasets, the 
Electricity Transformer Dataset (\sigla{ETD}) \footnote{\scriptsize\url{https://github.com/zhouhaoyi/ETDataset}} contains the oil temperature and $6$ power load features of electricity transformers from different regions of a province of China, over a period of two years. The \sigla{ETT-small} split is composed by two dataset coming from two electricity transformers located in different regions, where data points have been recorded every minute (\sigla{ETTm1, ETTm2}) or hour (\sigla{ETTh1, ETTh2}). Each dataset contains $70,080$ data points with $8$ features, including the date, the predictive value ``oil temperature'', and $6$ different external power load features.
Electricity Consuming Load (\sigla{ECL})\footnote{\scriptsize\url{https://archive.ics.uci.edu/ml/datasets/ElectricityLoadDiagrams20112014}} is another benchmark which collects the electricity consumption of $321$ clients from 00:00 of January 1st 2012 to January 31st 2014, registered every $15$ min. Each column represents the energy consumption of a client over this period.
\sigla{Weather}\footnote{\scriptsize\url{https://www.ncei.noaa.gov/data/local-climatological-data/}}
records $11$ climate features in time intervals of one hour from January 1st 2010 to December 31st 2013.
Since \sigla{ETT}, \sigla{ECL} and \sigla{Weather} are obtained through real-world monitoring operations (via sensors), we will refer to them as \sigla{SensorData} in Table \ref{tab:benchs}, and remark that they have been exploited in several works in the context of this survey \cite{huang2022encoding,zhang2023effectively,gu2021efficiently}.

\subsection{Reinforcement Learning}

Sequential models are also sparingly tested  for sequential decision making in reinforcement learning tasks. There is not a unified testbed in this setting. Didolkar et al. \cite{didolkar2022temporal} evaluate the 
\sigla{BabyAI} \cite{chevalier2018babyai} benchmark, which provides a suite of environments where the agent has to carry out a given instruction (e.g., going to an object, placing an
object beside another one, opening a door with a key, etc.) in a partially-observable maze. The goal is to produce an autoregressive generative model that predicts actions conditioned on the past context.
Also the Atari benchmark \cite{chen2021decision} is frequently considered, exploiting a  causal mask and supervised training, to match the actions in the offline dataset conditioned on the future expected returns and the past history. 
Other heterogeneous reinforcement learning benchmarks are evaluated in \cite{gu2020improving,pramanik2023recurrent,irie2022modern}.

\subsection{Long Range Arena}

The previous sections highlighted the absence of an established common set of benchmarks to evaluate sequential models that handle long-range tasks. Indeed, when comparing the experimental setup of the previously described papers, it can be observed that there is not a consensus on a common testbed. Models are evaluated on heterogeneous tasks and datasets ; even when using the same dataset, different works often preprocess it in a different manner, making the comparison of their performances very difficult. Moreover, many of the benchmarks do not directly target long-range modeling ability. Finally, recent works fall short in decoupling the effect of large-scale pretraining from the inductive bias arising from a certain data distribution.

In an effort to standardize long-range models evaluation, Tay et al. \cite{tay2020long} proposed the \sigla{Long Range Arena} (\sigla{LRA}) benchmark suite. \sigla{LRA} includes synthetic and real-world tasks to assess the ability of sequential architectures to reason in such scenarios,  considering different types of data and conditions. It focuses on understanding how well architectures can handle long sequences having certain hierarchical or spatial structures, and aims to compare their performance across various settings. In details, it consists of $6$ tasks:
($i$) \sigla{Long ListOps}, ($ii$) \sigla{Text} ($iii$) \sigla{Retrieval} ($iv$) \sigla{Image} ($v$) \sigla{Pathfinder} and its extreme long version, ($vi$) \sigla{Path-X}.
Sequences constituting these task have a length $\dseq \in [1k, 16k]$, encompassing modalities and objectives that require similarity, structural, and visuospatial reasoning. The \sigla{LRA} suite is becoming increasingly popular in the scientific literature and has been exploited by several recent works mentioned in the context of this survey
\cite{gu2021efficiently,gupta2022diagonal,gu2022parameterization,smith2022simplified,hasani2022liquid,ma2023mega,qin2023hierarchically,QinSDLWLYKZ22,liu2022ecoformer,qin2022devil,peng2021random,zucchet2023online,orvieto2023resurrecting,gupta2022simplifying,li2022makes}.

\parafango{Long ListOps} \sigla{Long ListOps} is a longer version of the ListOps task proposed in~\cite{nangia2018listops}, aimed at assessing the model ability to process hierarchically structured data in a long-context scenario.       
In each sequence, some mathematical operators (e.g.,  {max}, {mean}, {median} and {sum\_mod}) and integer operands (range 0 to 9) are enclosed by delimiters (brackets) in  prefix notation. The goal is to compute the integer result of the mathematical expression, as exemplified by the following (short) sequence, 
      
      \begin{minipage}{\columnwidth}
      \scriptsize
      {
      \fontsize{8}{10}\fontfamily{pcr}\selectfont
    \textbf{INPUT}: [{MAX} 4 3 [{MIN} 2  3  ]  1  0  [{MEDIAN} 1  5  8 9, 2]] 
       ~~~~~ \textbf{OUTPUT}: 5
      }\end{minipage}

Input symbols are encoded via one-hot vectors, having $17$ unique possible configurations (opening brackets and operators
 are grouped into a single token). 
\sigla{Long ListOps} tests the ability to reason hierarchically while handling long contexts with varying length, up to $\dseq=2$k, with a reserved end-of-sequence token. The final goal is ten-way classification (the integer result of the expression). There are $96$k training sequences, $2$k
 validation sequences, and $2$k test sequences. 
      
\parafango{Text} \sigla{Text} is a byte-level text classification task,  based upon the \sigla{IMDb} binary sentiment classification previously described (i.e., classify whether a movie review is positive/negative). Text sequences are processed at byte/character-level (characters are encoded as one-hot vectors, with $129$ unique values possible), with a fixed max length of $\dseq=4$k, and truncated or padded when necessary. 
The evaluation metric is the accuracy. Compared to ListOps,   the model needs to reason with compositional unsegmented real-world data, with less defined boundaries, that must be learnt from data. Differently from character-level language modeling, where the local context is sufficient to infer the next character, here the model is required to grasp the overall sentiment of the processed text, hence it needs to compose characters into words and then higher-level concepts. 
There are $25$k training examples and $25$k test examples, while no
 validation set is provided. 

\parafango{Retrieval}  \sigla{Retrieval} can be used to assess the  model capability of encoding long sequences into representations suitable for similarity-based matching. Couples of texts are compressed and concatenated, then provided to a linear classifier. The text is sampled from the ACL Anthology Network \cite{radev2013acl} dataset.  {Given two textual citations, where characters are encoded as a sequence of integer tokens (one-hot vector with $97$ unique values), the task is to classify whether the two citations are equivalent.}
Byte/character level processing is performed, resulting in sequence length of $\dseq=4$k for each document (i.e., total text length $8$k for this task). The performances of this binary classification task are measured with accuracy. There are $147,086$ training pairs, $18,090$ validation pairs, and $17,437$ test pairs. 

\parafango{Image} The \sigla{Image} task corresponds to the already introduced \sigla{sCIFAR}, an image classification task on sequences of pixels, described in Section \ref{sec:compvis}.
A $32 \times 32$ pixels image is flattened in raster scan into a sequence of $1024$ pixels. The goal is to classify ($10$-way classification) the image by processing the 1D pixel sequence. Models are required to capture 2D spatial hierarchical structure between pixels even if they are processed sequentially. Images belong to the {CIFAR-10} \cite{krizhevsky2009learning} dataset, and are processed in a single gray-scale channel (each pixel represented with 8-bit pixel intensity, hence vocabulary size of $256$).\footnote{Some recent works process the coloured images instead \cite{orvieto2023resurrecting}.}.
There are $45$k training examples, $5$k validation examples, and $10$k test examples. 

\parafango{Pathfinder} \sigla{Pathfinder}  is a synthetic visual task  originally introduced in ~\cite{linsley2018learning,Kim2020Disentangling} for learning long-range spatial dependencies.
\begin{figure}
        \centering
        \includegraphics[width=0.25\columnwidth]{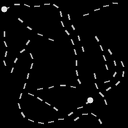}   
        \hskip 1.5cm \includegraphics[width=0.25\columnwidth]{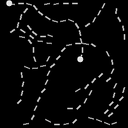}
        \caption{Negative (left) and positive (right) samples from \sigla{Pathfinder}. }
        \label{fig:path_finder}
    \end{figure}
A $32 \times 32$ grayscale image contains two points (a start and an end point, represented as small circles---see Figure \ref{fig:path_finder}). There are dashed lines over the image. 
The task is to make a binary decision whether the two points are connected by a path (Figure \ref{fig:path_finder}-left) or not (Figure \ref{fig:path_finder}-right). 
Models process the image pixel sequence (raster scan), thus $\dseq=1024$.  There are  $160$k training examples, $20$k validation examples, and $20$k test examples. 

\parafango{Path-X} Finally, \sigla{Path-X} is an harder version of \sigla{Pathfinder} composed by $128 \times 128$ images, resulting in sequences of $\dseq=16$k.

\subsection{Empirical Comparison: Transformers vs. SSMs}\label{sec:exp-comparison}

{We now provide an empirical comparison between attention-based Transformers and \sigla{SSMs} across a range of synthetic and real-world sequence modeling tasks. By evaluating performance on memory-intensive benchmarks like the \sigla{Copy} task, \sigla{Induction Heads}, \sigla{Associative Recall}, and \sigla{LRA}, the analysis highlights where each architecture excels or falls short, especially as sequence lengths grow or memory constraints tighten.}

\parafango{Copy task}
{The synthetic \sigla{Copy} task \cite{hochreiter1997long} remains the most direct probe of a model’s raw memory capacity: the network is asked to reproduce an input string
verbatim after a sentinel token. Recent results demonstrate a decisive
advantage for attention-based Transformers over \sigla{SSMs} on both data-efficiency and length
generalisation. Figure~\ref{fig:transformers-vs-ssm-copy}, revisited from \cite{jelassi2024repeat}, contrasts the strongest publicly reported results
for Transformers and \sigla{SSMs}.  \sigla{Hard-ALiBi} \cite{jelassi2024repeat}, a Transformer architecture exploiting positional encodings that bias query-key attention scores with a penalty  proportional to their distance, reaches
$\ge 95\%$ string-level accuracy on length-300 sequences after only
$1\times10^{4}$ training examples; conversely,  \sigla{Mamba-360M} needs two orders of magnitude more (Figure~\ref{fig:transformers-vs-ssm-copy}--left).}\footnote{The authors use their standard 300-token setup
to keep compute and optimisation schedules identical across models.} {When evaluated on out-of-distribution lengths (train on~$\leq50$,
test on~300\,/\,1000), the Transformer preserves $\geq90\%$ accuracy, while
\sigla{Mamba} and other gated \sigla{SSMs} collapse to chance
(Figure~\ref{fig:transformers-vs-ssm-copy}--right).}

\begin{figure}
\centering

\begin{minipage}[t]{0.48\linewidth}
\centering
\begin{tikzpicture}
\begin{axis}[
width=4.4cm,
height=3.4cm,
xlabel={\footnotesize{\# Training examples}},
ylabel={\footnotesize{Accuracy (\%)}},
xmode=log,
log basis x=10,
xmin=8000, xmax=2000000,
ymin=0, ymax=105,
xtick={1e4,1e5,1e6},
xticklabels={$10^4$, $10^5$, $10^6$},
ytick={0,25,50,75,100},
legend pos=south east,
grid=major
]
\addplot+[smooth, thick, color=red!60!brown, mark=*, mark options={fill=red!60!brown, draw=red!60!brown}] coordinates {(8000,0) (10000,0) (50000,0) (1e5,0) (1e6,10) (1.7e6,93) (1.9e6,92) (2e6,100)};

\addplot+[smooth, thick, color=blue!80!black, mark=asterisk,
mark options={fill=blue!80!black, draw=blue!80!black}] coordinates {(8000,0) (10000,5) (20000,90) (30000,85) (50000,97) (70000,95) (1e5,100) (1e6,100) (2e6,100)};
\end{axis}
\end{tikzpicture}
\end{minipage}
\hfill
\begin{minipage}[t]{0.48\linewidth}
\centering
\begin{tikzpicture}
\begin{axis}[
    width=4.4cm,
    height=3.4cm,
    xlabel={\footnotesize{\# Characters in string}},
    ylabel={\footnotesize{Accuracy (\%)}},
    xmode=log,
    log basis x=10,
    xtick={50,200,1000},
    xticklabels={50,200,$10^3$},
    ytick={0,25,50,75,100},
    ymin=0, ymax=105,
    xmin=0, xmax=1000,
    grid=major,
    legend style={at={(0.5,-0.25)}, anchor=north, legend columns=3}
]

\addplot[dashed, thick, color=purple!30!black] coordinates {(50,0) (50,100)};
\addplot[dashed, thick, color=green!30!black] coordinates {(200,0) (200,100)};

\addplot+[smooth, thick, color=red!60!brown] coordinates {
    (40,100) (50,100) (60,90) (70,60) (80,10) (90,0) (100,0) (200,0) (500,0) (1000,0)
};
\addplot+[smooth, thick, color=blue!80!black] coordinates {
    (50,100) (100,100) (150,100) (200,99) (300,98) (400,97) (800,95) (1000,93)
};
\end{axis}
\end{tikzpicture}
\end{minipage}

\vspace{0.5em}
\begin{tikzpicture}
\begin{axis}[
hide axis,
xmin=0, xmax=1,
ymin=0, ymax=1,
legend columns=2,
legend style={at={(0.5,-0.2)}, anchor=north, draw=none}
]
\addlegendimage{smooth, thick, color=red!60!brown, mark=*, mark options={fill=red!60!brown, draw=red!60!brown}}
\addlegendentry{\sigla{Mamba-360M}}
\addlegendimage{smooth, thick, color=blue!80!black, mark=asterisk, mark options={fill=blue!80!black, draw=blue!80!black}}
\addlegendentry{\sigla{Hard-ALiBi}}
\end{axis}
\end{tikzpicture}

\caption{{Revisited from \cite{jelassi2024repeat}. (Left) The authors train models to copy strings of length $\leq 300$ and evaluate string-level accuracy on strings of length $300$. Transformers train much faster than \sigla{SSMs}. Indeed, the \sigla{Mamba-360M} needs two orders of magnitude more w.r.t. \sigla{Hard-ALiBi} Transformer. (Right) The authors train models to copy on strings of length $\leq 50$ until all models are perfect in-distribution and evaluate string-level accuracy. Purple dashed line indicates maximum training string length and green dashed line indicates context window during training. Evaluating on longer inputs, the transformer models dramatically outperform the \sigla{SSMs.}}}
\label{fig:transformers-vs-ssm-copy}
\end{figure}


\parafango{Induction Heads and Associative Recall} 
{Synthetic memory benchmarks such as \sigla{Induction Heads} (IH) and \sigla{Associative Recall} (AR) provide a minimal yet revealing playground for analysing how sequence models store and retrieve information. Because the training context is kept short while the evaluation context can reach millions of tokens, these tasks expose the limits of each architecture’s memory mechanism. We report in Table~\ref{tab:ih_ar_results} the results in the original papers \cite{dao2022hungry,gu2023mamba}.  All models are two‑layer variants trained on length-$256$ sequences for $2\times 10^{5}$ updates and evaluated on powers of two up to $L=1,048,576$ tokens. These tasks expose three qualitatively different memory regimes that map cleanly onto the architectures in Table~\ref{tab:ih_ar_results}.} 
{Plain linear \sigla{SSMs} such as \sigla{S4D} and \sigla{GSS} compress the entire past into a fixed-width latent state. As the number of distinct pairs approaches that width the representation collides, which explains the sharp drop to $\approx35\%$ accuracy on IH even for 256--token contexts. \sigla{H3} augments the recurrent path with a learned \emph{shift} channel so that every incoming token is explicitly available in the next state's slot, and combines this with multiplicative gating. The resulting mechanism behaves like a differentiable hash table that stores each token exactly once while using the gate for equality tests. This prior yields $99.8\%$ accuracy on AR and perfect IH, with the remaining $0.2\%$ error attributable to rare collisions when two identical keys land in the same step. \sigla{Mamba} replaces fixed convolutions with input--dependent transition matrices. By deciding \emph{at runtime} which information to propagate, \sigla{Mamba} mimics the selective read--write behaviour of attention yet keeps both time and memory linear in $L$. Consequently it sustains $100\%$ accuracy from 64 up to $1{,}048{,}576$ tokens.}
{When GPU memory allows, Transformers remain the simplest drop--in solution and excel at many-to-many interactions. \sigla{H3} and especially \sigla{Mamba} become preferable for million-token contexts where linear memory is a hard requirement.}

\begin{table}[ht]
\centering
\caption{{Sequence‑level accuracy on Induction Heads (IH) and Associative Recall (AR)--results taken from \cite{dao2022hungry,gu2023mamba}.}}
\label{tab:ih_ar_results}
{
\begin{tabular}{p{2.8cm}|p{2cm}|p{1cm}|p{1cm}}

\toprule
Model & Parameters & IH & AR \\
\midrule
Transformer \cite{vaswani2017attention} & 0.10M & 100\% & 100\% \\
S4D \cite{gu2021combining} & 0.10M & 35.6\% & 86.0\% \\
GSS \cite{mehta2022long} & 0.10M & 6.8\% & 78.0\% \\
H3 \cite{dao2022hungry} & 0.15M & 100\% & 99.8\% \\
Mamba \cite{gu2023mamba} & 0.07M & 100\% & 100\% \\
\bottomrule
\end{tabular}}
\end{table}

\parafango{Long Range Arena}
{The \sigla{LRA} benchmark \cite{tay2020long} comprises six tasks with sequence
lengths ranging from \(1{,}024\) to \(16{,}384\) tokens, as described above. We compare the classical Transformer baseline against different \sigla{SSMs} architectures, reporting results from \cite{smith2022simplified,orvieto2023resurrecting} in Table~\ref{tab:lra}.}
\begin{table}[t]
  \centering
  \caption{{Exact test accuracy on the six Long-Range Arena tasks. Results are taken from table of the original papers \cite{smith2022simplified,orvieto2023resurrecting}.}}
  \label{tab:lra}
  {\begin{tabular}{l|p{1cm}|p{1cm}|p{1cm}|p{1cm}}
    \toprule
    \textbf{Task (length)} &
    \textbf{Trans.} &
    \textbf{Mega} &
    \textbf{S5} &
    \textbf{LRU} \\[2pt]
    \midrule
    ListOps (2 k)        & 36.37\% & 63.14\% & 62.15\% & 60.2\% \\
    Text (4 k)           & 64.27\% & 90.43\% & 89.31\% & 89.4\% \\
    Retrieval (4 k)      & 57.46\% & 91.25\% & 91.40\% & 89.9\% \\
    sCIFAR-10 (1 k)      & 42.44\% & 90.44\% & 88.00\% & 89.0\% \\
    Pathfinder (1 k)     & 71.40\% & 96.01\% & 95.33\% & 95.1\% \\
    Path-X (16 k)        & 7.00\% & 97.98\% & 98.58\% & 94.2\% \\
    \bottomrule
  \end{tabular}}
\end{table}
{Three main trends emerge from Table~\ref{tab:lra}. \sigla{Mega}, which augments a standard self-attention block with a diagonal state-space kernel, delivers the top score on four of the six tasks—peaking at \(91.25\%\) on the document–pair \sigla{Retrieval} task and \(96.0\%\) on \sigla{Pathfinder}. The classical Transformer, by contrast, collapses to chance-level accuracy (7 \%) on the 16k-token \sigla{Path-X} and lags by 20–50 pp everywhere else. \sigla{S5} dominates the very long regime. When sequence length jumps to 16k, the continuous-time \sigla{SSM} of \sigla{S5} reaches \(98.58\%\) on \sigla{Path-X}. It also ties or edges Mega on the image and \sigla{Pathfinder} tasks while keeping \(\mathcal{O}(L)\) memory and compute. \sigla{LRU} shows how far a linear RNN can go. With a purely recurrent core, stable-exponential parameterisation and a single $\gamma$-norm, \sigla{LRU} matches \sigla{S5} within one percentage point on five tasks and even surpasses it on \textsc{sCIFAR-10} (89.0 \% versus 88.0 \%).}
{Taken together, these results suggest a clear trade-off. If memory footprint is no object, the \sigla{Mega} transformer still gives the best average accuracy. For workloads dominated by \(\ge\!10\,000\)-token sequences or tight latency budgets, however, \sigla{S5} and the much simpler \sigla{LRU} offer comparable accuracy with strictly linear scaling and constant-time autoregressive generation.}

{In summary, Transformers outperform \sigla{SSMs} on data efficiency and generalization in the \sigla{Copy} task, requiring far fewer samples and maintaining high accuracy on longer sequences. In synthetic memory tests (IH, AR), advanced \sigla{SSMs} like \sigla{H3} and \sigla{Mamba} match Transformer performance while scaling linearly with sequence length. On \sigla{LRA} benchmarks, \sigla{Mega} leads in accuracy, but \sigla{SSMs} like \sigla{S5} and \sigla{LRU} offer competitive results with better efficiency, especially on very long sequences. Overall, Transformers are ideal when resources allow, while modern \sigla{SSMs} are better suited for long-context, resource-constrained settings.}

\begin{table*}[h!]
\centering
\caption{Sequence processing benchmarks. A list of the most common benchmarks (first column) and papers exploiting them (other columns). Each column report papers described in a different sections of this survey.}
\vskip 1mm
\footnotesize
\begin{tabular}{cc|p{3cm}|p{3cm}|p{3cm}}
\toprule
& Benchmarks & {Transformers embracing Recurrence (Sec.~\ref{sec:transformers})} & {Deep State-Space models (Sec.~\ref{sec:ssm})} & {Novel \sigla{RNNs} architectures (Sec.~\ref{sec:other_rnns}--\ref{sec:forward})} \\
\midrule
\multirow{4}{*}{\rotatebox{90}{\scriptsize\textsc{Synthetic}}} & Adding Problem \cite{hochreiter1997long}  & \cite{le2015simple,arjovsky2016unitary} & - & \cite{li2018independently,gu2020improving,rusch2020coupled,rusch2021long,kag2021training} \\
& Copy Task \cite{hochreiter1997long} & \cite{arjovsky2016unitary,menick2020practical,ba2016layer,bulatov2022recurrent,didolkar2022temporal} & \cite{gu2020hippo,zucchet2023online,de2024griffin,gu2023mamba} & \cite{gu2020improving,kerg2019non,menick2020practical, irie2023exploring, park2023persistent} \\
& Induction Heads &  & \cite{de2024griffin,gu2023mamba} & - \\
& Associative Recall & \cite{yang2023gated,bulatov2022recurrent,peng2024eagle} & \cite{dao2022hungry,gu2023mamba,polihyena}  & \cite{beck2024xlstm,liao2018reviving}  \\ 
\midrule
\multirow{6}{*}{\rotatebox{90}{\scriptsize\textsc{\hskip -0.4cm Computer Vision}}} & \sigla{sMNIST} & \cite{voelker2019,gu2020hippo,gu2021combining,gu2021efficiently,smith2022simplified}&  \cite{gu2020improving,kerg2019non,rusch2020coupled,limnoisyrnn,erichson2020lipschitz,rusch2021long} & \cite{kag2021training,ororbia2019biologically} \\
& \sigla{psMNIST} & & & \cite{rusch2020coupled,rusch2021unicornn,erichson2020lipschitz,rusch2021long,kag2021training, park2023persistent}  \\
& \sigla{genMNIST} & \cite{katharopoulos2020transformers} &  & \cite{ororbia2020continual} \\
& CIFAR & & & \cite{lim2023parallelizing,gu2020improving,kag2021training} \\ 
& \sigla{npCIFAR}  & & &  \cite{rusch2020coupled,rusch2021unicornn,rusch2021long} \\
& ImageNet & \cite{irie2022modern,liu2022ecoformer,choromanski2020rethinking}& \cite{li2022makes,polihyena}& \cite{qin2023hierarchically}  \\

\midrule
\multirow{12}{*}{\rotatebox{90}{\scriptsize\textsc{\hskip -0.7cm Natural Language}}} & WikiText103  \cite{merity2016pointer}&\cite{peng2021random,irie2021going,schlag2021linear,qin2022devil,huang2022encoding,dai2019transformer,bulatov2022recurrent,LiLL0LZW023,kasai2021finetuning,mao2022fine,QinSDLWLYKZ22,wu2022memformer,katsch2023gateloop,qin2023hierarchically} & \cite{gu2021efficiently,ma2023mega,li2022makes,polihyena}& \cite{gu2020improving,beck2024xlstm,menick2020practical} \\

& \sigla{PTB}\cite{marcus1993building} & \cite{dai2019transformer}& & \cite{kerg2019non,rusch2021long,li2018independently,bradbury2016quasi,kag2019rnns,kag2021training} \\
& \sigla{The Pile}\cite{gao2020pile} & \cite{sun2023retentive,bulatov2023scaling,qin2023hierarchically,peng2023rwkv}&\cite{polihyena,gu2023mamba} & \\
& \sigla{PG-19} \cite{rae2019compressive}& \cite{pilault2023block, hutchins2022block, wu2022memformer,peng2024eagle} & & \\ 
& \sigla{Enwik8} \cite{mahoney2011large} & \cite{zhai2021attention,huang2022encoding,dai2019transformer,bulatov2022recurrent} & & \cite{lei2018simple} \\
& ArXiv \& GitHub \cite{wu2022memorizing} & \cite{hutchins2022block,pilault2023block,LiLL0LZW023}& &  \\
&\sigla{LM-Eval-Harness} \cite{leogao} & \cite{yang2023gated,peng2023rwkv,qin2023scaling,sun2023retentive,peng2024eagle}& \cite{gu2023mamba,de2024griffin} & \cite{beck2024xlstm} \\
&\sigla{WMT}\cite{bojar2014findings,kocmi2022findings} & \cite{peng2021random,schlag2021linear,hao2019modeling,chen2019recurrent,kasai2021finetuning,chen2018best} & \cite{ma2023mega} &  \\
&\sigla{IWSLT}\cite{cettolo2014report} & \cite{peng2021random}& & \cite{bradbury2016quasi}  \\ 
&\sigla{IMDb}\cite{maas2011} & \cite{QinSDLWLYKZ22} & \cite{gu2020hippo} & \cite{rusch2020coupled,rusch2021unicornn,bradbury2016quasi}  \\
\midrule
\multirow{5}{*}{\rotatebox{90}{\scriptsize\textsc{Time Series}}}  & \sigla{Eigenworms} \cite{brown2013dictionary} & & & \cite{irie2022neural,rusch2021unicornn,lim2023parallelizing} \\ 
& \sigla{HAR-2} \cite{anguita2012human} & & \cite{kag2019rnns,rusch2020coupled}& \\ 
& \sigla{Mackey-Glass} \cite{mackey1977oscillation} & & \cite{voelker2019,gu2020hippo} & \\ 
& \sigla{BDIMC}\cite{tan2020monash} & & \cite{gu2021combining} & \cite{rusch2021unicornn}\\
& \sigla{SensorData} & \cite{huang2022encoding} & \cite{gu2021combining,zhang2023effectively} & \\
\midrule
& \sigla{Speech Commands} \cite{warden2018speech} & & \cite{gu2021combining,gu2021efficiently,gupta2022diagonal,gu2022parameterization,smith2022simplified} & \cite{rusch2021long,kidger2020neural,kag2019rnns} \\ 
\midrule
& \sigla{Long Range Arena} \cite{tay2020long} & \cite{QinSDLWLYKZ22,liu2022ecoformer,qin2022devil,peng2021random} & \cite{zucchet2023online,orvieto2023resurrecting,gupta2022simplifying,li2022makes,gu2022parameterization,gu2021efficiently,gupta2022diagonal,smith2022simplified,hasani2022liquid,ma2023mega} & \cite{qin2023hierarchically,beck2024xlstm} \\
\bottomrule
\label{tab:benchs}
\end{tabular}
\end{table*}

\section{Discussion and Future Directions}
\label{sec:discus}
In this Section, we analyze some open problems and issues that have been recently pointed out in the works described in this survey, as well as highlighting possible future avenues for research. \added{Despite the individual strengths of the architectures surveyed in previous sections, it is now increasingly evident that no single paradigm—be it pure Transformer, \sigla{RNN}, or \sigla{SSM}—can fully address the challenges posed by long-sequence modeling. Transformers offer excellent performance and global receptive fields, but suffer from quadratic complexity and require full sequence availability. \sigla{RNNs} and \sigla{SSMs}, on the other hand, offer efficient online processing and constant memory per time step, but are limited in terms of expressivity. These complementary strengths and weaknesses have naturally motivated the exploration of \textit{hybrid} models that aim to combine the parallelism and global context handling of Transformers with the recurrence and efficiency of \sigla{RNNs} and \sigla{SSMs}. This section builds on this perspective, outlining both open issues and emerging design trends that point toward unified architectures capable of overcoming the fundamental trade-offs of current solutions.
}


\subsection{Limitations of Transformers and Recurrent Models}
\label{sec:expressivitygap}

{The previous sections explored recent progress in Transformers and \sigla{SSMs}, highlighting the resurgence of recurrent mechanisms. Future research directions should be informed by an understanding of the current limitations of these model classes. There is potential for trade-offs and hybrid approaches, alongside a need to co-design models with the computational hardware they target.}

\parafango{{Transformers}}
{A series of recent studies have evaluated Transformers across various tasks to uncover their limitations. Hanh et al. \cite{hahn2020theoretical} demonstrated that Transformers struggle with recognizing simple patterns, such as parity or properly nested parentheses, when faced with long inputs. Peng et al. \cite{peng2024limitations} further investigated these limitations, focusing on function composition. For instance, tasks involving multi-step reasoning over several facts remain challenging for Transformers. Although chain-of-thought prompting can alleviate some issues, it substantially increases the number of required tokens.
Furthermore, Transformers have shown shortcomings in time series forecasting tasks, where simpler architectures have outperformed them \cite{zeng2023transformers}.
Early theoretical work established that, under idealized assumptions such as infinite‑precision arithmetic and arbitrarily powerful feed‑forward subnets, transformers are “Turing complete” and can simulate a general‑purpose machine \cite{perez2019turing}. However, once we impose realistic constraints—most critically, limiting attention to hard (discrete) patterns—their power collapses into well‑studied Boolean circuit classes. Boolean circuits are acyclic networks of logical gates (e.g., AND, OR, NOT) whose depth and size determine how “parallel” or “sequential” a computation can be. Using circuit‐complexity techniques, Hahn \cite{hahn2020theoretical} and Hao et al.\ \cite{hao2022formal} showed that transformers with hard attention correspond to non‑uniform $\mathrm{AC}^0$ circuits, which cannot even compute the majority function on $n$ bits. Allowing softer attention extends them to L‑uniform $\mathrm{TC}^0$ (the class of constant‑depth, polynomial‑size threshold circuits constructible in logarithmic space--informally, it can be
thought of as the class of problems that can be solved with
extremely parallel constant-depth computation), but Merrill \& Sabharwal \cite{merril2023,merrill2023logic} demonstrated that even this model cannot express inherently sequential computations. In particular, $\mathrm{NC}^1$‑hard state‑tracking problems—such as permutation composition—fall outside $\mathrm{TC}^0$ and thus lie beyond the reach of any realistic transformer model \cite{liu2022transformers}. Permutation composition underpins fundamental tasks like playing chess, evaluating code, or tracking entities in a narrative; since these problems require carrying and updating an internal state step by step, transformers constrained by finite‑precision or limited attention cannot solve them in the worst case.
}

\parafango{{Recurrent models}} Despite the computational advantages of recurrence-based models, performances are not always on par with respect to vanilla attention when considering specific tasks, also in synthetic settings \cite{dao2022hungry,jelassi2024repeat}. Indeed, the capability of recurrent models to effectively solve a target task is strongly conditioned upon how efficiently the processed sequence is \textit{compressed} into a finite-size state \cite{gu2023mamba}, which acts as an information bottleneck. 

The role of \textit{gating} mechanisms turns out to be fundamental, given their influence on the model ability to control, in a context/input dependent manner, how information propagates or interacts along the temporal dimension. 
In fact, the relevance of an appropriate gating mechanism can be appreciated when considering that such a mechanism is present in many high-impact works described in this survey \cite{jing2019gated,yang2023gated,zucchet2023gated,schlag2021linear,irie2021going,peng2023rwkv,sun2023retentive,katsch2023gateloop}, going well beyond the original case of \sigla{LSTMs}.
Indeed, the \sigla{Mamba} model \cite{gu2023mamba} shows how a \textit{selection} mechanism (i.e., a term which refers to the mechanistic action of a model to select or ignore inputs and facilitate data interaction along
the sequence length, inspired by the original gating in \cite{hochreiter1997long}) improves the capability of State-Space Models to solve tasks such as \sigla{Selective copy} or \sigla{Induction Heads} (see Section \ref{sec:benchmarks}). In these task,  content-aware reasoning is fundamental to be able to memorize relevant tokens and filter out the irrelevant ones. While the Linear Time Invariant (\sigla{LTI}) nature of previous \sigla{SSMs} (\sigla{LSSL, S4} \cite{gu2021combining,gu2021efficiently}) allows to achieve efficient parallelizable solutions, their dynamics do not easily allow to select the correct information from the current context, making it harder to encode the information stored in the hidden state. Conversely, an appropriate selection mechanism introduces context-awareness, allowing the model to focus on or to filter out information.
This element is also interesting in light of recent studies (see Section \ref{sec:other_rnns}) suggesting that certain \sigla{RNNs} (equipped with linear recurrent layers interconnected by feedforward paths with multiplicative gating) might be implementing attention/selection ``under the hood'', when trained to solve simple in-context learning tasks \cite{zucchet2023gated}.

However, despite such advantages brought by gating/selection mechanisms, 
the performance of recurrence-based solutions  still falls short of fully closing the gap with attention-based models.
Several recent works investigated such a recurrence \textit{expressivity gap} \cite{dao2022hungry,jelassi2024repeat} with respect to attention. Indeed, inference in Transformers leverages the entire sequence processed so far, which, in autoregressive models, is stored in the \sigla{KV-cache} (see Section \ref{sec:transformers}), to produce the final output. By definition, the memory usage of this approach increases over time, which becomes extremely expensive. 
At the cost of inefficiency, the attention mechanism possesses properties that standard stateful models are not able to embrace: explicitly recalling earlier tokens in the sequence (which influences their ability to copy from the input sequence, possibly by multiplying the attention scores with $V$) and comparing tokens across the sequence (which is achievable thanks to the quadratic attention matrix $QK'$)).
Jelassi et al. \cite{jelassi2024repeat} exploit the copy task (see Section \ref{sec:benchmarks}) to theoretically and empirically prove this property. 
In short, they show that from the theoretical point of view a simple Transformer model could be capable (when empowered by proper positional encodings and under some other mild assumptions) of copying strings of length that are exponential in the number of heads of the Transformer. This result relies on the ability of the Transformer to implement a mechanism of “storage” and retrieval of sequences of $n$ tokens ($n$-grams). Noticeably, the number of  parameters depends only logarithmically on the input sequence length.
The authors show that, conversely, any state-space model fails to solve the copy task, unless its latent state grows linearly with the sequence length. That is, \sigla{SSMs} cannot accurately copy strings with more bits than the size of the latent state.

Such theoretical findings have been confirmed empirically. Indeed, Transformers are both much more efficient at learning to copy (indeed, they need $\approx 10^4$ training examples to learn the task, whereas \sigla{Mamba} \cite{gu2023mamba} needs more than $\approx 10^6$ samples, as mentioned in Section \ref{sec:exp-comparison}) and also generalize better to longer inputs (when trained on strings $\dseq=50$, they generalize to sequences with $\dseq>1000$, whilst \sigla{Mamba} fails to generalize out of the trained context). This finding confirm the theoretical results, i.e., to be able to copy the entire input sequence, stateful models need to fully store it in its state space, which requires the memory to grow linearly with the sequence length.
These results hold even when considering pre-trained models, both in copy tasks and retrieval benchmarks (e.g., \sigla{Phonebook Lookup}, see Section \ref{sec:benchmarks}).  

These results are interesting in light of the fact that the \textit{selection mechanism} underlying \sigla{Mamba} (see Section \ref{sec:ssm})  was introduced specifically to be able to selectively propagate or forget information along the temporal dimension, depending on the current token.  
Similar intuitions have previously inspired the work of Dao et al. \cite{dao2022hungry}, who proposed the \sigla{H3} model. As we detailed in Section \ref{sec:ssm}, their intuition on reducing the attention-recurrence expressivity gap is to empower the copy and retrieval ability of \sigla{SSMs} by ($i$) adding a layer with a shift matrix, that allows the state to copy from the input and pass that information to the next state, ($ii$) stacking two SSMs with multiplicative interaction, in order to compare information from previous time steps (output of the lower shifted \sigla{SSMs}) with the input at the current time steps -- thus measuring similarity between tokens, which mimics local multiplicative interactions in linear attention. 
Noticeably, the combination of shift \sigla{SSM} and the following multiplicative interaction acts as a gate. Empirically, these solutions allow to completely solve the \sigla{Induction Heads} task and almost completely fit the \sigla{Associative Recall}, while other \sigla{SSMs} (e.g., \sigla{S4D} \cite{gu2022parameterization}, \sigla{GSS} \cite{mehta2022long}) struggle with it.

These findings show that recurrent architectures working in the state-space {still} have room for improvement in terms of representational capability when compared to attention-based methods.
{Merril et al.~\cite{merrillIllusionStateStateSpace2024} showed that, similarly to Transformers, \sigla{RNNs} with a diagonal
transition matrix (thus, most of deep \sigla{SSMs}) could only represent functions in TC\textsuperscript{0}. 
They also showed that the expressive power of \sigla{SSMs} can be extended by making the state transition matrices input-dependent, enabling the theoretical capability to solve NC\textsuperscript{1}-hard problems and achieving better empirical results, as evidenced by models like \sigla{Liquid-S4} \cite{hasani2022liquid}. 
Grazzi et al.~\cite{grazzi2025unlocking} demonstrate that typical linear \sigla{RNNs} like \sigla{Mamba} and \sigla{DeltaNet} are limited by positive-only eigenvalues in their state-transition matrices, which prevents them from solving simple tasks such as parity. By allowing negative eigenvalues and involving non-diagonal state transition matrices, they significantly improve performance on state-tracking, while maintaining stability and efficiency at scale.}
{Concurrently, \sigla{RWKV-7}~\cite{peng2025rwkv} demonstrates remarkable expressivity, with the ability to recognize all regular languages using a small constant number of layers. This is grounded in the architecture's generalized delta rule, which incorporates vector-valued gating and in-context learning rates. These advancements build upon and extend the capabilities introduced in \sigla{Gated DeltaNet}~\cite{yang2025gateddeltanetworksimproving}, a linear recurrent neural network architecture that combines the strengths of \sigla{Mamba2}’s gating mechanism for adaptive memory decay and \sigla{DeltaNet}’s delta update rule for precise key-value memory updates.}
{Building on \cite{grazzi2025unlocking}, Siems et al.~\cite{siems2025delta} propose \sigla{DeltaProduct}, a generalization of \sigla{DeltaNet} where each token update applies a product of multiple Householder transformations. This increases the rank of the state-transition matrix, offering a tunable balance between expressivity and efficiency.}

\parafango{{Overcoming limitations with hybrid models}}
{The expressivity gap emphasized by Jelassi et al. \cite{jelassi2024repeat} highlights why \sigla{SSM}-based architectures still lag behind attention-only models like \sigla{Mistral 7B} \cite{jiang2023mistral} in demanding tasks such as language modeling. To address this, recent models like \sigla{Griffin} \cite{de2024griffin} incorporate hybrid strategies, blending \sigla{SSMs} with local (sliding window) attention. This form of attention enables each token to access a limited context of past tokens, reducing FLOPs and constraining KV-cache sizes.}
In practice, local attention allows each position to attend only to a fixed number of tokens in the past, reducing the number of FLOPs with respect to standard attention and bounding the size of the \sigla{KV-cache} to the size of window, making it no longer quadratic in the sequence length. \sigla{Griffin} matches the learning speed of Transformers (in terms of necessary training steps/samples to achieve a certain performance), and is capable to extrapolate to evaluation sequences several orders of magnitude longer than the training sequence lengths.
{Other architectures have adopted different hybrid schemes, layering attention and \sigla{SSMs} in alternation. For instance, \sigla{Jamba} and \sigla{Jamba 1.5} \cite{team2024jamba} propose a Transformer-\sigla{Mamba} mix, combined with a mixture-of-experts module, to achieve a balance between performance and computational cost.
Poli et al. \cite{polimechanistic} conducted a comprehensive analysis and found that a compute-optimal configuration includes roughly one-fourth self-attention layers, with the remainder composed of SSMs.
\sigla{Zamba} \cite{glorioso2024zamba}, a 7B hybrid model, demonstrates strong performance relative to other open-weight models by pairing a \sigla{Mamba} backbone with a shared attention layer, thereby capturing the benefits of attention while keeping parameter usage minimal.}

Moreover, it is of extreme interest to understand the role of model pre-training. Indeed, Amos et al. \cite{amos2023train} suggested that self-supervised pretraining on the task data (referred to as \textit{self pretraining}), rather than training from scratch, significantly improves the performance on long sequences (i.e., on \sigla{LRA}) both in vanilla Transformers and with simple diagonal \sigla{RNNs}, showing the redundancy of sophisticated architectures and manually-designed biases \cite{gu2022parameterization,ma2023mega}.

\subsection{Hardware-aware Efficiency: FLOPs vs. Memory Bandwidth}
\label{sec:flopsvsmem}


Reducing the computational and memory requirements of vanilla Transformers is one of the main goal of the works described in this survey. In Section \ref{sec:transformers} we showcased several attention approximations that indeed succeed in this goal, achieving complexities linear or near-linear in sequence length \cite{katharopoulos2020transformers,schlag2021linear,peng2023rwkv} by reducing the amount of computations. However, when used in practice, many of these solutions do not display substantial wall-clock speedup when compared to standard attention.  
Recent works such as \sigla{FlashAttention} \cite{dao2022flashattention,dao2023flashattention} showed how floating-point operations (\sigla{FLOPs}) reduction may not correlate with fast wall-clock speed, when memory access overheads are not appropriately taken care of. 

This is a direct consequence of the way the hardware of common accelerators (e.g., GPUs, TPUs, etc.) is structured. For instance, GPUs compute speed has out-paced memory speed. When considering  Transformers, most operations are bottlenecked by memory accesses \cite{dao2022flashattention,ivanov2021data}. Generally speaking, accelerators perform reads and writes to different levels of fast and slow memory, i.e., between fast GPU on-chip \sigla{SRAM} and relatively slow GPU high bandwidth memory (\sigla{HBM}). 
As a key example, the \sigla{NVIDIA A100} is composed by 40-80 GB of \sigla{HBM}, with bandwidth of 1.5-2.0 TB/s and 192KB of on-chip SRAM
per each of 108 streaming multiprocessors (\sigla{SMs} -- that execute groups of thread, referred to as thread blocks, in parallel) with bandwidth estimated around 19 TB/s \cite{jia2021dissecting,dao2022flashattention}.  Each thread loads input from \sigla{HBM} to registers and on-chip SRAM for computation before writing back to \sigla{HBM}. Memory-bound operations (e.g., elementwise sum, reductions, softmax, batch norm, layer norm) often pose significant bottlenecks due to slower memory speeds.
Vanilla Transformers store tensors such as weights and the \sigla{KV-cache} in \sigla{HBM} for the duration of the decoding. These tensors need to be
transferred from \sigla{HBM} to the compute cores of the chip once per forward pass of the model, thus giving rise to a large amount of memory traffic.


Additionally, specialized hardware units, such as  TensorCores in NVIDIA GPUs \cite{markidis2018nvidia} and MXUs in Google TPUs \cite{norrie2021design}, are optimized for classic architectures relying on matrix multiplications and convolutions with high FLOPs-to-byte ratios. While \sigla{RNNs} benefit from this due to their dense recurrence matrices, operations happening in diagonal RNNs (see \sigla{SSMs} and other diagonal models described in Sections \ref{sec:ssm} and \ref{sec:other_rnns}) pose computational challenges on existing accelerators. For instance, the element-wise update equation in \sigla{Griffin} results in a low FLOPs-to-byte ratio, leading to memory-bound computations.

Many recent works \cite{gu2021combining,yang2023gated,katsch2023gateloop,sun2023retentive} propose alternative computational forms (i.e., recurrent, parallel, chunkwise) for their models, offering varying tradeoffs which can be chosen according to specific needs, but still not taking into account the I/O cost. 
Whilst basic implementations of the recurrent forms minimize the total FLOPs, they incur in high I/O costs due to storing 2D hidden states in \sigla{HBM} \cite{mao2022fine}.  Katharopoulos
et al. \cite {katharopoulos2020transformers} reduce this cost by avoiding state materialization and recomputing hidden states during the backward pass (hereinafter referred to as \textit{recomputation}), while Katsch \cite{katsch2023gateloop} leverages the parallel scan algorithm to parallelize a recurrent form over the sequence length. 
However, the absence of parallelizable operations, such as matrix multiplications, hinder the benefits from Tensorcores,  thus not translating to actual wall-time efficiency. 
In contrast, the quadratic-complexity form of Transformers exhibits high FLOPs, making long-sequence training expensive, even if Qin et al. \cite{qin2023scaling} showed that it can be as efficient as \sigla{FlashAttention}.
The chunkwise form (see \cite{hua2022transformer,sun2023retentive}) provides an interesting compromise, allowing fine-grained optimization by interpolating between the parallel and recurrent forms with the user-selected chunk size.
Differently from the recurrent form, most operations can be done via matmuls, enabling the use of tensor cores. However, most implementations
are not I/O-aware and thus slower than \sigla{FlashAttention} for moderate sequence lengths (e.g., 2K-4K).


When considering solutions to the aforementioned issues, \sigla{FlashAttention} \cite{dao2022flashattention} showed how even vanilla Transformers could benefit from I/O aware techniques. The authors of \cite{dao2022flashattention} restructure (\textit{tiling}) the attention computation to prevent reading and writing (\textit{materializing}) the large $L \times L $ attention matrix to and from the slow \sigla{HBM}. Meanwhile, in large-scale training and long-sequence modeling scenarios where the batch size tends to be small, parallelizing over the temporal dimension enables high GPU occupancy \cite{dao2023flashattention}.


Following similar intuitions, few recent works are starting to devise novel I/O-aware mechanisms. 
The \sigla{GLA} model proposed by Yang et al. \cite{yang2023gated} (see Section \ref{sec:transformers}) borrows concepts from Linear Transformers, gating mechanism and \sigla{FlashAttention}, resulting in the proposal of \sigla{FlashLinearAttention}, an hardware-efficient implementation of Linear Attention in a chunkwise form, also generalized to a more expressive variant with data-dependent gates. The authors propose a memory-efficient version where chunk-level states are not materialized in \sigla{HBM} memory, by leveraging tiling and recomputation, and a  materialization version, which leverages sequence-level parallelism and thus allows for higher training throughput at the cost of a slightly increasing memory footprint.
Overall, \sigla{FlashLinearAttention} is faster than \sigla{FlashAttention-2} \cite{dao2023flashattention}, despite the introduction of an expressive gating mechanism. 
A concurrent work by Qin et al. \cite{qin2024lightning} also proposed an I/O-aware version of linear attention, which is similar to the non-materialization version
of \sigla{FlashLinearAttention}. 


The selection mechanism underlying \sigla{Mamba} \cite{gu2023mamba} results in a time-varying \sigla{SSM}. Time varying solutions (such as Hippo \cite{gu2020hippo}) had been previously overtaken by LTI (non-selective) models (S4, S5 and all derivatives)  due to the fact that the latter can be efficiently expressed in the form of global convolutions or associative scans, while the former require to compute and materialize the large latent state into \sigla{HBM}. 
However, the additional expressive power achieved by means of ``selection'' procedures, calls for the need of an efficient hardware-aware implementation. Indeed, similarly to \sigla{GLA}, \sigla{Mamba} leverages kernel fusion (to reduce the amount of memory I/Os),
parallel scan (to avoid slowdowns due to the sequential recurrence), and recomputation (the intermediate states are not
stored but recomputed in the backward pass when the inputs are loaded from \sigla{HBM} to SRAM). Overall, \sigla{Mamba} shows the same  memory requirements as an optimized transformer with \sigla{FlashAttention}.

To address the low FLOPs-to-byte ratio due to elementwise operations,  the \sigla{Griffin} model \cite{de2024griffin} is implemented in TPUs and leverages a custom kernel for the computation of the state update equation 
exploiting linear scans. The implemented kernel minimizes memory transfers by keeping hidden states in the TPUs Vector Memory (\sigla{VMEM}) instead of \sigla{HBM} and performing memory transfers in larger chunks, resulting in a $3 \times$ speedup with respect to a vanilla implementation. As an interesting remark, the gating mechanism in \sigla{Griffin} is not compatible with the usual convolutional view or associative scans proper of LTI \sigla{SSMs}. Even if the authors could have used associative scans similarly to \sigla{GateLoop} \cite{katsch2023gateloop}, they found that these operations reduce the number of FLOPs, but do not reduce memory overheads, which is their primary bottleneck. 
Summing up, when dealing with long-sequence processing, a very relevant role is played by the way computations are optimized, and identifying primary performance bottlenecks on the target hardware or modern accelerators represents a fundamental step for future research.  Novel models in this research field could greatly benefit from ad-hoc hardware and algorithmic solutions.

\subsection{Infinite-length Sequences: Lifelong Learning}

This survey describes multiple architectural and algorithmic solutions  that allow neural networks to handle sequences of increasing length, by reducing the overall computational complexities and leveraging a stateful recurrent representation.
In the previous subsections, we showcased two very relevant research avenues for future studies, i.e., the expressivity improvement of recurrence-based models with respect to full-fledged attention, and the need for hardware-aware solutions capable of effectively dealing with very long sequences. 
However, there is still a significant gap that requires specific studies to move beyond current technologies, that shows up when thinking of transitioning from {\it long} sequences to {\it infinite-length} sequences. Indeed, while researchers aim to mimic human learning abilities from sequential data, the most common existing approaches primarily deal with finite-length datasets \sigla{BPTT} in stateful models, or standard Back-Propagation, whilst the alternative learning algorithms we described in Section \ref{sec:forward} are still not widespread in the literature. This stands in stark contrast to human learning from continuous perceptual stimuli, which unfold over time without being pre-segmented into finite-length sequences for stochastic gradient descent optimization, and without the need to explicitly store exemplars into buffers or the need of building data collections~\cite{collectionlessAI}.
Moreover, this misalignment deviates from the principles of seminal works in recurrent models. Indeed, the goal of works such as the one by Williams and Peng \cite{williams1990efficient} was to devise {\it ``an online algorithm, designed to be used to train a network while it runs; no manual state resets or segmentations of the training stream {are} required''}. Even the original formulations of \sigla{LSTMs}, among others \cite{hochreiter1997long}, were introduced with a learning algorithm that unlike \sigla{BPTT} is {\it ``local in space and time''}, where {\it ``there is no need to store activation values observed during sequence processing in a stack with potentially unlimited size''}.

In contrast, as we described in the previous Sections, the most popular research activities in the field of \sigla{RNNs} are not considering the case of learning {\it online} from a continuous, possibly infinite, stream of data (i.e., infinite-length sequence), without resets or segmentations \cite{gori_collas2022}. 
This extremely challenging scenario involves two primary demands: ($i.$) computing meaningful gradients at time $t$ based on processed data up to that point, without fully unrolling the network across the entire (potentially infinite) input sequence (in some cases, only relying on information from times $t$ and $t-1$), and updating model parameters at each time step; ($ii.$) possessing the ability to adapt to the characteristics of the streamed data, retaining acquired skills over time, as emphasized in continual learning contexts \cite{parisi2020online}.
In Section \ref{sec:forward} we described the scientific efforts aimed at developing alternatives to  \sigla{BPTT}. We showcased online learning algorithms for \sigla{RNNs}, particularly addressing requirement ($i.$).  \sigla{RTRL} stands as the most recognized method for online learning with recurrent models, which has been recently refined to mitigate computational and memory constraints \cite{irie2023exploring}, alongside various direct or indirect adaptations \cite{marschall2020unified}. 
We described recent studies that explore forward computation of gradients, enabled by state-space modeling \cite{zucchet2023online}, albeit reliant on finite-length sequences processed without updating learnable parameters at each time step, thereby not fully satisfying requirement ($i.$).
Considering demand ($ii.$) leads to a scenario akin to Online Continual Learning \cite{parisi2020online} with substantial overlap with Streaming Learning, as outlined in \cite{gunasekara2023survey}. It's important to distinguish this from approaches focused solely on continual learning from streams of finite-length sequences \cite{cossu2021continual,ehret2020continual}.

Summing up, the scenario described in this section remains largely unexplored. Despite advancements in related domains, meeting requirements ($i.$) and ($ii.$) independently poses significant challenges, and their combined fulfillment presents an even more daunting task. Nonetheless, this scenario reflects the reality of learning from continuous streams of visual data, auditory stimuli, perpetual video feeds, language inputs,  {and more}. Whilst common benchmarks for sequential data (see Section \ref{sec:benchmarks}) do provide valuable opportunities to assess the fulfillment of requirement ($i.$), it is  crucial to operate within the typical framework of online learning problems, where all data contributes to training and errors accumulate during learning.\footnote{The input at time $u_t$ is processed and the produced output $y_t$ is compared with the available ground truth to update the accuracy estimate; then, $y_t$ is used to evaluate a loss function, compute gradients at $t$ and update the model.}
Certain tasks such as time-series forecasting offer greater flexibility in this context, compared to classification tasks, due to the fact that they provide ground truth at each $t$. Additionally, the properties of the series {might} change over time, partially allowing exploration of requirement ($ii.$). However, a comprehensive evaluation of property ($ii.$) remains challenging with existing datasets.

All these considerations suggest that there is room for future research. Providing sources for streaming data becomes extremely important, in a context where data privacy and storage become more and more relevant \cite{collectionlessAI}. For instance, in computer vision, a paradigm-shift from image-based processing to videos is happening \cite{selva2023video,tiezzi2022stoca,faggi2023local,betti2020learning,meloni2021evaluating,carreira2023learning,videoworldsimulators2024}, thus requiring to rethink data handling pipelines and data engineering. 
From the pure machine learning side, research on novel models and learning methods which are better at capturing regularities on the input stream in an online manner are needed. At the same time, model evaluation and selection procedures must evolve \cite{ghunaim2023real}, to be capable of handling settings where models self-adapt their parameters over time continuously, potentially in a truly lifelong learning manner.

\section{Conclusions}
\label{sec:conclusions}
While Transformers initially overshadowed Recurrent Neural Networks, recent developments in large-context Transformers and (deep) State-Space Models have highlighted the importance of recurrent computations as a viable road {to} go beyond the limitations of Transformers both in terms of performance on long sequences and scalability. This survey outlines the latest trends and approaches for sequential data processing, offering insights {into} recent architectural and novel algorithmic solutions.

Interestingly, the recent literature in the context of recurrent long-context Transformers and  State-Space Models share the same features, leading towards a computational model based on stateful computations, confirming the advantages brought by the idea of working in the state space. We surveyed the most important works, and we discussed novel trends in learning algorithms for sequential data, offering a broad and detailed perspective on the last generation of neural architectures. This survey also outlined novel research opportunities. In fact, we emphasized how the research direction of lifelong online learning from a stream of data, intended to be an infinite-length sequence, represents a challenge that is still very open and might be faced starting from this renewed interest in recurrent models.

\bibliography{biblio}{}
\bibliographystyle{elsarticle-num}

\end{document}